\documentclass{article}

\usepackage[english]{babel}

\usepackage[letterpaper,top=2cm,bottom=2cm,left=3cm,right=3cm,marginparwidth=1.75cm]{geometry}

\usepackage{amsmath,amssymb,amsthm}
\usepackage{mathtools}
\usepackage{graphicx}
\usepackage{xcolor}
\usepackage[colorlinks=true, allcolors=blue]{hyperref}
\usepackage[sort]{natbib}
\usepackage{xspace}
\usepackage{booktabs}
\usepackage{array}
\usepackage{enumitem}
\usepackage{algorithm}
\usepackage{algorithmic}
\usepackage{float}
\usepackage{microtype}
\usepackage{authblk}

\graphicspath{{./}{../}}

\newcommand{\bone}{\boldsymbol{1}} 
\newcommand{\bx}{\boldsymbol{x}} 
\newcommand{\by}{\boldsymbol{y}} 
\newcommand{\bz}{\boldsymbol{z}} 
\newcommand{\bg}{\boldsymbol{g}} 
\newcommand{\bu}{\boldsymbol{u}} 
\newcommand{\btheta}{\boldsymbol{\theta}} 

\newcommand{\sA}{\mathcal{A}} 
\newcommand{\sK}{\mathcal{K}} 
\newcommand{\sL}{\mathcal{L}} 
\newcommand{\sO}{\mathcal{O}} 
\newcommand{\sE}{\mathsf{Alg}} 
\newcommand{\sZ}{\mathcal{Z}} 
\newcommand{\sN}{\mathcal{N}} 
\newcommand{\sC}{\mathcal{C}} 

\newcommand{\seq}[1]{\left(#1\right)}
\newcommand{\set}[1]{\left\{#1\right\}}
\newcommand{\norm}[1]{\left\|{#1}\right\|}
\newcommand{\abs}[1]{\left| #1 \right|}
\newcommand{\inn}[2]{\left\langle{#1}, {#2}\right\rangle}

\newcommand{\R}{\mathbb{R}}

\DeclareMathOperator{\Cov}{Coverage}
\DeclareMathOperator{\MisCov}{MisCov}
\DeclareMathOperator{\Regret}{Regret}
\DeclareMathOperator{\LinRegret}{LinRegret}

\newcommand{\MultiQT}{\texttt{MultiQT}\xspace}
\newcommand{\ACI}{\texttt{ACI}\xspace}
\newcommand{\UPOCP}{\texttt{UP-OCP}\xspace}
\newcommand{\KT}{\texttt{KT}\xspace}
\newcommand{\PNormOMD}{\texttt{$p$-norm OMD}\xspace}
\newcommand{\Iso}{\texttt{ISO}\xspace}
\theoremstyle{plain}
\newtheorem{theorem}{Theorem}[section]
\newtheorem{proposition}[theorem]{Proposition}
\newtheorem{lemma}[theorem]{Lemma}
\newtheorem{corollary}[theorem]{Corollary}
\theoremstyle{definition}
\newtheorem{definition}[theorem]{Definition}
\newtheorem{assumption}[theorem]{Assumption}
\theoremstyle{remark}
\newtheorem{remark}[theorem]{Remark}

\newcommand{\bG}{\boldsymbol{G}}
\newcommand{\bv}{\boldsymbol{v}}
\newcommand{\bw}{\boldsymbol{w}}
\newcommand{\bzero}{\boldsymbol{0}}
\newcommand{\bdelta}{\boldsymbol{\delta}}
\newcommand{\bxtilde}{\tilde{\boldsymbol{x}}} 

\newcommand{\sB}{\mathcal{B}}

\newcommand{\sH}{\mathcal{H}}
\newcommand{\sI}{\mathcal{I}}

\newcommand{\sS}{\mathcal{S}}
\newcommand{\sX}{\mathcal{X}}

\DeclareMathOperator{\IS}{IS}
\DeclareMathOperator{\WIS}{WIS}
\DeclareMathOperator{\dist}{dist}

\newcommand{\sign}{{\rm sign}}

\newcommand{\closure}{\mathrm{cl}}
\newcommand{\boundary}{\mathrm{bd}}

\DeclareMathOperator{\KL}{KL}

\renewcommand{\epsilon}{\varepsilon}

\title{A Black-Box Reduction from Regret to Multi-Level Coverage}

\author[1]{Tuo Liu\thanks{\texttt{tuo.liu@kaust.edu.sa}}}
\author[2]{Edgar Dobriban\thanks{\texttt{dobriban@wharton.upenn.edu}}}
\author[1]{Francesco Orabona\thanks{\texttt{francesco.orabona@kaust.edu.sa}}}

\affil[1]{King Abdullah University of Science and Technology, Saudi Arabia}
\affil[2]{University of Pennsylvania, USA}

\begin{document}
\maketitle

\begin{abstract}
Online Conformal Prediction (OCP) sequentially constructs prediction sets whose average empirical coverage converges to a target level over time, for arbitrary, potentially adversarial datasets.
In the multi-level setting, prediction sets are produced at $d$ coverage levels at once.
Prior work argued for the use of online gradient descent with isotonic projections, which produces sets nested across levels, but with a $\sqrt{d}$ dependence in the per-level coverage convergence rate.
Here, we both generalize and improve prior work. First, we show that isotonic projections can be used for the construction of multi-level online conformal prediction sets from \emph{any} online learning algorithm (not just gradient descent).
Then, through a black-box reduction, we show that the linearized regret of the online algorithm itself automatically gives control over both per-level coverage and distributional consistency of multi-level predictions via Fenchel duality.
Finally, by using the Weighted Interval Score, a standard scoring rule for assessing multi-level interval forecasts, we prove that these projections reduce both the size of the sets and their under-coverage.
We study two instantiations of our theory, $p$-norm Online Mirror Descent and coordinate-wise Universal Portfolio methods, both improving the dependence on the number of levels to $\sqrt{\log d}$ for the multi-level OCP problem.
Experiments on several datasets show favorable performance compared to  baselines.
\end{abstract}

\addtocontents{toc}{\protect\setcounter{tocdepth}{-1}}

\section{Introduction}
\label{sec:intro}

Constructing prediction sets with reliable coverage guarantees is a central goal of uncertainty quantification. Conformal prediction provides a distribution-free framework for this task: it constructs a prediction set that contains the realized outcome with a user-specified probability, while making few assumptions about the underlying predictive model~\citep{saunders1999transduction,vovk1999machine,papadopoulos2002inductive,vovk2005algorithmic,shafer2008tutorial}. 
In the classical exchangeable setting, conformal methods have been developed for regression, classification, and structured prediction~\citep[see, e.g.,][]{lei2013distribution,lei2018distribution,barber2021predictive,romano2020classification}.
This assumption, however, can be violated in settings where the data distribution shifts, such as sequential decision-making and time-series forecasting.

Online conformal prediction (OCP) addresses these limitations by operating in a \emph{fully adversarial sequential setting}, where the data may be generated by an arbitrary and possibly adversarial process.
A growing body of work has developed algorithms for single-level OCP, in which the goal is to maintain long-run coverage at a single target level $1 - \alpha$ for any data sequence. 
These include Adaptive Conformal Inference (\ACI)~\citep{gibbs2021adaptive}, an FTRL-based approach~\citep{zhang2024discounted}, conformal PID control~\citep{angelopoulos2023conformal}, meta-algorithms~\citep{zaffran2022adaptive,bastani2022practical,bhatnagar2023improved,gibbs2024conformal}, and betting-based methods~\citep{podkopaev2024adaptive,liu2026online}.
Recently, \citet{liu2026online} established a black-box reduction showing that \emph{any} online learning algorithm with a linearized regret guarantee can be converted into a single-level OCP algorithm with provable coverage. 
This reveals a rich landscape of available algorithms for the single-level problem.

However, many real-world applications can benefit from prediction sets that have simultaneous validity at multiple different levels of coverage. 
This can allow users to inspect the prediction sets \emph{ex post facto} and choose one of them that fits their needs in terms of coverage, length, as well as any potential downstream risk calculations.

Motivated by this, we consider the setting of multi-level online conformal prediction recently proposed by \citet{ding2025calibrated}.
Observed data $((Z_{t}, Y_{t}))_{t \geq 1}$ arrive over time $t=1, 2, \dots$.
At each round $t$, a base model produces a point forecast $\hat{Y}_t$ for the response $Y_t$, using all the previously observed data points, as well as the current features $Z_{t}$.
We say coverage is attained at $t$ for the prediction set $\hat{C}_{t}(\hat{Y}_t, x)$, parameterized by a threshold $x$, iff it contains $Y_t$, i.e., $Y_{t} \in \hat{C}_{t}(\hat{Y}_t, x)$.
We fix $d$ target miscoverage levels $\sA = \set{\alpha_1, \dots, \alpha_d} $, where $0<\alpha_1 < \alpha_2 < \dots < \alpha_d<1$. The conformal predictor announces a vector of thresholds $\bx_t = (x_{t, 1}, \dots, x_{t, d})$ before $Y_t$ is revealed, where $x_{t, i}$ targets a long-run coverage rate of $1 - \alpha_i$.
Two properties are essential for multi-level forecasts:
\begin{enumerate}[nosep, leftmargin=*]
    \item Coverage: for any sequence of data $((Z_{t}, Y_{t}))_{t\geq1}$, even adversarially chosen, the long-run observed coverage at each level should converge to the nominal target as the time horizon $T$ grows:
    \[
        \lim_{T \to \infty} \MisCov_{T, i} = 0, \ \forall i \in [d], \text{ where }
        \MisCov_{T, i} := \abs{\frac{1}{T} \sum_{t = 1}^T \bone\set{S_t \leq x_{t, i}} - (1 - \alpha_i)}.
    \]
    \item Distributional consistency. 
    Forecasts should be ordered across quantile levels, i.e., larger levels should give larger sets: $x_{t, 1} \geq x_{t, 2} \geq \dots \geq x_{t, d}$, for all $t = 1, 2, \dots$.
\end{enumerate}
We will refer to  these two properties as \emph{multi-level coverage}.

\citet{ding2025calibrated} proposed the \MultiQT algorithm, which wraps Online Gradient Descent (OGD) with a lazy isotonic projection step to enforce ordering.
The algorithm provably achieves asymptotic multi-level coverage.
However, their result applies to a single algorithm instance. 
Meanwhile, as said above, the single-level OCP literature offers a diverse toolbox of algorithms, provably better than OGD.
It remains unclear how to migrate these algorithms to the multi-level setting (ideally, with a black-box reduction), while preserving multi-level coverage.

\textbf{Contributions.} In this work, we answer the above open questions with the following contributions:
\begin{itemize}[nosep,leftmargin=*]
\item In Section~\ref{sec:reduction}, we present a general \emph{black-box reduction}: any online learner satisfying a linearized regret bound properly wrapped with an isotonic projection layer yields a multi-level OCP algorithm that provably guarantees multi-level coverage.

\item Instantiating our framework with $p$-norm Online Mirror Descent~\citep{Gentile03} and coordinate-wise Universal Portfolio algorithms~\citep{cover2002universal} yields per-level miscoverage rates that depend on $\sqrt{\log d}$, improving over the $\sqrt{d}$ dependence of the general guarantee in \citet{ding2025calibrated}, and, for the latter, with no parameter tuning under polynomially growing scores. The improvement is in the dependence on $d$ (rather than on the horizon $T$; see Table~\ref{tab:prior-comparison} for the regime where \MultiQT attains a faster rate in $T$).

\item In Section~\ref{sec:projection-benefits}, we show that when the isotonic projection acts, it strictly reduces the \emph{sharpness reward} and weakly reduces the \emph{coverage penalty} of the \emph{Weighted Interval Score} (WIS)~\citep{winkler1972decision,gneiting2007strictly,bracher2021evaluating}.
This gives a precise theoretical understanding for the empirical observation that projection improves the Pareto frontier of coverage-sharpness trade-offs.

\item In Section~\ref{sec:experiments}, we evaluate our algorithms on synthetic and real-world datasets, demonstrating consistent improvements in the Pareto frontiers and in the WIS over existing baselines.
\end{itemize}

\textbf{Related Work.}
Our work sits at the intersection of online conformal prediction, online learning, and multi-level coverage. Adaptive conformal inference and its variants treat the conformal radius as a sequentially updated decision variable, often optimized through the pinball loss~\citep{gibbs2021adaptive,zaffran2022adaptive,gibbs2024conformal}. Strongly adaptive and control-based methods improve local responsiveness or empirical stability, but require method-specific choices such as expert grids, gain parameters, or range assumptions~\citep{angelopoulos2023conformal,bhatnagar2023improved}. Parameter-free approaches obtain optimal regret guarantees uniformly over all comparators, by avoiding explicit stepsize tuning~\citep{podkopaev2024adaptive,zhang2024discounted,liu2026online}, and can be extended to group-conditional coverage at a single level~\citep{bharti2026parameter}. Among these works, the closest one to our work is \citet{liu2026online}, which proves a black-box linearized-regret-to-coverage implication, but only for scalar OCP, i.e., a single coverage level.
Related analyses distinguish standard regret from the stronger conditions needed for coverage, including gradient-equilibrium and proximal-regret viewpoints~\citep{angelopoulos2025gradient,cai2024tractable}.

The multi-level setting is connected to, but distinct from, simultaneous and multivalid forms of conformal inference, where one controls empirical coverage across groups or threshold families~\citep{bastani2022practical}, while simultaneous conformal and conformal risk-control methods handle many inference targets or losses at once~\citep{angelopoulos2022conformal,lee2024simultaneous}. Instead, our focus is an adversarial online protocol for an ordered family of conformal sets indexed by miscoverage levels.
Finally, our sharpness interpretation is aligned with proper interval and quantile scoring rules~\citep{winkler1972decision,gneiting2007strictly,bracher2021evaluating}.
Table~\ref{tab:prior-comparison} in Appendix~\ref{app:prior-comparison} compares our results with the two  works we build on closely.

\section{Problem setup}
\label{sec:setup}

In this section, we formalize the multi-level OCP problem outlined in Section~\ref{sec:intro}.

\textbf{Notation.}
We collect a few standard tools from convex analysis in $\R^d$~\citep[see, e.g.,][]{rockafellar1997convex} for reference.
For a function $f \colon \R^d \to \R \cup \{+\infty\}$, a vector $\bg \in \R^d$ is a \emph{subgradient} of $f$ at $\bx \in \R^d$ if $f(\by) \geq f(\bx) + \inn{\bg}{\by - \bx}$ for every $\by \in \R^d$, and the set of all such $\bg$ is the \emph{subdifferential} $\partial f(\bx)$.
The \emph{indicator function} $\bone\{A\}$ is equal to $1$ if $A$ is true and 0 otherwise.
The \emph{Fenchel conjugate} of $f \colon \R^d \to [-\infty, +\infty]$ is $f^\star(\btheta) \coloneqq  \sup_{\bx \in \R^d} \inn{\btheta}{\bx} - f(\bx)$, which is always convex and closed, i.e., its epigraph is a closed set (equivalently, $f^\star$ is lower semicontinuous).
The \emph{Euclidean projection} onto a nonempty closed convex set $\mathcal{C} \subseteq \R^d$ is $\Pi_\mathcal{C}(\bx) \coloneqq  \arg\min_{\bz \in \mathcal{C}} \norm{\bx - \bz}_2$, which is well-defined and unique.
We denote by $[d] \coloneqq  \{1, \dots, d\}$.

\textbf{Multi-level OCP.}
We now describe the sequential protocol of OCP with multiple quantile levels.
We choose to focus on symmetric prediction sets of the form $\hat{C}_{t}(\hat{Y}_t, x) \coloneqq  [\hat{Y}_t - x, \hat{Y}_t + x]$, defined as empty if $x < 0$~\citep{lei2018distribution}.
Our methodology is applicable beyond this setting: it only needs a nested family of sets parameterized by one scalar per level; e.g., it can be used to adjust the prediction intervals based on conditional quantile regression models~\citep{romano2019conformalized}, but we avoid the details for brevity. We refer the reader to \citet{gupta2022nested} for more versions of the nested prediction sets for which our methodology is applicable.
Let $S_{t} \geq 0$ denote the radius of the smallest prediction set that contains the true response $Y_{t}$, i.e.,
$S_{t} \coloneqq  \inf \{x \in [0,\infty): Y_{t} \in \hat{C}_{t}(\hat{Y}_t, x)\} = |Y_{t} - \hat{Y}_{t}|$.
With this choice, $Y_t \in \hat{C}_t(\hat{Y}_t, x) \iff S_t \leq x$ for $x > 0$, and the coverage event at level $1 - \alpha_i$ collapses to the single threshold comparison $\bone\{S_t \leq x_{t, i}\}$. We will also refer to $S_t$ as the non-conformity score \citep{vovk2005algorithmic}.

Note that a negative radius does not describe a valid interval, so our algorithms output the empty set instead, represented by a $0$ radius, the Lebesgue measure of $\emptyset$ (Section~\ref{sec:reduction}). Accordingly, a played radius of $0$ is read as $\emptyset$, which never covers $Y_t$, even when $S_t = 0$. 

As we said in Section~\ref{sec:intro}, we require per-level \emph{coverage}
and \emph{distributional consistency}. For the second one, the increasing miscoverage levels $\alpha_1 < \cdots < \alpha_d$ should be matched by non-increasing thresholds $x_{t, 1} \geq \cdots \geq x_{t, d}$. Equivalently, $\bx_t \in \sK$ for every $t$, where $\sK \coloneqq  \{\bx \in \R^d : x_1 \geq x_2 \geq \cdots \geq x_d\}$ is the \emph{isotonic cone}.\footnote{By convention, we choose the non-increasing cone.}

\textbf{Pinball loss and its aggregate.}
As in the single-level case, we introduce the \emph{pinball (or quantile) loss} for online quantile learning \citep{gibbs2021adaptive,podkopaev2024adaptive}, defined as
\begin{equation} \label{eq:pinball-loss}
    \ell^{(1 - \alpha)}(x, S)
    \coloneqq  \max \{(1 - \alpha) (S - x), \alpha (x - S)\},
\end{equation}
where $S$ is the non-conformity score and $x$ is the radius of the prediction interval.
This loss is convex and $L$-Lipschitz in the first argument, where $L \coloneqq  \max \{1 - \alpha, \alpha\}$.
Intuitively, the pinball loss is a proper surrogate for the $(1 - \alpha_i)$-th quantile, as under any distribution of $S$, the expectation of $\ell^{(1 - \alpha_i)}(x, S)$ in $x$ is minimized exactly at the $(1 - \alpha_i)$-th quantile of $S$~\citep{gneiting2007strictly}.

To handle the $d$ levels jointly, we work with the \emph{aggregate pinball loss} $\sL_t(\bx) \coloneqq  \sum_{i = 1}^d \ell^{(1 - \alpha_i)}(x_i, S_t)$,  for $\bx \in \R^d$.
We select the subgradient $\bg_t \in \partial \sL_t(\bx_t)$ with coordinates
\begin{equation} \label{eq:pinball-subgradient}
    g_{t, i} = \bone\{x_{t, i} \geq S_t\} - (1 - \alpha_i), \quad \forall i \in [d].
\end{equation}
Note that $\sL_t$ is Lipschitz in both the $\ell_2$ and $\ell_1$ (dual of $\ell_\infty$) norms: $\norm{\bg_t}_2\leq \sqrt{d}$ and
$\norm{\bg_t}_\infty \leq \max\{1 - \alpha_1, \alpha_d\} \leq 1$.
These are favorable conditions for online learnability~\citep[see, e.g.,][]{hazan2016introduction,orabona2019modern,cesa2021online}.
Performance on the aggregate loss is measured by the standard notion of \emph{regret}. Against a fixed comparator $\bu \in \R^d$, the regret of an algorithm producing $\seq{\bx_t}_{t \geq 1}$ is $
\Regret_T(\bu) \coloneqq  \sum_{t = 1}^T \sL_t(\bx_t) - \sum_{t = 1}^T \sL_t(\bu)$.

Finally, we note that the subgradient \eqref{eq:pinball-subgradient} admits a direct interpretation as a miscoverage error:
\begin{equation}
    \label{eq:miscov-subgrad}
    \MisCov_{T, i} \coloneqq  \frac{1}{T} \abs{\sum_{t = 1}^T g_{t, i}}, \quad \forall i \in [d].
\end{equation}

We use a single, mild assumption on the polynomial growth of the score sequence.
\begin{assumption}
\label{ass:poly-growth}
There exist constants $D > 0$ and $\gamma \geq 0$ such that $S_t \leq D T^\gamma$ for every $t \in [T]$.
\end{assumption}
This assumption generalizes the bounded-score assumption $S_t \leq D$ used in much of the OCP literature (recovered by $\gamma = 0$), and accommodates non-stationary streams whose magnitudes grow with the horizon (e.g., financial returns with volatility clustering~\citep{liu2026online}).

\section{From regret to multi-level coverage: a black-box reduction}
\label{sec:reduction}

Our goal is to design a black-box reduction for an online algorithm: (1) to have iterates $\bx_t$ lie in the isotonic cone $\sK$ at every round; (2) to secure per-level coverage by controlling the cumulative subgradient $\bG_T \coloneqq  \sum_{t = 1}^T \bg_t$, via~\eqref{eq:miscov-subgrad}.
Linearized regret is the natural interface for (2): $G_{T,i}$ is $T$ times the signed coverage error at level $i$, and a bound on $\sum_t \inn{\bg_t}{\bx_t - \bu}$ holding for every comparator $\bu$ is, by duality, a bound on $\bG_T$.
We resolve both issues through a single mechanism: a \emph{lazy isotonic projection layer, applied as a black-box wrapper around any unconstrained online learner $\sE$ with a linearized-regret guarantee}.

\textbf{The lazy isotonic projection wrapper.}
Let $\sE$ be a generic online learner. We make no assumption on its internal structure: $\sE$ is a black box that, at each round, produces a vector of radii $\tilde{\bx}_t \in \R^d_+$ and accepts an arbitrary subgradient. The wrapper inserts $\sE$ with a single Euclidean projection step:

\begin{algorithm}[h]
\caption{Lazy projection wrapper $\sE \rightarrow$ multi-level OCP}
\label{alg:lazy-projection}
\begin{algorithmic}[1]
  \STATE \textbf{Input:} base learner $\sE$ with outputs in $\R^d_+$; isotonic cone $\sK$
  \FOR{$t = 1, 2, \dots$}
    \STATE Receive hidden iterate $\bxtilde_t \in \R^d_+$ from $\sE$
    \STATE Play projected iterate $\bx_t = \Pi_\sK(\bxtilde_t)$
    \STATE Observe $S_t$ and compute $\bg_t \in \partial \sL_t(\bx_t)$ via~\eqref{eq:pinball-subgradient}
    \STATE Feed $\bg_t$ to $\sE$, which produces $\bxtilde_{t+1}$
  \ENDFOR
\end{algorithmic}
\end{algorithm}

The defining feature of this protocol is the asymmetry between the played iterate $\bx_t$ (projected onto $\sK$) and the gradient input $\bg_t$ to $\sE$ (evaluated at the projected iterate, then routed back into the unconstrained dynamics). This pattern is known as \emph{lazy projection}~\citep{Zinkevich03}.
Despite being embedded in \MultiQT~\citep{ding2025calibrated}, the generality of such a structure was largely unrecognized.
Here, we treat the projection as a strictly external layer.

Any online learner on $\R^d$ becomes a base learner with outputs in $\R^d_+$ through the convention of Section~\ref{sec:setup}: whenever its raw radius $z_{t, i}$ is negative, it outputs the empty set at level $i$, represented as $\tilde{x}_{t, i} = \max(z_{t, i}, 0) = 0$. The learner's update rule and internal state are unchanged. It also makes the input of the isotonic layer a valid vector of radii, which Section~\ref{sec:projection-benefits} exploits.

To start off, observe that, since $\bx_t \in \sK$ at every round, the empirical coverage rates inherit the same monotonicity: $\Cov_{T, 1} \geq \Cov_{T, 2} \geq \cdots \geq \Cov_{T, d}$, where $\Cov_{T, i} \coloneqq  (1 / T) \sum_{t = 1}^T \bone \{S_t \leq x_{t, i}\}$. This is an automatic property from the algorithm, coherent to multi-level coverage, with no extra analysis.

\textbf{Linearized regret preservation through inward flow.}
Two features of the wrapper need to be reconciled with the regret guarantee: (1) The projection step moves $\tilde{\bx}_t$ inside $\sK$ before predicting, and (2) $\sE$ is updated with $\bg_t \in \partial \sL_t(\bx_t)$ evaluated at the played iterate rather than at the iterate $\sE$ produced. Both could in principle inflate the regret. The \emph{inward flow} property of~\citet{ding2025calibrated} ensures they do not. We restate it for any abstract loss-set pair, since this generality is essential below.
\begin{definition}[Inward flow~\citep{ding2025calibrated}]
\label{def:inward-flow}
A function $f$ paired with a closed convex set $\sC \subseteq \R^d$ satisfies the \emph{inward flow} property if there exists a subgradient $\bg(\bx) \in \partial f(\bx)$ such that $-\bg(\bx) \in T_{\sC}(\bx)$ for every boundary point $\bx \in \boundary(\sC)$, where $T_{\sC}(\bx) \coloneqq  \closure\{\bv: \exists \delta > 0, \bx + \varepsilon \bv \in \sC \text{ for } \varepsilon \in (0, \delta]\}$ is the tangent cone of $\sC$ at $\bx$, $\boundary(\sS)$ denotes the boundary of a set $\sS$, and $\closure(\sS)$ denotes the closure of $\sS$.
\end{definition}
Geometrically, the chosen subgradient at the boundary points outward, so the descent direction points inward, and a small step opposite $\bg(\bx)$ stays feasible. For the isotonic cone, the meaning is clear: two tied levels $x_i = x_{i + 1}$ share the same coverage indicator, so $g_i = \bone\{x_i \geq S_t\} - (1 - \alpha_i) \leq g_{i + 1}$, and a descent step raises the radius at level $i$ at least as much as at level $i + 1$, never creating a crossing. \citet[Lemma~2]{ding2025calibrated} verifies inward flow for the aggregated pinball $\sL_t$ paired with the (decreasing) isotonic cone $\sK$ under the choice of subgradient as in~\eqref{eq:pinball-subgradient}; we adopt this throughout.
The structural consequence relevant to our analysis is the following projection-and-gradient inequality.

\begin{lemma}[Projection inequality~\citep{ding2025calibrated}]
\label{lem:projection-inequality}
Let $f$ with subgradient $\bg$ and a closed convex set $\sC \subseteq \R^d$ satisfy inward flow. For every $\bx \in \R^d$, $\inn{\bg(\Pi_{\sC}(\bx))}{\bx} \geq \inn{\bg(\Pi_{\sC}(\bx))}{\Pi_{\sC}(\bx)}$.
\end{lemma}
A self-contained proof, which leverages the duality between the tangent cone (containing $-\bg_t$) and the normal cone (containing the projection direction $\tilde{\bx}_t - \bx_t$), is in Appendix~\ref{app:projection-inequality}. We state it for a general $\sC$ because inward flow composes under intersection with polyhedral sets, so further polyhedral constraints can be layered on top of $\sK$ (Appendix~\ref{app:beyond-isotonic}). 

Lemma~\ref{lem:projection-inequality} converts a regret bound on the hidden iterates of $\sE$ into a regret bound on the played iterates and allows us to prove the following regret guarantee for Algorithm~\ref{alg:lazy-projection}.
\begin{proposition}[Linearized regret preservation]
\label{prop:regret-preservation}
Suppose $\sE$ admits a linearized regret bound $F_T(\bu)$ on the gradient sequence $(\bg_t)_{t \in [T]}$, that is, $\sum_{t=1}^T \inn{\bg_t}{\bxtilde_t - \bu} \leq F_T(\bu)$ for every $\bu \in \R^d$. The played iterates $\bx_t = \Pi_\sK(\bxtilde_t)$ produced by Algorithm~\ref{alg:lazy-projection} satisfy
\begin{equation}
\label{eq:regret-preservation}
  \LinRegret_T(\bu) \coloneqq  \sum_{t=1}^T \inn{\bg_t}{\bx_t - \bu} \leq F_T(\bu),
  \quad \forall \bu \in \R^d.
\end{equation}
\end{proposition}

A consequence that is worth highlighting is that~\eqref{eq:regret-preservation} holds for every $\bu \in \R^d$, not just for $\bu \in \sK$, so the projection is free of charge against any comparator the base learner could compete with. The proof in Appendix~\ref{app:central-proof} is a direct combination of Lemma~\ref{lem:projection-inequality}, the assumption on linearized regret, and convexity of $\sL_t$.

\textbf{Characterization of miscoverage level sets.}
We now translate the regret guarantee into a finite-time bound on the cumulative subgradient. We show a sketch of the reasoning, while the full argument is given in Appendix~\ref{app:central-proof}.

For notational convenience, define $\bG_T \coloneqq  \sum_{t = 1}^T \bg_t \in \R^d$. First, note that \eqref{eq:regret-preservation}, by regret--reward duality~\citep{orabona2019modern}, is equivalent to
\[
  F_T^\star(- \bG_T)
  = F_T^\star\left(- \sum_{t = 1}^T \bg_t\right)
  \leq -\sum_{t = 1}^T \inn{\bg_t}{\bx_t}
  \leq \sum_{t=1}^T (\sL_t(\bzero)-\sL_t(\bx_t))
  \leq \sum_{t=1}^T \sL_t(\bzero),
\]
where in the second and third inequality we used the convexity and non-negativity of $\sL_t$. Define $B_T \coloneqq  \sum_{t = 1}^T \sL_t(\bzero)$.
Since $S_t \geq 0$, $\sL_t(\bzero) = \sum_{i = 1}^d (1 - \alpha_i) S_t \leq d  S_t$, so under Assumption~\ref{ass:poly-growth}, $B_T \leq d D T^{1 + \gamma}$. Putting all together, we have the following theorem.
\begin{theorem}[Miscoverage level set]
\label{thm:level-set}
Suppose $\sE$ admits the linearized regret bound $F_T$ on the gradient stream produced by Algorithm~\ref{alg:lazy-projection}, and that each $(\sL_t, \sK)$ satisfies inward flow. Then
\begin{equation}
\label{eq:central-inequality}
  F_T^\star\left(-\bG_T\right) \leq B_T,
\end{equation}
where $F_T^\star$ is the Fenchel conjugate of $F_T$ on $\R^d$. Equivalently, $\bG_T$ lies in the level set
\begin{equation}
\label{eq:level-set}
  \sZ_T(F_T) \coloneqq  \set{\bz \in \R^d : F_T^\star(-\bz) \leq B_T}.
\end{equation}
\end{theorem}
The proof follows that of \citet[Thm.~3.1]{liu2026online} with Lemma~\ref{lem:projection-inequality} and convexity of $\sL_t$ in place of their Lemma~A.1; the direct benefit of this combination is the generality of our results (Table~\ref{tab:prior-comparison}).
The geometry of the coverage-error region is characterized entirely in $F_T^\star$, and, through~\eqref{eq:miscov-subgrad}, it dictates how the values $\MisCov_{T, i}$ for $i\in [d]$ are controlled.
One implication is immediate: \emph{Smaller regret implies tighter coverage.}
In particular, optimizing the \emph{dependence of a regret bound on problem parameters} directly tightens the coverage-error region. We show it formally in Appendix~\ref{app:regret-coverage-monotonicity}.
\begin{remark}
\citet{ding2025calibrated} demonstrated that naive approaches to combining single-level coverage with ordering constraints can fail: both post-hoc isotonic projection of independently coverage-controlled quantiles and (non-lazy) projected OGD on the aggregate pinball loss can break per-level coverage.
Our results do not contradict their result. Instead, Algorithm~\ref{alg:lazy-projection} and Theorem~\ref{thm:level-set} generalize their result by showing that any generic online learning algorithm can be used directly to obtain \emph{multi-level coverage}. In fact, we can convert \emph{any} regret bound for $\sE$ into a deterministic constraint on the miscoverage components $\MisCov_{T, i}$, while guaranteeing distributional consistency too.
\end{remark}
The next two subsections instantiate algorithms based on Algorithm~\ref{alg:lazy-projection} and Theorem~\ref{thm:level-set}.

\subsection{Online mirror descent in the $p$-norm}
\label{sub:pnorm}

Fix $p \in (1,2]$ and let $q = p/(p-1) \in [2,\infty)$. Take $\sE$ to be Online Mirror Descent with regularizer $\psi(\tilde{\bx}) = \tfrac{1}{2}\norm{\tilde{\bx}}_p^2$ and constant step size $\eta > 0$, initialized at $\bxtilde_1 = \bzero$. The resulting wrapped algorithm has a standard mirror-descent dual-update form with an isotonic projection at prediction time; we defer the explicit pseudocode (Algorithm~\ref{alg:pnorm-omd}) to Appendix~\ref{app:algorithms}.

Standard mirror-descent analysis gives the regret bound and its conjugate~\citep[see, e.g.,][Sec.~6.7]{orabona2019modern}
\[
  F_T(\bu) = \frac{\norm{\bu}_p^2}{2\eta} + \frac{\eta}{2(p-1)}\sum_{t=1}^T \norm{\bg_t}_q^2,
  \quad
  F_T^\star(\btheta) = \frac{\eta}{2}\norm{\btheta}_q^2 - \frac{\eta}{2(p-1)}\sum_{t=1}^T \norm{\bg_t}_q^2,
\]
where the conjugate uses $(\tfrac{1}{2}\norm{\cdot}_p^2)^\star = \tfrac{1}{2}\norm{\cdot}_q^2$. Substituting into Theorem~\ref{thm:level-set} and bounding $\norm{\bg_t}_q \leq d^{1/q}$ (since $\norm{\bg_t}_\infty \leq 1$) yields the following $\ell_q$ coverage-error ball.

\begin{corollary}[$\ell_q$ coverage-error ball]
\label{cor:p-norm}
Algorithm~\ref{alg:pnorm-omd} with step size $\eta>0$ satisfies
\begin{equation}
\label{eq:lq-coverage}
    \max_{i \in [d]} \ \MisCov_{T, i}
    \leq \frac{\norm{\bG_T}_q}{T}
    \leq \frac{1}{T} \sqrt{\frac{2 B_T}{\eta} + \frac{d^{2/q}  T}{p - 1}}.
\end{equation}
\end{corollary}

The bound~\eqref{eq:lq-coverage} interpolates between the Euclidean case ($q=2$, dependence $\sqrt{d/T}$) and the geometry of the unit hypercube ($q\to \infty$). 
We optimize the parameter pair $(p, \eta)$. Assuming $d \geq 3$ and choosing $q \coloneqq  2 \log d$ (so $p - 1 = 1 / (2 \log d - 1)$ and $d^{2/q} = e$), the second term inside the square root becomes $e (2 \log d - 1) T$. The right-hand side of~\eqref{eq:lq-coverage} is strictly decreasing in $\eta$, so~\eqref{eq:lq-coverage} alone would favor an arbitrarily large step size. We instead balance its two terms by setting $\eta = 2 B_T / (e (2 \log d - 1) T)$, which gives the $\eta \to \infty$ limit up to a factor $\sqrt{2}$ at a finite step size, and yields
\[
    \max_{i \in [d]} \ \MisCov_{T, i}
    \leq \frac{\norm{\bG_T}_{2\ln d}}{T}
    \leq \sqrt{\frac{2 e (2 \log d - 1)}{T}}
    = \sO\left(\sqrt{\frac{\log d}{T}}\right).
\]
A detailed derivation, including verification that the balanced $\eta$ is well-defined under Assumption~\ref{ass:poly-growth}, is in Appendix~\ref{app:p-norm}.

We emphasize that the $\sqrt{\log d}$ dimension dependence represents an exponential improvement over the $\sqrt{d}$ dependence of OGD, i.e., of \MultiQT, which our analysis recovers as the $p=2$ case.
The improvement lies in $d$, not $T$: for a common point forecast with bounded scores, (our setting with $\gamma = 0$), \citet[Cor.~1]{ding2025calibrated} prove a fast $\sO(1/T)$ rate for a \emph{fixed} step size, for which the regret bound grows linearly in $T$, as for any constant step size; balancing the two yields $\sO(T^{-1/2})$ for both~\citep[Sec.~4.3]{ding2025calibrated}. Appendix~\ref{app:prior-comparison} states both regimes with constants.

The optimal step size $\eta$ in $p$-norm OMD, as in \MultiQT, depends on $B_T$, hence on the unknown polynomial-growth parameters $D$ and $\gamma$. In Appendix~\ref{app:eta-sensitivity}, we give a heuristic and a doubling-trick variant with provable guarantees.

\subsection{Coordinate-wise Universal Portfolio}
\label{sub:upocp}

The cleanest way to remove the dependence on $B_T$ in the choice of step size is to replace OMD by a parameter-free algorithm~\citep[Sec.~10]{orabona2019modern}. Since the aggregate pinball loss decomposes coordinate-wise, $\sL_t(\bx) = \sum_{j=1}^d \ell^{(1-\alpha_j)}(x_j, S_t)$, a natural instantiation is to run an independent copy of the scalar universal-portfolio learner \UPOCP \citep{liu2026online} at each level $\alpha_j$, with no shared state between levels, and to apply Algorithm~\ref{alg:lazy-projection} on the joint vector. We refer to this composed algorithm as coordinate-wise \UPOCP and defer the complete pseudocode (Algorithm~\ref{alg:cw-upocp}) to Appendix~\ref{app:algorithms}.

The total regret of $d$ independent instances is the sum of their per-coordinate regrets~\citep[Sec.~10.3]{orabona2019modern}: $F_T(\bu) = \sum_{j=1}^d F_T^{(j)}(\abs{u_j})$, where $F_T^{(j)}$ is the linearized-regret bound of \UPOCP at level $\alpha_j$~\citep[Theorem~4.3]{liu2026online}. The Fenchel conjugate factorizes accordingly, $F_T^\star(\btheta) = \sum_{j=1}^d (F_T^{(j)})^\star(\abs{\theta_j})$, and Theorem~\ref{thm:level-set} becomes
\begin{equation}
\label{eq:cw-central}
  \sum_{j=1}^d \big(F_T^{(j)}\big)^\star\big(\abs{G_{T,j}}\big) \leq B_T.
\end{equation}
The Fenchel conjugate of \UPOCP is non-decreasing in $\abs{\theta_j}$ and $(F_T^{(j)})^\star(0) \geq - 1$, so dropping all but one term in~\eqref{eq:cw-central} yields a per-coordinate budget of $B_T + d - 1$.
Inverting the resulting Bernstein-type inequality~\citep[App.~F.1]{liu2026online} gives the following theorem.
\begin{theorem}[Per-level coverage of coordinate-wise \UPOCP]
\label{thm:cw-upocp-per-level}
Run Algorithm~\ref{alg:cw-upocp}. For every $j \in [d]$,
\begin{equation}
\label{eq:cw-per-level}
  \MisCov_{T,j} \leq \hat\varepsilon_{T,j} + \sqrt{2\alpha_j(1-\alpha_j)\hat\varepsilon_{T,j}},
  \quad
  \hat\varepsilon_{T,j} \coloneqq  \frac{1}{T}\left[\ln(B_T + d) + \tfrac{1}{2}\ln\big(\pi(T+1)\big)\right].
\end{equation}
In particular, with Assumption~\ref{ass:poly-growth}, $\max_{j \in [d]} \ \MisCov_{T, j} = \sO\left(\sqrt{((1 + \gamma) \log T + \log d) / T}\right)$.
\end{theorem}

The per-level rate matches the $p$-norm bound of Corollary~\ref{cor:p-norm} in its dependence on $d$, up to an additional $\log T$ factor that is the price of being parameter-free. Crucially, no quantity used by the algorithm requires knowledge of $D$, $\gamma$, or $T$. The factor $\alpha_j(1-\alpha_j)$ also guarantees a faster rate for extreme quantile levels.

\textbf{Extensions to continuous quantile levels.}
An important advantage of our framework is that it can be easily generalized from a finite collection $\sA = \{\alpha_1, \dots, \alpha_d\}$ to a compact interval $\sX \subseteq (0,1)$ by replacing $\R^d$ with the Hilbert space $\sH = L^2(\sX, \mu)$ for a finite Borel reference measure $\mu$ with full support, and replacing the isotonic cone by the cone of non-increasing functions. We collect the formal statements, including a coordinate-wise \UPOCP analogue obtained via Donsker--Varadhan change-of-measure, in Appendix~\ref{app:continuous}.%

\section{Sharpness and coverage improvement of isotonic projection}
\label{sec:projection-benefits}

Section~\ref{sec:reduction} used the lazy projection structurally, to make the played iterate feasible without inflating the coverage bound. We now show that it does strictly more: whenever the hidden iterate $\tilde{\bx}_t$ violates isotonicity, projecting it onto $\sK$ \emph{strictly} reduces a measure of prediction width while \emph{weakly} reducing a measure of miscoverage, so the empirical Pareto frontier of width versus coverage shifts monotonically toward the ideal corner of small width and accurate coverage.

We frame our analysis in terms of the Weighted Interval Score (WIS)~\citep{bracher2021evaluating}, the standard scoring rule for assessing multi-level interval forecasts. WIS is an extension of the classical Interval Score~\citep{winkler1972decision,gneiting2007strictly} to the multi-level setting.
For a collection of interval forecasts $[l_\alpha, u_\alpha]$, $\alpha \in \sA$, the WIS is defined as
\begin{equation} \label{eq:wis}
  \WIS_{\sA}\left(\set{[l_\alpha, u_\alpha]}_{\alpha \in \sA}, y\right)
  \coloneqq  \underbrace{\sum_{\alpha \in \sA} \alpha \left(u_\alpha - l_\alpha\right)}_\textit{sharpness reward} + \underbrace{\sum_{\alpha \in \sA} 2 \cdot \dist\left(y, [l_\alpha, u_\alpha]\right)}_\textit{coverage penalty},
\end{equation}
where $\dist(y, [l_{\alpha}, u_{\alpha}]) = \inf_{x \in [l_{\alpha}, u_{\alpha}]} \abs{y - x}$.
For symmetric prediction sets $\hat{C}_t(\hat{Y}_t, x) = [\hat{Y}_t - x, \hat{Y}_t + x]$ at miscoverage levels $\sA = \{\alpha_1, \dots, \alpha_d\}$, a direct calculation (Appendix~\ref{app:wis-pinball}) gives
\begin{equation} \label{eq:wis-pinball-identity}
  \sL_t(\bx) + S_t \sum_{i = 1}^d \alpha_i
  = \frac{1}{2}\WIS_{\sA}\left(\set{\hat{C}_t(\hat{Y}_t, x_i)}_{i}, Y_t\right)
  = \underbrace{\sum_{i = 1}^d \alpha_i x_i}_{\textit{sharpness reward}}
  + \underbrace{\sum_{i = 1}^d \max(S_t - x_i,0)}_{\textit{coverage penalty}}.
\end{equation}
This means that, up to an additive term that is constant in $\bx_t$, the aggregate pinball loss \emph{is} half the WIS of our prediction sets.
The sharpness term is an $\alpha$-weighted measure of total predicted width, with higher miscoverage levels (smaller intervals) weighted more; the coverage penalty $\sum_i \max(S_t - x_i,0)$ is a one-sided $\ell_1$ residual that penalizes under-coverage at every level.

Our main result in this section is that the projection step in Algorithm~\ref{alg:lazy-projection} reduces both terms of~\eqref{eq:wis-pinball-identity} at every round (and hence WIS itself), and strictly so whenever the hidden iterate is not already isotonic.
\begin{theorem}[Per-round reduction]
\label{thm:per-round-reduction}
For every round $t$, the hidden iterate $\bxtilde_t \in \R^d_+$ and the played iterate $\bx_t = \Pi_\sK(\bxtilde_t) \in \sK \cap \R^d_+$ satisfy $
  \sL_t(\bx_t) \leq \sL_t(\bxtilde_t)$.
Moreover, both the $\alpha$-weighted prediction sum and the total coverage penalty are individually reduced:
\begin{align}
  \sum_{i = 1}^d \alpha_i x_{t, i}
  \leq \sum_{i = 1}^d \alpha_i \tilde{x}_{t, i}\quad \text{ and } \quad
  \sum_{i = 1}^d \max(S_t - x_{t, i},0)
  \leq \sum_{i = 1}^d \max(S_t - \tilde{x}_{t, i},0).
\end{align}
When $\bxtilde_t \notin \sK$, the sharpness reduction and the loss reduction are strict.
\end{theorem}
The proof hinges on the characterization of Pool-Adjacent-Violators (PAVA) algorithm used for the isotonic projection~\citep{robertson1988order}.
The full proof, including the supporting lemmas, is in Appendix~\ref{app:projection-benefits}.
The inequalities hold for any $\bxtilde_t \in \R^d$; with valid hidden radii (Section~\ref{sec:reduction}), both sides are WIS components of genuine prediction intervals, since PAVA averages non-negative radii.

\citet{ding2025calibrated} obtained a decomposition analogous to \eqref{eq:wis-pinball-identity} under the assumption that $\sA$ is symmetric about $1/2$, but they only used it as a motivation to minimize the regret with $\sL_t$. In fact, they observed a better empirical performance of \MultiQT compared to the \emph{naive forecasters}, but they did not provide any explanation for it.
Our Theorem~\ref{thm:per-round-reduction} provides this missing explanation.

\section{Experiments}
\label{sec:experiments}

Here, we evaluate our framework on both synthetic and real-world datasets.

\paragraph{Datasets.}
We follow the OCP literature~\citep{gibbs2021adaptive,angelopoulos2023conformal,gibbs2024conformal,podkopaev2024adaptive} and use three categories of streams: (1) daily opening prices of major US stocks (AAPL, AMZN, GOOGL) from 2006 to 2017~\citep{nguyen2018s}; (2) electricity demand records from New South Wales~\citep{harries1999splice}; and (3) synthetic environments following~\citet[App.~F.5]{angelopoulos2023conformal} that bypass forecasting and directly simulate the nonconformity scores $S_t$ under three patterns (sinusoidal with Gaussian noise; constant and quadratic trends with sparse bumps), allowing repeated runs for error bars. Forecasting pipelines (daily-refit Prophet for stocks~\citep{taylor2018forecasting}, AR(3) for electricity) and full synthetic-stream specifications are in Appendix~\ref{app:experiments-datasets}.

\paragraph{Methods.}
We compare four method families: \UPOCP (Algorithm~\ref{alg:cw-upocp}, parameter-free), \PNormOMD (Algorithm~\ref{alg:pnorm-omd}, with $p = 2 \log d / (2 \log d - 1)$ from Corollary~\ref{cor:p-norm} and a constant step size chosen per dataset by oracle grid search on a base-$2$ logarithmic grid), \KT (Algorithm~\ref{alg:cw-kt}, parameter-free reference), and the OGD-based prior baseline \MultiQT~\citep{ding2025calibrated} (adaptive; uses a data-dependent learning-rate schedule). The first three are run both as vanilla learners and wrapped with the lazy projection of Algorithm~\ref{alg:lazy-projection} (suffix~\textsc{-Iso}); \MultiQT is by construction projected. Update rules, hyperparameter grids, and selected $\eta$ values are in Appendix~\ref{app:algorithms}.

\textbf{Metrics.}
We track, per method, (1) the \emph{empirical coverage} $\Cov_{T,j}$ as a function of the target $1-\alpha$; (2) the per-round \emph{prediction-set width} $|\hat C_t(\hat Y_t, x_{t,j})|$, summarized by its mean (the standard sharpness measure), median, and $95$th percentile; (3) the \emph{full distribution} of widths at selected levels; and (4) the \emph{WIS}~\eqref{eq:wis}, summed over all $d$ levels and averaged over rounds, with its two components.

The scores are the absolute residuals $S_t = |Y_t - \hat{Y}_t|$; none is exactly zero (the smallest is about $10^{-5}$), so the empty-set convention never changes a coverage indicator. 
In all cases we calibrate $d = 100$ levels uniformly spaced in $[0.001, 0.5]$ for each method, with warm-up periods of $T_{\text{burn-in}} = 100$ rounds ($300$ for \texttt{elec2}) during which the predictor outputs sets but is not updated. 
Results are reported on the practically interesting window $\alpha \in [0.03, 0.25]$ (target coverage $1 - \alpha \in [0.75, 0.97]$).
We focus on the AAPL dataset and the synthetic sinusoidal stream; full results for the other datasets are in Appendix~\ref{app:experiments-extended}.

\begin{figure}[!t]
  \centering
  \includegraphics[width=\textwidth]{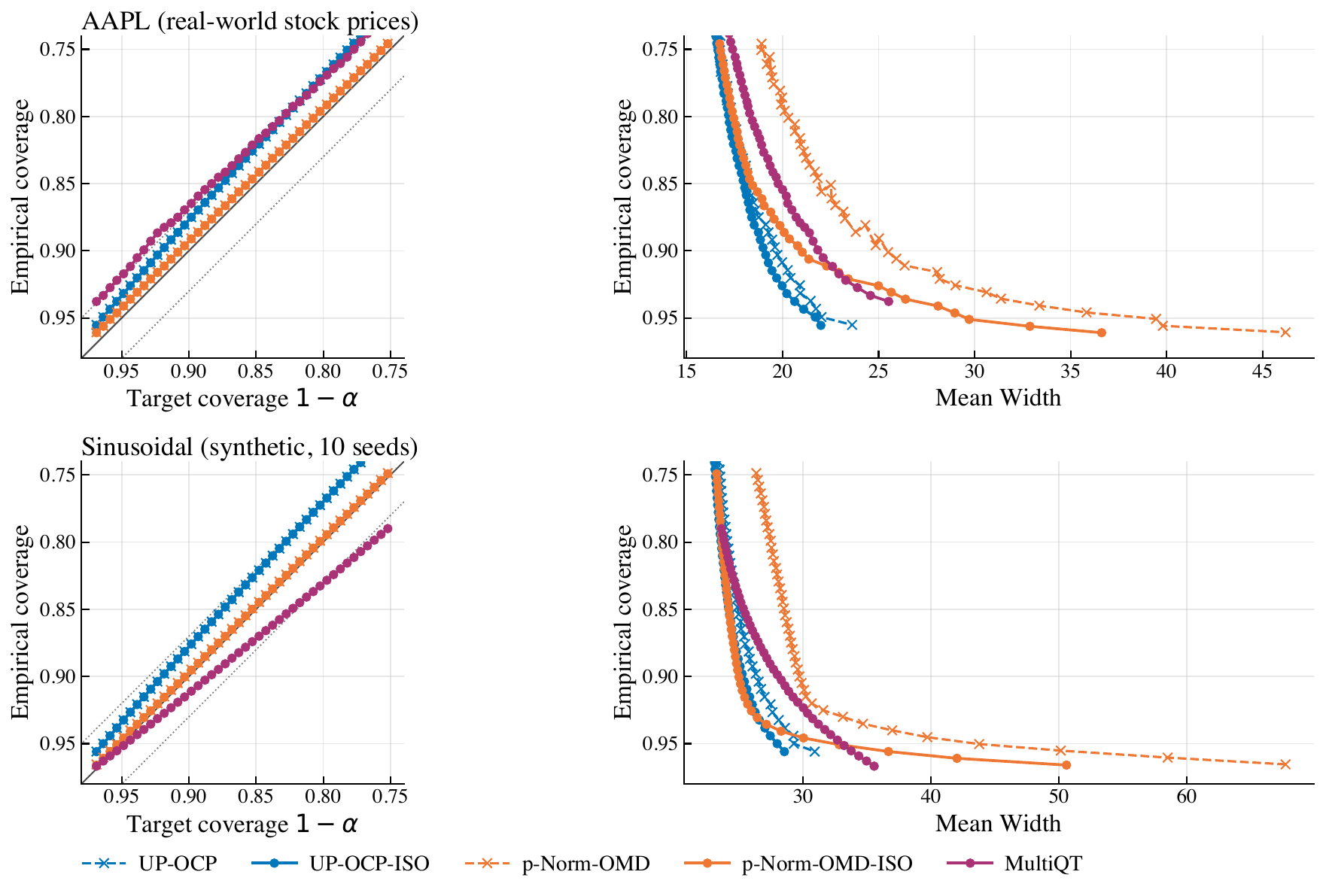}
  \caption{Calibration (left column: empirical versus target coverage $1 - \alpha$) and Pareto frontiers (right column: empirical coverage versus mean prediction-set width), for AAPL (top row) and the sinusoidal dataset (bottom row, mean across $10$ seeds). The solid diagonal implies perfect calibration $y = x$, the dotted lines are the $\pm 0.03$ error band. Better performance is closer to the bottom-left corner.}
  \label{fig:paper-main}
\end{figure}

\begin{figure}[t]
  \centering
  \includegraphics[width=\textwidth]{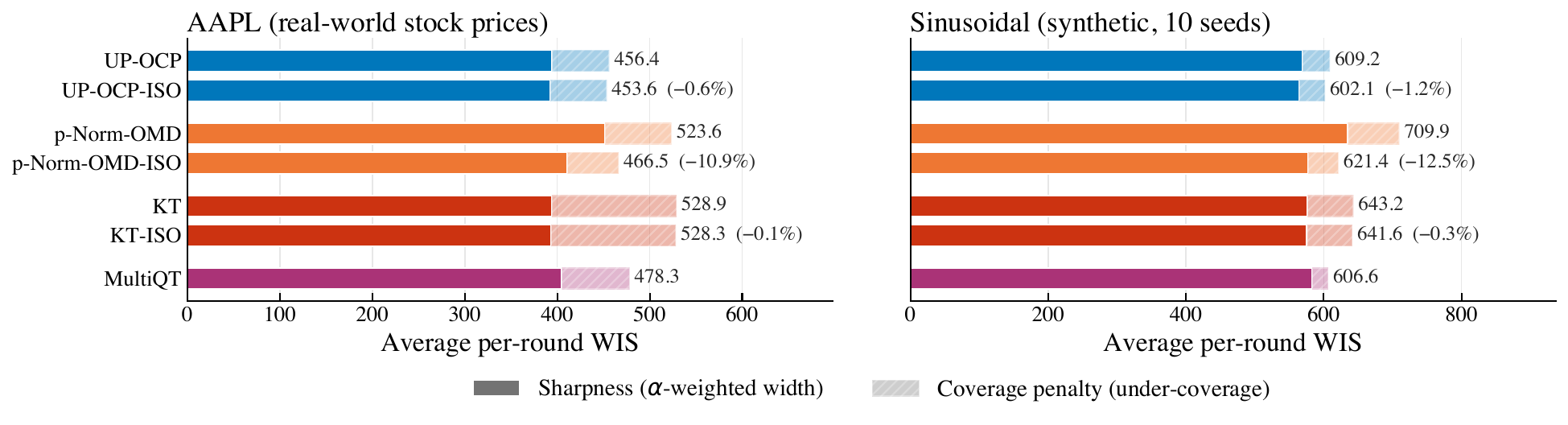}
  \caption{Average per-round WIS over all $d = 100$ levels (post burn-in; sinusoidal: mean over $10$ seeds), split into the sharpness (solid) and coverage-penalty (hatched) components of~\eqref{eq:wis}. Percentages give the change of each projected variant relative to its unprojected counterpart.}
  \label{fig:paper-wis}
\end{figure}

\textbf{Calibration and Pareto frontiers.}
Higher coverage requires larger sets, but this trade-off varies across algorithms. Following~\citet{liu2026online}, we assess it through a \emph{Pareto frontier}: Figure~\ref{fig:paper-main} shows, per dataset, a calibration curve (left) and the empirical coverage against the mean width (right) for $\alpha \in [0.03, 0.25]$. We prioritize the mean width, which reflects the cumulative cost of uncertainty downstream and, unlike quantiles of the width, captures unreasonably large sets. We exclude \KT, whose width is substantially larger at almost all levels and would compress every other curve; it is included in Figure~\ref{fig:paper-wis} and Appendix~\ref{app:experiments-extended}. Several observations follow.

\emph{Calibration is preserved by the projection.} In each calibration panel, the projected (\Iso) and unprojected curves are visually indistinguishable; the $p$-norm curves lie inside the $\pm 0.03$ tolerance band over the entire $\alpha$ window, and the \UPOCP curves leave it only slightly, at the lowest target coverages. This is consistent with Proposition~\ref{prop:regret-preservation}: the projection does not consume any regret, so it does not consume any coverage.

\emph{Sharpness is strictly improved by the projection, for almost all $\alpha$}. 
On the right-hand panels, every solid (\textsc{-Iso}) curve sits on or to the left of its dashed (unprojected) counterpart, i.e., it is narrower at the same coverage, with the gap widening at small $\alpha$ (high coverage), where the hidden iterate is most likely to violate isotonicity.
The reduction is largest for \PNormOMD, whose unprojected mean width is up to $33\%$ (AAPL) and $39\%$ (sinusoidal) larger than the projected one near $1 - \alpha = 0.97$; for \UPOCP it is about $7$--$8\%$. 

\emph{Coordinate-wise \UPOCP uniformly improves over \PNormOMD and \MultiQT in width.} On AAPL, \UPOCP-\textsc{Iso} is narrower than both \PNormOMD variants and \MultiQT at every $\alpha$. On AAPL, tuned $p$-norm OMD with the optimal $p$ achieves the highest empirical coverage but pays for it with much wider sets. On both datasets, the constant step size selected from our grid by Pareto dominance is $\eta = 64$. Oracle tuning gives \PNormOMD an advantage that is unavailable in practice, while \UPOCP and \KT are parameter-free; still, the results are not fragile to it: every $\eta \in [16, 128]$ stays within about $10\%$ of the best WIS on both datasets, and a plug-in step size using only revealed scores comes within $4\%$ (Appendix~\ref{app:eta-sensitivity}).

\begin{remark}[Slight miscoverage of the parameter-free and adaptive instances]
  There is a fundamental trade-off between marginal coverage and prediction set size in OCP, as shown in recent lower bounds arguments~\citep{srinivas2026online}: any algorithm whose prediction sets have size at most $\mu$ times the optimal must incur $\Omega\left(\log(1/\varepsilon_{\min})/\log \mu \cdot\alpha T\right)$ coverage errors, where $\varepsilon_{\min}$ is the smallest relevant scale. As suggested by~\citet{liu2026online}, empirically, this can be fixed by a simple $\alpha$-correction technique.
\end{remark}

Tables in Appendix~\ref{app:experiments-results} report the same comparison at three representative target-coverage levels $1 - \alpha \in \set{0.95, 0.9, 0.8}$ for every dataset, including \KT and the median and $95$th-percentile widths (Table~\ref{tab:AAPL_summary} for AAPL).

\textbf{Weighted Interval Score.} The projection lowers the total WIS (Figure~\ref{fig:paper-wis}) of every base learner on both datasets, most for \PNormOMD ($-10.9\%$ on AAPL, $-12.5\%$ on the sinusoidal stream), and \UPOCP-\textsc{Iso} attains the smallest WIS on both. Since Figure~\ref{fig:paper-wis} compares runs with different trajectories, we also checked Theorem~\ref{thm:per-round-reduction} in its exact within-run form: the hidden iterate violates isotonicity in $99.6\%$ of the rounds, and in every one of them the projection strictly reduced the WIS and its sharpness component without increasing the coverage penalty (Appendix~\ref{app:wis-results}, which also reports the WIS on every dataset).

\textbf{Further results.} The projection also removes the heavy right tail of the per-round width distribution (Appendix~\ref{app:width-distributions}). As $d$ grows from $5$ to $1{,}000$ on the sinusoidal stream, the realized worst-level errors grow at most slowly, so the worst-case $\sqrt{d}$ and $\sqrt{\log d}$ scalings are not visible there; the methods designed through the tighter analysis are nonetheless more accurate at every $d$ (Appendix~\ref{app:d-ablation}).

\section{Conclusion, limitations, and future work}
\label{sec:conclusion}

We presented a black-box reduction that turns any online learner with a linearized regret bound into a multi-level OCP algorithm with \emph{multi-level coverage}, by wrapping the learner in a lazy isotonic projection layer. The reduction preserves regret (and therefore coverage) while strictly improving sharpness, formalized through the Weighted Interval Score. Instantiating the framework with $p$-norm OMD and coordinate-wise Universal Portfolio replaces the $\sqrt{d}$ dependence in the per-level miscoverage rate of \MultiQT with $\sqrt{\log d}$, with the coordinate-wise Universal Portfolio instantiation being parameter-free. Extensive experiments confirm that the projection wrapper is a free (in terms of order of computational complexity) per-round improvement on every base learner we tested, and that \UPOCP delivers smaller and better-behaved prediction sets than the tuned baselines.

\textbf{Limitations and future work.} Our guarantees concern long-run \emph{marginal} coverage, the standard target in adversarial OCP. Coverage conditional on groups (attainable at a single level~\citep{bharti2026parameter}) or on time intervals remains open in the multi-level setting and would likely require multi-validity~\citep{bastani2022practical} or strongly adaptive online learning~\citep{bhatnagar2023improved}.

\section*{Acknowledgements}
ED's work was supported in part by the US NSF, ARO, ONR, and the Sloan Foundation.

\bibliographystyle{plainnat}
\bibliography{references}

\newpage
\appendix
\addtocontents{toc}{\protect\setcounter{tocdepth}{3}}

{\hypersetup{linkcolor=black}
\setcounter{tocdepth}{3}
\renewcommand{\contentsname}{Appendix}
\tableofcontents
}

\section{Deferred Proofs and supporting material for Section~\ref{sec:reduction}}
\label{app:section3}

This appendix collects the proofs and auxiliary results deferred from Section~\ref{sec:reduction}. We restate each statement before its proof. Throughout, we write
\[
  \bdelta_T \;\coloneqq \; \frac{\bG_T}{T} \;\in\; \R^d
\]
for the time-averaged cumulative subgradient, so that $\delta_{T,i} = G_{T,i}/T$ is the per-level coverage gap and~\eqref{eq:miscov-subgrad} reads $\MisCov_{T,i} = \abs{\delta_{T,i}}$.

\subsection{Projection inequality (Lemma~\ref{lem:projection-inequality})}
\label{app:projection-inequality}

We first record an immediate consequence of inward flow, then prove Lemma~\ref{lem:projection-inequality}. The argument is in $\R^d$, but every step extends verbatim to a Hilbert space, which is exploited in Appendix~\ref{app:continuous}. Recall the \emph{normal cone} of a closed convex set $C$ at $\bx \in C$, $N_C(\bx) \coloneqq  \{\bv \in \R^d : \inn{\bv}{\by - \bx} \leq 0 \text{ for all } \by \in C\}$, is the polar of the tangent cone $T_C(\bx)$, i.e., $\inn{\bv}{\bw} \leq 0$ for every $\bv \in N_C(\bx)$ and $\bw \in T_C(\bx)$~\citep[][Sec.~6]{rockafellar1997convex}.

\begin{lemma}
\label{lem:normal-cone-nonneg}
Let $\ell$ with subgradient $\bg$ and a closed convex set $C \subseteq \R^d$ satisfy inward flow. For every $\bx \in \boundary(C)$ and $\bv \in N_C(\bx)$,
\(
  \inn{\bv}{\bg(\bx)} \geq 0.
\)
\end{lemma}

\begin{proof}
By inward flow, $-\bg(\bx) \in T_C(\bx)$. By tangent--normal cone polarity, $\inn{\bv}{-\bg(\bx)} \leq 0$ for every $\bv \in N_C(\bx)$, i.e., $\inn{\bv}{\bg(\bx)} \geq 0$.
\end{proof}

\begin{proof}[Proof of Lemma~\ref{lem:projection-inequality}]
If $\bxtilde \in C$, then $\Pi_C(\bxtilde) = \bxtilde$ and the inequality is an equality. Otherwise, let $\bx_0 = \Pi_C(\bxtilde)$. By the variational characterization of Euclidean projection, there exists $\bv \in N_C(\bx_0)$ such that $\bxtilde = \bx_0 + \bv$. Then
\[
  \inn{\bg(\bx_0)}{\bxtilde} = \inn{\bg(\bx_0)}{\bx_0} + \inn{\bg(\bx_0)}{\bv}
  \geq \inn{\bg(\bx_0)}{\bx_0},
\]
where the inequality is Lemma~\ref{lem:normal-cone-nonneg} with $\bx = \bx_0 \in \boundary(C)$.
\end{proof}

\subsection{Regret preservation and the central inequality}
\label{app:central-proof}

\begin{proof}[Proof of Proposition~\ref{prop:regret-preservation}]
By convexity of $\sL_t$ and $\bg_t \in \partial \sL_t(\bx_t)$, $\sL_t(\bx_t) - \sL_t(\bu) \leq \inn{\bg_t}{\bx_t - \bu}$. Summing over $t$ gives $\Regret_T(\bu) \leq \LinRegret_T(\bu)$, so the bound~\eqref{eq:regret-preservation} proved below also bounds the regret. By Lemma~\ref{lem:projection-inequality} applied to $C = \sK$, $\inn{\bg_t}{\bx_t} \leq \inn{\bg_t}{\bxtilde_t}$ for every $t$, so
\[
  \sum_{t=1}^T \inn{\bg_t}{\bx_t - \bu}
  \leq \sum_{t=1}^T \inn{\bg_t}{\bxtilde_t - \bu}
  \leq F_T(\bu),
\]
where the last step is the regret bound of $\sE$. \qedhere
\end{proof}

\begin{proof}[Proof of Theorem~\ref{thm:level-set}]
By regret-reward duality~\citep[Sec.~6.4]{orabona2019modern}, the regret bound of $\sE$ is equivalent to
\(
  F_T^\star\!\left(-\sum_{t=1}^T \bg_t\right) \leq -\sum_{t=1}^T \inn{\bg_t}{\bxtilde_t}.
\)
Inward flow of $(\sL_t,\sK)$ together with Lemma~\ref{lem:projection-inequality} gives $\inn{\bg_t}{\bx_t} \leq \inn{\bg_t}{\bxtilde_t}$, hence $-\inn{\bg_t}{\bxtilde_t} \leq -\inn{\bg_t}{\bx_t}$. Convexity of $\sL_t$ at $\bx_t$ with comparator $\bzero$ yields
\[
  \sL_t(\bzero) \geq \sL_t(\bx_t) + \inn{\bg_t}{\bzero - \bx_t}
  \quad\Longrightarrow\quad
  -\inn{\bg_t}{\bx_t} \leq \sL_t(\bzero) - \sL_t(\bx_t) \leq \sL_t(\bzero),
\]
where we used $\sL_t(\bx_t) \geq 0$ since each pinball loss is non-negative. Summing over $t$ gives
\[
  F_T^\star\!\left(-\bG_T\right)
  \leq -\sum_{t=1}^T \inn{\bg_t}{\bxtilde_t}
  \leq \sum_{t=1}^T \sL_t(\bzero)
  = B_T. \qedhere
\]
\end{proof}

\subsection{Inward flow for the isotonic cone}
\label{app:isotonic-inward-flow}

\begin{proof}[Verification of inward flow for the isotonic cone]
Fix $t$ and $\bx \in \boundary(\sK)$. Let $I(\bx) \coloneqq  \{i \in [d-1] : x_i = x_{i+1}\}$ be the set of tight constraints; $I(\bx)$ is non-empty since $\bx$ is on the boundary. Because $\sK$ is polyhedral with constraints $x_i - x_{i+1} \geq 0$, the tangent cone at $\bx$ is
\[
  T_\sK(\bx) = \set{\bv \in \R^d : v_i - v_{i+1} \geq 0 \text{ for every } i \in I(\bx)}.
\]
Inward flow asks $-\bg_t \in T_\sK(\bx)$, equivalently $g_{t,i} \leq g_{t,i+1}$ for every $i \in I(\bx)$. With the subgradient~\eqref{eq:pinball-subgradient}, we have $g_{t,i} = \bone\{x_i \geq S_t\} - (1-\alpha_i)$, where the coverage indicator follows the empty-set convention of Section~\ref{sec:setup}; in particular, it depends on $x_i$ only through its value, so tied coordinates share it. Take $i \in I(\bx)$ so that $x_i = x_{i+1}$.

\emph{Case 1: both levels cover, i.e., $x_i = x_{i+1} \geq S_t$ and $x_i > 0$.} Then $g_{t,i} = \alpha_i$ and $g_{t,i+1} = \alpha_{i+1}$. Since $\alpha_i < \alpha_{i+1}$, $g_{t,i} < g_{t,i+1}$.

\emph{Case 2: neither level covers, i.e., $x_i = x_{i+1} < S_t$, or $x_i = x_{i+1} = 0$ (the empty set).} Then $g_{t,i} = -(1-\alpha_i)$ and $g_{t,i+1} = -(1-\alpha_{i+1})$. Since $\alpha_i < \alpha_{i+1}$, $1-\alpha_i > 1-\alpha_{i+1}$, hence $g_{t,i} < g_{t,i+1}$.

In either case $g_{t,i} \leq g_{t,i+1}$, completing the proof.
\end{proof}

\subsection{Detailed derivation for Corollary~\ref{cor:p-norm}: $p$-norm OMD}
\label{app:p-norm}

The regret bound for $p$-norm OMD with regularizer $\psi(\bx) = (1/2) \norm{\bx}_p^2$ at constant step size $\eta > 0$ is standard~\citep[Theorem~6.8]{orabona2019modern}: $\psi$ is $(p - 1)$-strongly convex with respect to $\norm{\cdot}_p$ by~\citep[Lemma~6.36]{shalev2007thesis}, yielding
\[
    F_T(\bu)
    = \frac{\norm{\bu}_p^2}{2 \eta}
    + \frac{\eta}{2 (p - 1)} \sum_{t = 1}^T \norm{\bg_t}_q^2.
\]
The Fenchel conjugate of $a \norm{\cdot}_p^2$ is $(1 / (4 a)) \norm{\cdot}_q^2$. Applying this with $a = 1 / (2 \eta)$ and absorbing the additive $T$-dependent term gives
\[
    F_T^\star(\btheta)
    = \frac{\eta}{2} \norm{\btheta}_q^2
    - \frac{\eta}{2 (p - 1)} \sum_{t = 1}^T \norm{\bg_t}_q^2.
\]
The central inequality~\eqref{eq:central-inequality} then becomes
\(
    (\eta / 2) \norm{\bG_T}_q^2 - (\eta / (2 (p - 1))) \sum_{t = 1}^T \norm{\bg_t}_q^2 \leq B_T,
\)
i.e.,
\[
    \norm{\bG_T}_q^2
    \leq \frac{2 B_T}{\eta} + \frac{1}{p - 1} \sum_{t = 1}^T \norm{\bg_t}_q^2.
\]
Since $\norm{\bg_t}_\infty \leq 1$ and $\norm{\bg_t}_q \leq d^{1/q} \norm{\bg_t}_\infty \leq d^{1/q}$ by H\"older,
\(
    \sum_{t = 1}^T \norm{\bg_t}_q^2 \leq d^{2/q} \, T.
\)

For the optimization, set $q = 2 \log d$. Then $p - 1 = 1/(q - 1) = 1/(2 \log d - 1)$ and $d^{2/q} = d^{1/\log d} = e$. Plugging in,
\[
    \norm{\bG_T}_q^2 \leq \frac{2 B_T}{\eta} + e (2 \log d - 1) T.
\]
Setting $\eta \coloneqq  2 B_T / (e (2 \log d - 1) T)$ balances the two terms and gives $\norm{\bG_T}_q^2 \leq 2 e (2 \log d - 1) T$. This choice of $\eta$ is well-defined whenever $B_T > 0$; the trivial case $B_T = 0$ corresponds to $S_t = 0$ for all $t$, where~\eqref{eq:lq-coverage} holds for every $\eta > 0$ and already gives $\norm{\bG_T}_q^2 \leq e (2 \log d - 1) T$. Under Assumption~\ref{ass:poly-growth}, the upper bound $B_T \leq d D T^{1 + \gamma}$ gives $\eta \leq 2 d D T^\gamma / (e (2 \log d - 1)) = O(d D T^\gamma / \log d)$ as a function of $T$.

The per-level miscoverage bound follows from $\max_{i \in [d]} \abs{\delta_{T, i}} = \max_i \abs{G_{T, i}} / T \leq \norm{\bG_T}_q / T$, since $\norm{\cdot}_\infty \leq \norm{\cdot}_q$.

\paragraph{On the choice of $\eta$.}
The right-hand side of~\eqref{eq:lq-coverage} is strictly decreasing in $\eta$, with infimum $\frac{1}{T}\sqrt{d^{2/q} T / (p - 1)}$ as $\eta \to \infty$; the balanced $\eta$ above attains it up to a factor $\sqrt{2}$, at a finite step size. Small miscoverage alone is a weak target: the data-free predictor that outputs $\R$ on the first $\lceil (1 - \alpha_i) T \rceil$ rounds and $\emptyset$ afterwards attains $\MisCov_{T, i} \leq 1/T$ (see~\citep[App.~H.5]{liu2026online} for a horizon-free variant). The regret bound is what controls efficiency; for instance, when the scores are i.i.d.\ (under regularity conditions), sublinear regret drives the average radius to the oracle quantile~\citep[Thm.~D.3]{liu2026online}. Finally, there is no contradiction with Proposition~\ref{prop:regret-monotonicity}: for $\eta' > \eta$, the two OMD bounds satisfy $F_T^{\eta'}(\bu) > F_T^{\eta}(\bu)$ if and only if $\norm{\bu}_p^2 < \eta \eta' \sum_t \norm{\bg_t}_q^2 / (p - 1)$, so neither dominates the other uniformly and the proposition does not apply.

\subsection{Beyond the isotonic cone}
\label{app:beyond-isotonic}

The lazy-projection framework rests on a single geometric assumption---inward flow---and is therefore agnostic to the details of $\sK$. We record here a composition principle that allows additional polyhedral constraints to be layered on top of the isotonic cone without redoing the analysis.

\begin{proposition}[Inward flow under intersection]
\label{prop:inward-flow-intersection}
Let $C_1, C_2 \subseteq \R^d$ be closed convex sets and $\ell$ a loss with subgradient selection $\bg(\cdot)$. Suppose that
\begin{enumerate}[nosep]
\item[(i)] $(\ell, C_1)$ and $(\ell, C_2)$ both satisfy inward flow with the same selection $\bg$;
\item[(ii)] for every $\bx \in \boundary(C_1 \cap C_2)$, $T_{C_1 \cap C_2}(\bx) = T_{C_1}(\bx) \cap T_{C_2}(\bx)$.
\end{enumerate}
Then $(\ell, C_1 \cap C_2)$ satisfies inward flow with $\bg$. Condition~(ii) holds whenever $C_1$ and $C_2$ are polyhedral.
\end{proposition}

\begin{proof}
Fix $\bx \in \boundary(C_1 \cap C_2)$. For $i \in \{1,2\}$, either $\bx \in \boundary(C_i)$, in which case inward flow gives $-\bg(\bx) \in T_{C_i}(\bx)$, or $\bx \in \mathrm{int}(C_i)$, in which case $T_{C_i}(\bx) = \R^d$. Either way, $-\bg(\bx) \in T_{C_1}(\bx) \cap T_{C_2}(\bx) = T_{C_1 \cap C_2}(\bx)$ by~(ii).
\end{proof}

\paragraph{Example: bounded isotonic cone.}
Suppose Assumption~\ref{ass:poly-growth} holds with $\gamma = 0$, so $S_t \leq D$ for all $t$, and let $R \geq D$. The set
\(
  \sK_R \coloneqq  \sK \cap \set{\bx \in \R^d : x_j \leq R \text{ for all } j}
\)
enforces both isotonicity and an upper bound. To verify inward flow on $\sK_R$, by Proposition~\ref{prop:inward-flow-intersection} it suffices to verify it on the upper-bound polyhedron $C_R \coloneqq  \{\bx : x_j \leq R\}$. At a boundary point with $x_j = R \geq D \geq S_t$, the subgradient coordinate is $g_{t,j} = \alpha_j > 0$, so $-g_{t,j} < 0$, which lies in the half-space $v_j \leq 0$ defining the tangent cone of $C_R$ along $x_j = R$. Both $\sK$ and $C_R$ are polyhedral, so~(ii) holds, and Proposition~\ref{prop:inward-flow-intersection} applies. All bounds in Section~\ref{sec:reduction} carry over with $\sK$ replaced by $\sK_R$, with the additional benefit that the played radii are clipped at the largest plausible score.

\subsection{Tighter regret implies tighter coverage}
\label{app:regret-coverage-monotonicity}

The following monotonicity formalizes the second corollary of Theorem~\ref{thm:level-set}; it generalizes the scalar argument of \citet[Sec.~4]{liu2026online} to the vector setting and underlies the design of the case studies in Sections~\ref{sub:pnorm}--\ref{sub:upocp}.

\begin{proposition}
\label{prop:regret-monotonicity}
Let $F_T^A, F_T^B \colon \R^d \to \R$ be two valid linearized regret bounds for two base learners $\sE_A, \sE_B$ that satisfy $F_T^A(\bu) \leq F_T^B(\bu)$ for all $\bu \in \R^d$. Then their Fenchel conjugates satisfy $(F_T^A)^\star(\bz) \geq (F_T^B)^\star(\bz)$ for all $\bz \in \R^d$, and the coverage-error level sets~\eqref{eq:level-set} are nested:
\[
  \sZ_T(F_T^A) \;\subseteq\; \sZ_T(F_T^B).
\]
Consequently, if $\sE_A$ admits a uniformly tighter regret bound than $\sE_B$, the worst-case miscoverage region delivered by Theorem~\ref{thm:level-set} for $\sE_A$ is contained in that for $\sE_B$.
\end{proposition}

\begin{proof}
Fix $\bz \in \R^d$. By definition of the Fenchel conjugate,
\[
  (F_T^A)^\star(\bz)
  = \sup_{\bu \in \R^d} \big[\inn{\bz}{\bu} - F_T^A(\bu)\big]
  \geq \sup_{\bu \in \R^d} \big[\inn{\bz}{\bu} - F_T^B(\bu)\big]
  = (F_T^B)^\star(\bz),
\]
using $-F_T^A(\bu) \geq -F_T^B(\bu)$ pointwise. For every $\bz \in \sZ_T(F_T^A)$,
\(
  (F_T^B)^\star(-\bz) \leq (F_T^A)^\star(-\bz) \leq B_T,
\)
so $\bz \in \sZ_T(F_T^B)$. \qedhere
\end{proof}

This makes precise the operational principle invoked in the main text: minimizing the linearized regret of the base learner directly tightens the coverage-error region of the wrapped algorithm.

\subsection{Coordinate-wise UP-OCP: per-level and mean-absolute proofs}
\label{app:cw-l1}

We first record the analytic facts about the Bernstein-type function used by UP-OCP, then prove Theorem~\ref{thm:cw-upocp-per-level} and Theorem~\ref{thm:cw-upocp-l1}. Throughout, the Fenchel conjugate of the UP-OCP regret bound at level $\alpha$ has the form $(F_T^{(\alpha)})^\star(\abs{\theta}) = \exp(h_\alpha(\theta)) - 1$ with
\[
  h_{\alpha}(\theta) \coloneqq  \frac{\theta^2}{2\alpha(1-\alpha)T + (2/3)\theta} - \frac{1}{2}\log(\pi(T+1));
\]
see~\citep[Theorem~4.3 and App.~E.4]{liu2026online}.

\begin{lemma}[Bernstein inversion]
\label{lem:bernstein-inversion}
Let $\sigma, \theta \geq 0$ and $c > 0$. If
\(
  \theta^2/\big(2\sigma T + (2/3)\theta\big) \leq c,
\)
then $\theta \leq \sqrt{2\sigma T c} + 2c/3$.
\end{lemma}

\begin{proof}
Rearranging gives $\theta^2 - (2c/3)\theta - 2\sigma T c \leq 0$. The quadratic formula yields
\(
  \theta \leq c/3 + \sqrt{c^2/9 + 2\sigma T c} \leq c/3 + c/3 + \sqrt{2\sigma T c} = 2c/3 + \sqrt{2\sigma T c}.
\)
\end{proof}

\begin{lemma}
\label{lem:bernstein-convex}
For every fixed $\sigma > 0$ and $T > 0$, the map $\theta \mapsto \theta^2/(2\sigma T + (2/3)\theta)$ is convex and non-decreasing on $[0,\infty)$.
\end{lemma}

\begin{proof}
Write $f(\theta) = \theta^2/(a + b\theta)$ with $a = 2\sigma T$, $b = 2/3$. Direct calculation gives $f'(\theta) = \theta(2a + b\theta)/(a+b\theta)^2 \geq 0$ and $f''(\theta) = 2a^2/(a+b\theta)^3 > 0$.
\end{proof}

\begin{proof}[Proof of Theorem~\ref{thm:cw-upocp-per-level}]
Substituting the form of $(F_T^{(j)})^\star$ into~\eqref{eq:cw-central},
\(
  \sum_{j=1}^d \exp(h_{\alpha_j}(\abs{G_{T,j}})) \leq B_T + d.
\)
Each term is non-negative, so we may drop all but the $j$-th: $\exp(h_{\alpha_j}(\abs{G_{T,j}})) \leq B_T + d$, equivalently
\[
  \frac{\abs{G_{T,j}}^2}{2\alpha_j(1-\alpha_j)T + (2/3)\abs{G_{T,j}}}
  \leq \log(B_T + d) + \tfrac{1}{2}\log(\pi(T+1))
  = T\hat\varepsilon_{T,j}.
\]
Lemma~\ref{lem:bernstein-inversion} with $\sigma = \alpha_j(1-\alpha_j)$ and $c = T\hat\varepsilon_{T,j}$ gives
\(
  \abs{G_{T,j}} \leq T\sqrt{2\alpha_j(1-\alpha_j)\hat\varepsilon_{T,j}} + (2/3)T\hat\varepsilon_{T,j}.
\)
Dividing by $T$ and using $2/3 < 1$ yields~\eqref{eq:cw-per-level}. The asymptotic claim uses $B_T \leq dDT^{1+\gamma}$ to give $\hat\varepsilon_{T,j} = \sO((\log d + \log T)/T)$ and hence $\max_j \abs{\delta_{T,j}} = \sO(\sqrt{(\log d + \log T)/T})$.
\end{proof}

The per-level guarantee~\eqref{eq:cw-per-level} can be aggregated into a mean-absolute coverage bound across levels. By dropping the variance budgets at individual levels in favor of their envelope $\sigma_\sA$ and applying Jensen's inequality to the resulting sum, one obtains a single $\ell_1$-style bound whose rate depends on the level set only through $\sigma_\sA$ and $\log d$. The bound below is a vector analogue of \citet[Theorem~4.4]{liu2026online} and, by norm equivalence, improves any $\ell_2$ bound by a factor of $1/\sqrt{d}$ on the same quantity (cf.\ Remark~\ref{rmk:up-vs-multiqt} below).

\begin{theorem}[$\ell_1$ miscoverage error of coordinate-wise \UPOCP]
\label{thm:cw-upocp-l1}
Run Algorithm~\ref{alg:cw-upocp}, and let $\sigma_\sA \coloneqq  \max_{i \in [d]} \alpha_i(1-\alpha_i) \in [0, 1/4]$ and $C_T \coloneqq  B_T + d$. Then
\begin{equation} \label{eq:cw-upocp-l1}
  \frac{1}{d} \sum_{i = 1}^d \abs{\delta_{T, i}}
  \;\leq\; \sqrt{\frac{2 \sigma_\sA \Lambda_T}{T}} + \frac{2 \Lambda_T}{3 T},
  \qquad
  \Lambda_T \;\coloneqq \; \log\frac{C_T}{d} + \frac{1}{2} \log\!\big(\pi (T + 1)\big).
\end{equation}
Under Assumption~\ref{ass:poly-growth}, $\Lambda_T = \sO((1 + \gamma) \log T)$, hence
\(
  \frac{1}{d} \sum_{i = 1}^d \abs{\delta_{T, i}}
  = \sO\!\big(\sqrt{\sigma_\sA (1 + \gamma) \log T / T}\big).
\)
\end{theorem}

\begin{remark}[Comparison with the $\ell_2$-bound of \citet{ding2025calibrated}]
\label{rmk:up-vs-multiqt}
By norm equivalence, $\frac{1}{d}\norm{\bdelta_T}_1 \leq \frac{1}{\sqrt{d}}\norm{\bdelta_T}_2$, so~\eqref{eq:cw-upocp-l1} delivers a $1/\sqrt{d}$ improvement over any $\ell_2$ bound on the mean-absolute miscoverage. Combined with Theorem~\ref{thm:cw-upocp-per-level}, this improves the dependence on $d$ over the $\ell_2$ guarantee of \MultiQT, at the price of a $\log T$ factor; it is not uniformly tighter in $T$, since for a fixed step size the bounds of \MultiQT carry no $\log T$ factor and, for a common point forecast, decay as $\sO(1/T)$ (Appendix~\ref{app:prior-comparison}).
\end{remark}

\begin{proof}[Proof of Theorem~\ref{thm:cw-upocp-l1}]
Substituting the form of $(F_T^{(j)})^\star$ into~\eqref{eq:cw-central},
\[
  \sum_{j=1}^d \exp\!\big(h_{\alpha_j}(\abs{G_{T,j}})\big)
  \leq B_T + d = C_T.
\]
By definition of $\sigma_\sA$, $\alpha_j(1-\alpha_j) \leq \sigma_\sA$ for every $j$, so $h_{\alpha_j}(\theta) \geq h_{\sigma_\sA}(\theta)$ for every $\theta \geq 0$ (where we abuse notation and set $h_{\sigma_\sA}(\theta) \coloneqq  \theta^2/(2\sigma_\sA T + (2/3)\theta) - \tfrac{1}{2}\log(\pi(T+1))$). Hence
\[
  \sum_{j=1}^d \exp\!\big(h_{\sigma_\sA}(\abs{G_{T,j}})\big)
  \leq C_T.
\]
By Lemma~\ref{lem:bernstein-convex}, $\theta \mapsto h_{\sigma_\sA}(\theta) + \tfrac{1}{2}\log(\pi(T+1))$ is convex; the exponential is increasing and convex; so the composition $\theta \mapsto \exp(h_{\sigma_\sA}(\theta))$ is convex. Jensen's inequality applied to the uniform measure on $[d]$ gives
\[
  d\,\exp\!\big(h_{\sigma_\sA}(\bar G_T)\big)
  \leq \sum_{j=1}^d \exp\!\big(h_{\sigma_\sA}(\abs{G_{T,j}})\big)
  \leq C_T,
  \qquad
  \bar G_T \coloneqq  \frac{1}{d}\sum_{j=1}^d \abs{G_{T,j}}.
\]
Taking logarithms and rearranging,
\[
  \frac{\bar G_T^2}{2\sigma_\sA T + (2/3)\bar G_T}
  \leq \log\frac{C_T}{d} + \frac{1}{2}\log(\pi(T+1)) = \Lambda_T.
\]
Lemma~\ref{lem:bernstein-inversion} with $\sigma = \sigma_\sA$, $c = \Lambda_T$ gives $\bar G_T \leq \sqrt{2\sigma_\sA T \Lambda_T} + 2\Lambda_T/3$. Dividing by $T$ and noting $\frac{1}{d}\sum_j \abs{\delta_{T,j}} = \bar G_T/T$ yields~\eqref{eq:cw-upocp-l1}. For the asymptotic claim, $C_T/d = B_T/d + 1 \leq D T^{1+\gamma} + 1$ by Assumption~\ref{ass:poly-growth}, hence $\Lambda_T = \sO(\log T)$, and the right-hand side of~\eqref{eq:cw-upocp-l1} is $\sO(\sqrt{\sigma_\sA \log T/T})$.
\end{proof}

\subsection{Continuous quantile levels}
\label{app:continuous}

We extend the framework from a finite collection $\sA$ to a compact interval $\sX \subseteq (0,1)$. Fix a finite Borel reference measure $\mu$ with full support and work in the Hilbert space $\sH \coloneqq  L^2(\sX, \mu)$ with inner product $\inn{f}{h} \coloneqq  \int_\sX f(\alpha)h(\alpha)\,d\mu(\alpha)$. The radius forecast is now a function $b_t \in \sH$, and the score $S_t$ is a scalar with $S_t \leq D T^\gamma$ as in Assumption~\ref{ass:poly-growth}. Define the integrated pinball loss
\[
  \sL_t(b) \coloneqq  \int_\sX \ell^{(1-\alpha)}(b(\alpha), S_t)\,d\mu(\alpha),
\]
which is convex on $\sH$ and has subgradient $g_t(\alpha) \coloneqq  \bone\{b_t(\alpha) \geq S_t\} - (1-\alpha)$ in $L^\infty(\sX,\mu) \subseteq \sH$. The isotonic constraint becomes the cone of non-increasing functions:
\[
  \sK \coloneqq  \set{b \in \sH : b \text{ admits a non-increasing representative on } \sX}.
\]
$\sK$ is a closed convex cone in $\sH$, and the $L^2$-projection $\Pi_\sK$ is well-defined by the Hilbert projection theorem. Continuous PAVA characterizes $\Pi_\sK(b)$ as $-\frac{d\hat F}{d\mu}$, where $\hat F$ is the greatest concave majorant of $F(x) \coloneqq  -\int_{\alpha_{\min}}^x b(\alpha)\,d\mu(\alpha)$ \citep{robertson1988order}.

\paragraph{Inward flow, regret preservation, central inequality.} The proofs of Lemma~\ref{lem:projection-inequality}, Proposition~\ref{prop:regret-preservation}, and Theorem~\ref{thm:level-set} use only the inner-product structure of $\sH$ and so transfer verbatim. Inward flow of $(\sL_t, \sK)$ on $\sH$ is verified by the same case analysis as in Appendix~\ref{app:isotonic-inward-flow}, applied pointwise. The central inequality reads
\begin{equation}
\label{eq:central-continuous}
  F_T^\star\!\big({-G_T}\big) \leq B_T,
  \qquad
  G_T \coloneqq  \sum_{t=1}^T g_t \in \sH,
  \qquad
  B_T \coloneqq  \sum_{t=1}^T \sL_t(0).
\end{equation}
Under Assumption~\ref{ass:poly-growth}, $\sL_t(0) \leq \mu(\sX) \cdot S_t \leq \mu(\sX)\,D T^\gamma$, so $B_T \leq \mu(\sX)\,D T^{1+\gamma}$, depending on $\sX$ only through $\mu(\sX)$ and \emph{not} through any notion of dimensionality.

\paragraph{$L^2$ coverage via OGD.} Take $\sE$ to be OGD on $\sH$ with regularizer $\psi(f) = \tfrac{1}{2}\norm{f}_\sH^2$ and step size $\eta>0$. The regret bound is $F_T(u) = \norm{u}_\sH^2/(2\eta) + (\eta/2)\sum_t \norm{g_t}_\sH^2$, with conjugate $F_T^\star(\theta) = (\eta/2)\norm{\theta}_\sH^2 - (\eta/2)\sum_t \norm{g_t}_\sH^2$. Since $\norm{g_t}_\sH^2 = \int_\sX g_t(\alpha)^2\,d\mu(\alpha) \leq \mu(\sX)$, Theorem~\ref{thm:level-set} gives
\(
  \norm{G_T}_\sH^2 \leq 2 B_T/\eta + \mu(\sX)\,T.
\)
Choosing $\eta = 2 B_T/(\mu(\sX) T)$ to balance the two terms yields
\begin{equation}
\label{eq:l2-continuous}
  \norm{\delta_T}_{L^2(\sX,\mu)} \leq \sqrt{\frac{2\mu(\sX)}{T}},
  \qquad
  \delta_T \coloneqq  G_T/T.
\end{equation}
The bound is dimension-free: it depends on the size of the level set only through $\mu(\sX)$.

\paragraph{$L^q$ coverage via OMD.}
Take $\sE$ to be OMD with regularizer $\psi(f) = \tfrac{1}{2}\norm{f}_{L^p}^2$ for $p \in (1,2]$, $q = p/(p-1)$. The regularizer is $(p-1)$-strongly convex with respect to $\norm{\cdot}_{L^p}$~\citep[Sec.~4]{shalev2007thesis}, by the uniform convexity of $L^p$ in this range. The regret bound is $F_T(u) = \norm{u}_{L^p}^2/(2\eta) + (\eta/(2(p-1)))\sum_t \norm{g_t}_{L^q}^2$, with conjugate $F_T^\star(\theta) = (\eta/2)\norm{\theta}_{L^q}^2 - (\eta/(2(p-1)))\sum_t \norm{g_t}_{L^q}^2$. Since $\norm{g_t}_{L^q} \leq \mu(\sX)^{1/q}$,
\begin{equation}
\label{eq:lq-continuous}
  \norm{\delta_T}_{L^q} \leq \frac{1}{T}\sqrt{\frac{2 B_T}{\eta} + \frac{\mu(\sX)^{2/q}\,T}{p-1}}.
\end{equation}
Balancing the two terms gives $\norm{\delta_T}_{L^q} \leq \sqrt{2(q-1)\mu(\sX)^{2/q}/T}$.

\begin{remark}[Optimal $q$ depends on $\mu(\sX)$]
\label{rmk:lq-optimal-q}
Differentiating $g(q) \coloneqq  (q-1)\mu(\sX)^{2/q}$ in $q \in [2,\infty)$ shows that $g$ is non-decreasing on $[2, \infty)$ unless $\mu(\sX) > e^2$ (since $q^2 / (q - 1) \geq 4$ there), in which case the unique stationary point in $[2, \infty)$ satisfies $q^2 = 2(q-1)\log\mu(\sX)$ and is approximately $2\log\mu(\sX)$ for large $\mu(\sX)$.
For the Lebesgue measure or any probability measure on $\sX \subseteq (0,1)$, $\mu(\sX) \leq 1 < e^2$, so $g$ is increasing on $[2,\infty)$ and the optimum is $q = 2$, recovering the dimension-free $L^2$ bound~\eqref{eq:l2-continuous}. So OGD on $\sH$ already delivers the optimal rate among $L^q$ bounds.
\end{remark}

This corrects the optimization in the discrete-to-continuous translation: the discrete-case choice $q = 2\log d$ is beneficial only when the analogue $\mu(\sX) > e^2$ holds, which fails under Lebesgue and any normalized probability measure on $\sX$. In those cases, the $L^2$ bound~\eqref{eq:l2-continuous} \emph{is} the right answer.

\paragraph{Coordinate-wise universal portfolio on a continuum.}
Running an independent UP-OCP instance at each level $\alpha \in \sX$ with initial wealth $1$, the regret factorizes as $F_T(u) = \int_\sX F_T^{(\alpha)}(\abs{u(\alpha)})\,d\mu(\alpha)$, and similarly $F_T^\star(\theta) = \int_\sX (F_T^{(\alpha)})^\star(\abs{\theta(\alpha)})\,d\mu(\alpha)$. The central inequality~\eqref{eq:central-continuous} becomes
\begin{equation}
\label{eq:cw-central-cont}
  \int_\sX \exp\!\big(h_\alpha(\abs{G_T(\alpha)})\big)\,d\mu(\alpha) \leq C_T \coloneqq  B_T + \mu(\sX),
\end{equation}
where $h_\alpha$ is as defined in Appendix~\ref{app:cw-l1}.

To control miscoverage along an arbitrary direction in the level set, we record the Donsker--Varadhan change-of-measure inequality and then derive a coverage bound for any probability measure $\nu$ on $\sX$.

\begin{lemma}[Donsker--Varadhan change-of-measure; {\citep[Sec.~2.1]{boucheron2013concentration}}]
\label{lem:donsker-varadhan}
Let $(\sX, \sB)$ be a measurable space, $\bar\mu$ a probability measure on $\sX$, and $\nu$ a probability measure with $\nu \ll \bar\mu$. Then for every bounded measurable $\Phi \colon \sX \to \R$,
\[
  \int_\sX \Phi\,d\nu \leq \KL(\nu \| \bar\mu) + \log\!\int_\sX e^\Phi\,d\bar\mu.
\]
\end{lemma}

Fix any probability measure $\nu \ll \bar\mu$ on $\sX$ (with $\bar\mu \coloneqq  \mu/\mu(\sX)$), and define the $\nu$-variance envelope $\sigma_\nu \coloneqq  \sup_{\alpha \in \mathrm{supp}(\nu)} \alpha(1-\alpha)$ and the $\nu$-mean miscoverage $\bar G_\nu \coloneqq  \int_\sX \abs{G_T(\alpha)}\,d\nu(\alpha)$. Lemma~\ref{lem:donsker-varadhan} with $\Phi(\alpha) = h_\alpha(\abs{G_T(\alpha)})$, combined with~\eqref{eq:cw-central-cont} (so $\log\int e^\Phi\,d\bar\mu \leq \log(C_T/\mu(\sX))$), yields
\begin{equation}
\label{eq:nu-bound}
  \int_\sX h_\alpha(\abs{G_T(\alpha)})\,d\nu(\alpha)
  \leq \KL(\nu\|\bar\mu) + \log\frac{C_T}{\mu(\sX)}.
\end{equation}
Bounding $h_\alpha \geq h_{\sigma_\nu}$ $\nu$-almost everywhere and applying Jensen's inequality to the convex function $h_{\sigma_\nu}$ (Lemma~\ref{lem:bernstein-convex}) under the probability measure $\nu$ gives
\(
  h_{\sigma_\nu}(\bar G_\nu) \leq \int_\sX h_{\sigma_\nu}(\abs{G_T(\alpha)})\,d\nu(\alpha) \leq \KL(\nu\|\bar\mu) + \log(C_T/\mu(\sX)),
\)
i.e., $\bar G_\nu^2/(2\sigma_\nu T + (2/3)\bar G_\nu) \leq R_\nu$ with $R_\nu \coloneqq  \KL(\nu\|\bar\mu) + \log(C_T/\mu(\sX)) + \tfrac{1}{2}\log(\pi(T+1))$. Lemma~\ref{lem:bernstein-inversion} then yields the general coverage bound
\begin{equation}
\label{eq:nu-mean}
  \int_\sX \abs{\delta_T(\alpha)}\,d\nu(\alpha)
  \leq \sqrt{\frac{2\sigma_\nu R_\nu}{T}} + \frac{2 R_\nu}{3T}.
\end{equation}

\paragraph{Sub-interval coverage as a special case.}
Take $\sI \subseteq \sX$ measurable with $\mu(\sI) > 0$ and set $\nu = \mu|_\sI/\mu(\sI)$. Then $\KL(\nu\|\bar\mu) = \log(\mu(\sX)/\mu(\sI))$, $\sigma_\nu = \sigma_\sI \coloneqq  \sup_{\alpha \in \sI} \alpha(1-\alpha)$, and $\bar G_\nu = \bar G_\sI \coloneqq  \frac{1}{\mu(\sI)}\int_\sI \abs{G_T(\alpha)}\,d\mu(\alpha)$, so~\eqref{eq:nu-mean} reads
\begin{equation}
\label{eq:interval-mean}
  \frac{1}{\mu(\sI)}\int_\sI \abs{\delta_T(\alpha)}\,d\mu(\alpha)
  \leq \sqrt{\frac{2\sigma_\sI R_\sI}{T}} + \frac{2 R_\sI}{3T},
  \qquad
  R_\sI \coloneqq  \log\frac{C_T}{\mu(\sI)} + \tfrac{1}{2}\log(\pi(T+1)).
\end{equation}
Specializing further to $\sI = \sX$ recovers the global mean-absolute miscoverage of Theorem~\ref{thm:cw-upocp-l1}; specializing to a small interval near $\alpha = 0$ or $\alpha = 1$ exploits the smaller variance envelope $\sigma_\sI$, mirroring the level-adaptive behavior of the discrete case.

\paragraph{Reduction to the discrete setting.}
Specializing $\sX = \sA = \{\alpha_1, \dots, \alpha_d\}$ and $\mu$ the counting measure recovers $\mu(\sX) = d$ and reduces every continuous bound to its finite analogue: \eqref{eq:l2-continuous} becomes the OGD bound, \eqref{eq:lq-continuous} becomes Corollary~\ref{cor:p-norm}, \eqref{eq:cw-central-cont} becomes~\eqref{eq:cw-central}, and~\eqref{eq:interval-mean} with $\sI = \sX$ recovers Theorem~\ref{thm:cw-upocp-l1}. In particular, the dichotomy in Remark~\ref{rmk:lq-optimal-q} explains why the discrete choice $q = 2\log d$ helps for many levels (the counting measure makes $\mu(\sX) = d$, and $d > e^2 \approx 7.4$) but does not transfer to Lebesgue.

\subsection{Detailed comparison with prior work}
\label{app:prior-comparison}

\begin{table}[H]
\centering
\caption{Comparison with the closest prior work. Rates are for $\max_i \MisCov_{T, i}$; $d_\sA \coloneqq  \min_i \min\{\alpha_i, 1 - \alpha_i\}$ ($= \alpha_1$ when $\alpha_1 + \alpha_d \leq 1$).}
\label{tab:prior-comparison}
\footnotesize
\setlength{\tabcolsep}{4pt}
\renewcommand{\arraystretch}{1.1}
\begin{tabular}{@{}>{\raggedright\arraybackslash}p{1.85cm}>{\raggedright\arraybackslash}p{2.75cm}>{\raggedright\arraybackslash}p{3.45cm}>{\raggedright\arraybackslash}p{5.0cm}@{}}
\toprule
 & \citet{liu2026online} & \citet{ding2025calibrated} (\MultiQT) & This paper \\
\midrule
Scores & $S_t \leq D t^\gamma$ & $S_t \leq D$ & $S_t \leq D T^\gamma$ (Ass.~\ref{ass:poly-growth}) \\
Algorithms & any scalar learner with linearized regret & OGD with lazy isotonic projection & any learner on $\R^d$ with linearized regret (Alg.~\ref{alg:lazy-projection}, Prop.~\ref{prop:regret-preservation}) \\
Guarantee & scalar Fenchel level set & $\ell_2$ ball, via restorativity & level set $F_T^\star(-\bG_T) \leq B_T$ (Thm.~\ref{thm:level-set}) \\
Rate, general & single level only & $\sO\big(\sqrt{d / T} + d \sqrt{D / (\eta\, d_\sA T)}\big)$ & $\sO(\sqrt{\log d / T})$ (Cor.~\ref{cor:p-norm}); parameter-free: $\sO\big(\sqrt{(\log d + (1 + \gamma) \log T)/T}\big)$ (Thm.~\ref{thm:cw-upocp-per-level}) \\
Rate, common point forecast & n/a & $\sO\big(d^{3/2} (1 + D / \eta) / (d_\sA T)\big)$, fixed $\eta$, linear regret & not recovered by our analysis \\
Other results & none & sharpness gains (empirical) & WIS reduction (Thm.~\ref{thm:per-round-reduction}); continuous levels (App.~\ref{app:continuous}) \\
\bottomrule
\end{tabular}
\end{table}

Table~\ref{tab:prior-comparison} compares our results with the two works we build on, and this subsection details it. We state the results of \citet{ding2025calibrated} in our notation: with symmetric sets, every level shares the base forecast $\hat{Y}_t$, so in score space \MultiQT runs on the scores $(S_t)$ with the zero base forecast at every level, quantile levels $1 - \alpha_i$, and initialization $\bxtilde_1 = \bzero$. We write $d_\sA \coloneqq  \min_i \min\{\alpha_i, 1 - \alpha_i\}$, which equals $\alpha_1$ whenever $\alpha_1 + \alpha_d \leq 1$.

\paragraph{What the combination adds.}
The ingredients of Section~\ref{sec:reduction} are known, and Proposition~\ref{prop:regret-preservation} and Theorem~\ref{thm:level-set} are short combinations of them (see the discussion after Theorem~\ref{thm:level-set}). The combination nonetheless changes what can be proved.
\begin{enumerate}[leftmargin=*,nosep]
\item \emph{Black-box generality.} The coverage proof of \citet{ding2025calibrated} is not a regret argument: it relies on the OGD update and on the restorativity of the aggregated pinball loss, which they derive from bounded residuals~\citep[Lemma~1]{ding2025calibrated}, so it does not apply to other learners. The duality argument of \citet{liu2026online} is scalar and unconstrained. Proposition~\ref{prop:regret-preservation} gives the played, projected sequence the regret bound of the base learner against every $\bu \in \R^d$, which is exactly what the duality argument needs, so any learner with a linearized-regret bound can be used.
\item \emph{Naive combinations fail.} Running independent single-level learners and sorting or projecting their outputs post hoc can break coverage, and so can (non-lazy) projected OGD~\citep[Props.~2 and~4]{ding2025calibrated}. Algorithm~\ref{alg:lazy-projection} avoids both failure modes because the projection is lazy and the subgradient is evaluated at the played iterate.
\item \emph{Two steps behind $\sqrt{d} \to \sqrt{\log d}$.} The geometry of the coverage-error region is a design choice ($q = 2 \log d$ in Corollary~\ref{cor:p-norm}), and for coordinate-wise \UPOCP, a single aggregate budget $B_T$ splits into $d$ simultaneous per-level bounds at a $\log d$ price (Theorem~\ref{thm:cw-upocp-per-level}).
\item \emph{Continuous levels.} Regret preservation transfers verbatim to a Hilbert space (Appendix~\ref{app:continuous}), which gives coverage guarantees for a continuum of levels, an extension that \citet{ding2025calibrated} leave to future work.
\item \emph{Sharpness.} Theorem~\ref{thm:per-round-reduction} explains the empirical advantage of \MultiQT over naive forecasters reported by \citet{ding2025calibrated}.
\end{enumerate}

\paragraph{The two regimes of \MultiQT.}
Under Assumption~\ref{ass:poly-growth} with $\gamma = 0$ (residual bound $D$) and a constant step size $\eta$, the general guarantee of \citet[Thm.~1]{ding2025calibrated} reads
\[
  \max_{i \in [d]} \MisCov_{T, i}
  \leq \sqrt{\frac{d}{T} + \frac{2 D d^2}{\eta\, d_\sA T}},
\]
and, since all levels share the same point forecast, their fast rate for point forecasts~\citep[Lemma~3 and Cor.~1]{ding2025calibrated} applies and reads
\[
  \max_{i \in [d]} \MisCov_{T, i}
  \leq \frac{\sqrt{d}}{T} + \frac{d^{3/2}}{2 d_\sA T} + \frac{D d^{3/2}}{d_\sA \eta T} + \frac{d^{3/2}}{\sqrt{3}\, T},
\]
an $\sO(1/T)$ rate for fixed $\eta$, $d$, $d_\sA$, and $D$ (both bounds hold for the $\ell_2$ norm of the coverage-error vector, hence for its largest coordinate). Our analysis does not recover this fast rate. Three qualifications put it in context.
\begin{itemize}[leftmargin=*,nosep]
\item \emph{Coupling with regret.} The regret bound of \citet[Thm.~3]{ding2025calibrated} reads $D^2 d / (\eta T) + \eta d / 2$ for the average regret. For a fixed $\eta$ it does not vanish, i.e., the cumulative regret bound grows linearly in $T$, and with it the control of the WIS, hence of the width, is lost. The same holds for any step size that does not decay with $T$, including the balanced one of Corollary~\ref{cor:p-norm} when $\gamma = 0$. Balancing calibration against regret, \citet[Sec.~4.3]{ding2025calibrated} select $\eta = \sO(T^{-1/2})$ and obtain $\sO(T^{-1/2})$ for both; coordinate-wise \UPOCP attains the same rate in $T$ up to a $\sqrt{\log T}$ factor together with sublinear regret, without any tuning (Theorem~\ref{thm:cw-upocp-per-level}).
\item \emph{Dependence on the levels.} The constants scale as $d^{3/2} / d_\sA$. At our configuration ($d = 100$ levels down to $\alpha_1 = 10^{-3}$, $T = 3{,}000$), the second term alone equals $10^3 / 6 \approx 167$, so the bound is vacuous, whereas Corollary~\ref{cor:p-norm} with $q = 2 \log d$ and the balanced $\eta$ gives $\sqrt{2e(2 \log d - 1)/T} \approx 0.12$.
\item \emph{Score growth.} Under Assumption~\ref{ass:poly-growth} with $\gamma > 0$, the residual bound becomes $D T^\gamma$, so for an $\eta$ fixed independently of $T$ the dominant terms degrade to $\sO(T^{\gamma - 1})$: slower than $T^{-1/2}$ for $\gamma > 1/2$, and no longer vanishing for $\gamma \geq 1$. Scaling $\eta \propto T^\gamma$ restores the rates but requires knowing $D$ and $\gamma$, as does the balanced step size of Corollary~\ref{cor:p-norm}; coordinate-wise \UPOCP needs no such knowledge.
\end{itemize}
Our improvements are therefore in the dependence on $d$, parameter-free operation under polynomially growing scores, and black-box generality, not in the dependence on $T$.

\section{Deferred proofs for Section~\ref{sec:projection-benefits}}
\label{app:section-projection-benefits}

This appendix collects the proofs deferred from Section~\ref{sec:projection-benefits}: the WIS--pinball identity~\eqref{eq:wis-pinball-identity} and the per-round structural reduction (Theorem~\ref{thm:per-round-reduction}).

\subsection{Pinball loss as a Weighted Interval Score (Equation~\eqref{eq:wis-pinball-identity})}
\label{app:wis-pinball}

We verify the identity~\eqref{eq:wis-pinball-identity}. The WIS is a function of the interval endpoints; a zero radius, which encodes the empty set, has width (Lebesgue measure) $0$ and is evaluated at the degenerate endpoints $l = u = \hat{Y}_t$. For symmetric central intervals $[\hat{Y}_t - x_{t, i}, \hat{Y}_t + x_{t, i}]$, $x_{t, i} \geq 0$, at miscoverage levels $\alpha_1 < \cdots < \alpha_d$ evaluated at the realized response $Y_t$,
\[
  \WIS_\sA\!\big(\set{[\hat{Y}_t - x_{t, i}, \hat{Y}_t + x_{t, i}]}_{i = 1}^d, Y_t\big)
  \;=\; 2 \sL_t(\bx_t) + 2 S_t \sum_{i = 1}^d \alpha_i.
\]

First of all, observe that $\WIS_\sA\!\big(\{[l_{\alpha_i}, u_{\alpha_i}]\}_{i = 1}^d, y\big)=\sum_{i=1}^d \alpha_i \IS_{\alpha_i}([l_{\alpha_i}, u_{\alpha_i}], y)$, where $\IS_{\alpha_i}([l_{\alpha_i}, u_{\alpha_i}], y) \coloneqq  (u_{\alpha_i}-l_{\alpha_i})+ \frac{2}{\alpha_i} \dist\left(y, [l_{\alpha_i}, u_{\alpha_i}]\right)$.

With $D \coloneqq  Y_t - \hat{Y}_t$ and $S_t = \abs{D}$, we have
\[
  \max(l_{\alpha_i} - Y_t,0) = \max(-D - x_{t, i},0),
  \quad \text{ and } \quad \max(Y_t - u_{\alpha_i},0) = \max(D - x_{t, i},0).
\]
Because $D$ has a definite sign and $x_{t, i} \geq 0$ (all played and hidden radii are valid by the convention of Section~\ref{sec:reduction}), exactly one of these two non-negative quantities can be non-zero, and it equals $\max(\abs{D} - x_{t, i},0) = \max(S_t - x_{t, i},0)$. Hence
\[
  \IS_{\alpha_i}([\hat{Y}_t - x_{t, i}, \hat{Y}_t + x_{t, i}], Y_t)
  = 2 x_{t, i} + \frac{2}{\alpha_i} \max(S_t - x_{t, i},0),
\]
and therefore $\alpha_i \IS_{\alpha_i} = 2 \alpha_i x_{t, i} + 2 \max(S_t - x_{t, i},0)$. Summing over $i \in [d]$,
\[
  \WIS_\sA = 2 \sum_{i = 1}^d \alpha_i x_{t, i} + 2 \sum_{i = 1}^d \max(S_t - x_{t, i},0).
\]
The pointwise identity $\ell^{(1 - \alpha)}(x, S) = \alpha (x - S) + \max(S - x,0)$ (a direct case analysis on $x \gtreqless S$) gives, after summation,
\[
  \sL_t(\bx_t) = \sum_{i = 1}^d \alpha_i x_{t, i} + \sum_{i = 1}^d \max(S_t - x_{t, i},0) - S_t \sum_{i = 1}^d \alpha_i,
\]
that is, $\sL_t(\bx_t) + S_t \sum_{i=1}^d \alpha_i = \sum_{i=1}^d \alpha_i x_{t,i} + \sum_{i=1}^d \max(S_t - x_{t,i},0)$, which, combined with the previous display, is~\eqref{eq:wis-pinball-identity}.

\subsection{Per-round structural reduction (Theorem~\ref{thm:per-round-reduction})}
\label{app:projection-benefits}

The proof of Theorem~\ref{thm:per-round-reduction} relies on two structural lemmas about the PAVA decomposition of the isotonic projection. Throughout, fix $\bxtilde \in \R^d$ (the lemmas are algebraic and need no sign condition), let $\bx \coloneqq  \Pi_\sK(\bxtilde)$ with PAVA blocks $B_1, \dots, B_p$ (each $B_k = \{l_k, l_k + 1, \dots, r_k\}$), and let $\overline{\tilde{x}}_{B_k} \coloneqq  |B_k|^{-1} \sum_{i \in B_k} \tilde{x}_i$ denote the block average, so $x_j = \overline{\tilde{x}}_{B_k}$ for all $j \in B_k$.

\begin{lemma}[Partial-average property of PAVA]
\label{lem:partial-average}
For every block $B_k = \{l, \dots, r\}$ and every $j \in [l, r]$,
\[
  \frac{1}{j - l + 1} \sum_{i = l}^j \tilde{x}_i \;\leq\; \overline{\tilde{x}}_{B_k}.
\]
\end{lemma}

\begin{proof}
Recall that $\bx$ solves the quadratic program
\(
  \min_{\bz \in \sK} \frac{1}{2} \norm{\bz - \bxtilde}_2^2,
\)
where $\sK = \{\bz : z_1 \geq z_2 \geq \cdots \geq z_d\}$. Introducing non-negative Lagrange multipliers $\lambda_i \geq 0$ for the constraints $z_i \geq z_{i + 1}$, $i \in [d - 1]$ (with the convention $\lambda_0 = \lambda_d = 0$), the KKT stationarity condition at the optimum reads, for $i \in [d]$,
\[
  \tilde{x}_i - x_i = \lambda_{i - 1} - \lambda_i.
\]
Within block $B_k = \{l, \dots, r\}$, the projected entries are constant, so the constraints $x_{l - 1} > x_l$ and $x_r > x_{r + 1}$ are strict at the boundary of the block (with the convention that $x_0 = +\infty$ and $x_{d + 1} = -\infty$). Complementary slackness then forces $\lambda_{l - 1} = \lambda_r = 0$. Summing the stationarity equation over $i = l, \dots, j$ and telescoping,
\[
  \sum_{i = l}^j (\tilde{x}_i - \overline{\tilde{x}}_{B_k})
  = \sum_{i = l}^j (\lambda_{i - 1} - \lambda_i)
  = \lambda_{l - 1} - \lambda_j
  = - \lambda_j \;\leq\; 0,
\]
where the last inequality uses dual feasibility $\lambda_j \geq 0$. Dividing by $j - l + 1$ gives the claim.
\end{proof}

\begin{lemma}[Weighted-sum reduction within a block]
\label{lem:weighted-sum}
Let $w_l \leq w_{l + 1} \leq \cdots \leq w_r$ be non-decreasing weights on a block $B_k = \{l, \dots, r\}$. Then
\[
  \sum_{i \in B_k} w_i \, x_i \;=\; \overline{\tilde{x}}_{B_k} \sum_{i \in B_k} w_i
  \;\leq\; \sum_{i \in B_k} w_i \, \tilde{x}_i.
\]
The inequality is strict whenever $\{\tilde{x}_i\}_{i \in B_k}$ is non-constant and $w$ is strictly increasing on $[l, r]$.
\end{lemma}

\begin{proof}
Let $c_i \coloneqq  \tilde{x}_i - \overline{\tilde{x}}_{B_k}$, so $\sum_{i = l}^r c_i = 0$ and, by Lemma~\ref{lem:partial-average}, $\sum_{i = l}^j c_i \leq 0$ for every $j \in [l, r]$. Abel summation yields
\[
  \sum_{i = l}^r w_i c_i
  = w_r \cdot \underbrace{\sum_{i = l}^r c_i}_{= 0}
    - \sum_{j = l}^{r - 1} (w_{j + 1} - w_j) \sum_{i = l}^j c_i
  = - \sum_{j = l}^{r - 1} \underbrace{(w_{j + 1} - w_j)}_{\geq 0} \underbrace{\sum_{i = l}^j c_i}_{\leq 0}
  \;\geq\; 0,
\]
which gives $\sum_{i = l}^r w_i \tilde{x}_i \geq \overline{\tilde{x}}_{B_k} \sum_{i = l}^r w_i = \sum_{i = l}^r w_i x_i$. For strictness, observe that if $\{\tilde{x}_i\}_{i \in B_k}$ is non-constant, then $\sum_{i = l}^j c_i < 0$ for some $j \in [l, r - 1]$; if in addition $w_{j + 1} > w_j$, the corresponding term in the Abel sum is strictly positive.
\end{proof}

For convenience, we restate Theorem~\ref{thm:per-round-reduction} below, attaching labels to its individual reduction inequalities so that the proof can refer to them.

\begin{theorem}[Per-round reduction; restatement of Theorem~\ref{thm:per-round-reduction}]
For every round $t$, the hidden iterate $\bxtilde_t \in \R^d_+$ and the played iterate $\bx_t = \Pi_\sK(\bxtilde_t) \in \sK \cap \R^d_+$ satisfy the per-round sharpness and coverage reductions
\begin{align}
  \sum_{i = 1}^d \alpha_i x_{t, i}
  &\leq \sum_{i = 1}^d \alpha_i \tilde{x}_{t, i}, \label{eq:sharpness-reduction} \\
  \sum_{i = 1}^d \max(S_t - x_{t, i}, 0)
  &\leq \sum_{i = 1}^d \max(S_t - \tilde{x}_{t, i}, 0), \label{eq:coverage-reduction}
\end{align}
and consequently the per-round loss reduction
\begin{equation} \label{eq:loss-reduction}
  \sL_t(\bx_t) \;\leq\; \sL_t(\bxtilde_t).
\end{equation}
Both~\eqref{eq:sharpness-reduction} and~\eqref{eq:loss-reduction} are strict whenever $\bxtilde_t \notin \sK$.
\end{theorem}

\begin{proof}[Proof of Theorem~\ref{thm:per-round-reduction}]
By the WIS decomposition~\eqref{eq:wis-pinball-identity}, the loss reduction~\eqref{eq:loss-reduction} is implied by the sharpness reduction~\eqref{eq:sharpness-reduction} and the coverage reduction~\eqref{eq:coverage-reduction}, since the constant $S_t \sum_i \alpha_i$ cancels. We prove each block by block.

\emph{Sharpness.} On any block $B_k$, the miscoverage levels $\alpha_l < \alpha_{l + 1} < \cdots < \alpha_r$ are strictly increasing (by our convention). Lemma~\ref{lem:weighted-sum} with $w_i = \alpha_i$ then gives $\sum_{i \in B_k} \alpha_i x_{t, i} \leq \sum_{i \in B_k} \alpha_i \tilde{x}_{t, i}$, with strict inequality whenever $\{\tilde{x}_{t, i}\}_{i \in B_k}$ is non-constant. PAVA pools coordinates only when consecutive (sub-)block averages violate the order, which forces a strict difference between elements; hence any block of size $|B_k| \geq 2$ has non-constant entries. Summing over $k$ gives~\eqref{eq:sharpness-reduction}, with strict inequality whenever $\bxtilde_t \notin \sK$ (equivalently, whenever some block has size $\geq 2$).

\emph{Coverage.} The map $\phi(x) \coloneqq  \max(S_t - x,0)$ is convex on $\R$. Jensen's inequality applied to $\bxtilde_{t}$ restricted to $B_k$ gives
\[
  \sum_{i \in B_k} \max(S_t - x_{t, i},0) \;=\; |B_k| \, \phi(\overline{\tilde{x}}_{B_k})
  \;\leq\; \sum_{i \in B_k} \phi(\tilde{x}_{t, i})
  \;=\; \sum_{i \in B_k} \max(S_t - \tilde{x}_{t, i},0).
\]
Summing over $k$ gives~\eqref{eq:coverage-reduction}.

\emph{Loss.} Summing~\eqref{eq:sharpness-reduction} and~\eqref{eq:coverage-reduction} and applying~\eqref{eq:wis-pinball-identity} gives~\eqref{eq:loss-reduction}, which is strict whenever~\eqref{eq:sharpness-reduction} is strict, i.e., whenever $\bxtilde_t \notin \sK$.
\end{proof}

\section{Experimental details and additional results}
\label{app:experiments-extended}

This appendix collects the dataset descriptions, forecasting-pipeline configurations, and complete results referenced from Section~\ref{sec:experiments}.

\subsection{Datasets and forecasting pipelines}
\label{app:experiments-datasets}

\paragraph{Synthetic streams.} All synthetic scores are non-negative trajectories, truncated at zero. Each trajectory is generated from a fixed base seed; the 10 seeds we use are $42, 43, \dots, 51$.
\begin{itemize}[leftmargin=*,nosep]
  \item \emph{Sinusoidal}: $S_t = \max(0, m + a \sin(2\pi t/p) + \epsilon_t)$ with center $m = 11$ and amplitude $a = 9$ (so that $S_t$ oscillates between $\{2, 20\}$), period $p = 200$, $\epsilon_t \sim \sN(0, 0.3^2)$, $T = 3{,}000$. Stresses cyclic non-stationarity.
  \item \emph{Stationary and mixed regimes}: a constant baseline or a quadratic trend overlaid with rare exponential spikes, smoothed by a $25$-step rolling maximum (Appendices~\ref{app:results-stationary} and~\ref{app:results-mixed}). Stresses heavy tails.
\end{itemize}

\paragraph{Real-world streams.}
\begin{itemize}[leftmargin=*,nosep]
  \item \emph{AAPL, AMZN, GOOGL}: daily opening prices of the three stocks, from~\citet{nguyen2018s}, covering the period 2006-01-03 to 2017-12-29 ($T = 3{,}020$ rounds: the $3{,}019$ trading days of each stock plus one date of the panel missing for them, filled by interpolation). Forecasting model: Prophet~\citep{taylor2018forecasting} re-fit at every step on log-prices using all data up to time $t-1$ to produce a one-step-ahead forecast $\hat Y_t$. Nonconformity score $S_t = \abs{Y_t - \hat Y_t}$.
  \item \emph{Electricity (\texttt{elec2})}: half-hourly New South Wales electricity demand from~\citep{harries1999splice}, restricted to a non-stationary regime; one-step-ahead forecasts produced by an autoregressive model.
\end{itemize}

\subsection{Detailed update rules}
\label{app:algorithms}

This subsection records the explicit update rules of the four method families compared in Section~\ref{sec:experiments}: coordinate-wise universal portfolio, $p$-norm online mirror descent, coordinate-wise Krichevsky--Trofimov bettor, and \MultiQT. The first three are base online learners that we run \emph{both} in their vanilla form and through the lazy projection wrapper of Algorithm~\ref{alg:lazy-projection}; \MultiQT, which is by construction a projected algorithm, is presented in its single canonical form. All wrapped variants share the same outer loop: receive the hidden iterate $\bxtilde_t \in \R^d_+$ from the base learner, project to obtain $\bx_t = \Pi_\sK(\bxtilde_t)$, observe the score $S_t$, compute the subgradient $\bg_t$ at $\bx_t$ via~\eqref{eq:pinball-subgradient}, and feed $\bg_t$ back to the base learner. The unprojected variants are obtained by replacing the projected play $\bx_t = \Pi_\sK(\bxtilde_t)$ by $\bx_t = \bxtilde_t$ throughout, leaving the rest of the update untouched. Every base learner, including \MultiQT, outputs valid radii following Section~\ref{sec:reduction}: its raw radius $z_{t,j}$ is replaced by $\tilde{x}_{t,j} = \max(z_{t,j}, 0)$, where $0$ encodes the empty set, while its internal state and update rule are left unchanged~\citep[App.~F.2]{liu2026online}.

\subsubsection{Coordinate-wise Universal Portfolio (\UPOCP)}
\label{app:cw-upocp-rules}

\paragraph{Algorithm.} Coordinate-wise \UPOCP runs an independent scalar UP-OCP instance~\citep{liu2026online} at each miscoverage level $\alpha_j$ and synchronizes only at prediction time through the projection. At level $\alpha_j$, the scalar instance treats the betting game as a two-asset \emph{conformal market} \citep[Def.~4.1]{liu2026online} with returns $w_{t,1}^{(j)} = -g_{t,j}/\alpha_j + 1$ and $w_{t,2}^{(j)} = 1 + g_{t,j}/(1 - \alpha_j)$, and chooses portfolio weight $\lambda_t^{(j)}$ as the wealth-weighted average over $[0,1]$ under the Beta$(1/2, 1/2)$ prior. Specializing the universal-portfolio update to this market admits the closed form
\[
  \lambda_t^{(j)}
  \;=\; \frac{1}{t}\!\left(\sum_{s = 1}^{t-1} \bone\{g_{s,j\} = -(1 - \alpha_j)} + \tfrac{1}{2}\right)
  \;=\; \frac{N_{t-1}^{(j)} + 1/2}{t},
\]
where $N_{t-1}^{(j)}$ counts past per-level miscoverage events. The hidden iterate is the wealth-to-radius mapping $\tilde{x}_{t,j} = \max\big(W_{t-1}^{(j)}\,(\lambda_t^{(j)} - \alpha_j)/(\alpha_j (1 - \alpha_j)), 0\big)$, clipped as in the scalar \UPOCP~\citep[App.~F.2]{liu2026online}, and the wealth recursion $W_t^{(j)} = W_{t-1}^{(j)} \cdot (\lambda_t^{(j)} w_{t,1}^{(j)} + (1 - \lambda_t^{(j)}) w_{t,2}^{(j)})$ simplifies, on the binary outcome $g_{t,j} \in \{-(1-\alpha_j), \alpha_j\}$, to the multiplicative updates in Algorithm~\ref{alg:cw-upocp}. The coordinate-wise nature is essential: the per-level wealth and miscoverage counter make every level adapt to its own variance budget $\alpha_j(1 - \alpha_j)$, which is what ultimately produces the $\sqrt{\log d}$ rate of Theorem~\ref{thm:cw-upocp-per-level}.

\paragraph{Hyperparameters.} \UPOCP is parameter-free; the initial wealth is fixed at $1$ for every dataset and every level (see Section~\ref{sec:experiments}), matching the theoretical analysis. No tuning is performed.

\paragraph{Projected vs.\ unprojected.} The projected variant \UPOCP-\textsc{Iso} plays $\bx_t = \Pi_\sK(\bxtilde_t)$ on Line~\ref{ln:cw-upocp-proj} of Algorithm~\ref{alg:cw-upocp} below; the unprojected variant simply plays $\bx_t = \bxtilde_t$. Both feed the gradient evaluated at the played iterate back into the per-level wealth recursions.

\begin{algorithm}[ht]
\caption{Coordinate-wise \UPOCP with lazy projection.}
\label{alg:cw-upocp}
\begin{algorithmic}[1]
  \STATE \textbf{Input:} miscoverage levels $\alpha_1 < \cdots < \alpha_d$
  \STATE Initialize $W_0^{(j)} \leftarrow 1$ and miscoverage count $N^{(j)} \leftarrow 0$ for each $j \in [d]$
  \FOR{$t = 1, 2, \dots$}
    \FOR{$j = 1, \dots, d$ (in parallel)}
      \STATE $\lambda_t^{(j)} \leftarrow (N^{(j)} + 1/2)/t$
      \STATE $\tilde{x}_{t,j} \leftarrow \max\big(W_{t-1}^{(j)} \cdot (\lambda_t^{(j)} - \alpha_j) / (\alpha_j(1 - \alpha_j)),\, 0\big)$ \quad (\emph{$0$ encodes $\emptyset$})
    \ENDFOR
    \STATE $\bx_t \leftarrow \Pi_\sK(\bxtilde_t)$ \quad (\emph{unprojected:} $\bx_t \leftarrow \bxtilde_t$);\quad observe $S_t$ \label{ln:cw-upocp-proj}
    \FOR{$j = 1, \dots, d$ (in parallel)}
      \IF{$\hat{C}_t(\hat{Y}_t, x_{t,j})$ does not cover $Y_t$, i.e., $S_t > x_{t,j}$ or $x_{t,j} = 0$}
        \STATE $N^{(j)} \leftarrow N^{(j)} + 1$;\quad $W_t^{(j)} \leftarrow W_{t-1}^{(j)} \cdot \lambda_t^{(j)} / \alpha_j$
      \ELSE
        \STATE $W_t^{(j)} \leftarrow W_{t-1}^{(j)} \cdot (1 - \lambda_t^{(j)}) / (1 - \alpha_j)$
      \ENDIF
    \ENDFOR
  \ENDFOR
\end{algorithmic}
\end{algorithm}

\subsubsection{$p$-norm Online Mirror Descent}
\label{app:pnorm-omd-rules}

\paragraph{Algorithm.} \PNormOMD\ is online mirror descent on $\R^d$ with regularizer $\psi(\bx) = \tfrac{1}{2}\norm{\bx}_p^2$ for $p \in (1, 2]$ and conjugate exponent $q = p/(p-1)$. Maintaining a dual iterate $\btheta_t \in \R^d$ initialized at $\bzero$, the primal mirror map is the gradient of the conjugate regularizer~\citep{shalev2007thesis,orabona2019modern}, which yields the explicit closed form of the raw radius (clipped as $\tilde{x}_{t,j} = \max(z_{t,j}, 0)$, Section~\ref{sec:reduction})
\[
  z_{t,j} \;=\; \norm{\btheta_t}_q^{2-q}\,\sign(\theta_{t,j})\,\abs{\theta_{t,j}}^{q-1},
  \qquad j = 1, \dots, d,
\]
followed by the standard dual update $\btheta_{t+1} = \btheta_t - \eta \bg_t$. The choice $p = 2 \log d/(2 \log d - 1)$ recommended by Corollary~\ref{cor:p-norm} interpolates between OGD ($p = 2$) and $\ell_\infty$-style updates as $d$ grows: it makes the regularizer $(p-1)$-strongly convex with respect to $\norm{\cdot}_p$ while keeping the conjugate gradient norm $\norm{\bg_t}_q^2$ at most $d^{2/q} = e$, eliminating the polynomial dimension dependence of $\ell_2$ regret. The case $p = 2$ recovers OGD with lazy isotonic projection, i.e., \MultiQT with a constant step size, and is the closest direct comparison with it.

\paragraph{Hyperparameters and tuning.} The constant step size $\eta$ is the only tuned knob. Following standard practice for adaptive conformal baselines~\citep{podkopaev2024adaptive,ding2025calibrated}, we run a per-dataset grid search over $\eta \in \{2^{-2}, 2^{-1}, 1, 2, 4, 8, 16, 32, 64, 128, 256, 512, 1024\}$ and report the value whose mean-width vs.\ empirical-coverage Pareto frontier dominates the others on $\alpha \in [0.03, 0.25]$. For both AAPL and the sinusoidal stream the selected $\eta$ is $64$. We emphasize that this oracle-style selection is a best-case, not realistic, deployment of the method: a practitioner without access to the full stream cannot select $\eta$ this way. The exponent $p$ is fixed at $2 \log d/(2 \log d - 1)$ from theory and not tuned. Initialization is $\btheta_1 = \bzero$.

\paragraph{Projected vs.\ unprojected.} The projected variant $p$-norm-OMD-\textsc{Iso} plays $\bx_t = \Pi_\sK(\bxtilde_t)$ on Line~\ref{ln:pnorm-proj} below; the unprojected variant plays $\bx_t = \bxtilde_t$. The dual update is identical in both cases.

\begin{algorithm}[ht]
\caption{$p$-norm online mirror descent with lazy projection.}
\label{alg:pnorm-omd}
\begin{algorithmic}[1]
  \STATE \textbf{Input:} $p \in (1,2]$ (with conjugate exponent $q \coloneqq  p/(p-1)$), step size $\eta > 0$. Set $\btheta_1 = \bzero$
  \FOR{$t = 1, 2, \dots$}
    \STATE $\tilde{x}_{t,j} \leftarrow \max\big(\norm{\btheta_t}_q^{2-q}\sign(\theta_{t,j}) \cdot \abs{\theta_{t,j}}^{q-1},\, 0\big)$ for $j = 1, \dots, d$ \quad (\emph{$0$ encodes $\emptyset$})
    \STATE $\bx_t \leftarrow \Pi_\sK(\bxtilde_t)$ \quad (\emph{unprojected:} $\bx_t \leftarrow \bxtilde_t$);\quad observe $S_t$;\quad $\bg_t \in \partial \sL_t(\bx_t)$ via~\eqref{eq:pinball-subgradient} \label{ln:pnorm-proj}
    \STATE $\btheta_{t+1} \leftarrow \btheta_t - \eta \, \bg_t$
  \ENDFOR
\end{algorithmic}
\end{algorithm}

\subsubsection{Coordinate-wise Krichevsky--Trofimov bettor (\KT)}
\label{app:cw-kt-rules}

\paragraph{Algorithm.} Coordinate-wise \KT~\citep{podkopaev2024adaptive,liu2026online} runs an independent KT bettor at each level $\alpha_j$, treating the per-level subgradients as the outcomes of a coin-betting game. At level $\alpha_j$, the bettor maintains wealth $W_t^{(j)}$ and a betting fraction $\beta_t^{(j)} = -\Sigma_{t-1}^{(j)}/t$ where $\Sigma_{t-1}^{(j)} = \sum_{s = 1}^{t-1} g_{s,j}$ is the cumulative subgradient~\citep{podkopaev2024adaptive}; the hidden iterate is $\tilde{x}_{t,j} = \max(\beta_t^{(j)} W_{t-1}^{(j)}, 0)$ and the wealth follows $W_t^{(j)} = W_{t-1}^{(j)} \cdot (1 - \beta_t^{(j)} g_{t,j})$. Unlike \UPOCP, the KT bettor uses a closed-form betting rule that is min-max optimal for symmetric coin-betting games (i.e., $\alpha_j = 1/2$); it does \emph{not} adapt to the asymmetry of the per-level pinball loss when $\alpha_j$ is small. This is the source of the under-performance of \KT at high target coverage in our experiments, and is the reason \UPOCP achieves the strictly tighter rate $\sqrt{\alpha_j (1 - \alpha_j) \log T / T}$.

\paragraph{Hyperparameters.} \KT is parameter-free; the initial wealth is fixed at $1$ across all datasets and levels, matching the theoretical analysis. No tuning is performed.

\paragraph{Projected vs.\ unprojected.} The projected variant \KT-\textsc{Iso} plays $\bx_t = \Pi_\sK(\bxtilde_t)$ on Line~\ref{ln:cw-kt-proj}; the unprojected variant plays $\bx_t = \bxtilde_t$.

\begin{algorithm}[ht]
\caption{Coordinate-wise Krichevsky--Trofimov bettor with lazy projection.}
\label{alg:cw-kt}
\begin{algorithmic}[1]
  \STATE \textbf{Input:} miscoverage levels $\alpha_1 < \cdots < \alpha_d$
  \STATE Initialize $W_0^{(j)} \leftarrow 1$ and cumulative gradient $\Sigma^{(j)} \leftarrow 0$ for each $j \in [d]$
  \FOR{$t = 1, 2, \dots$}
    \FOR{$j = 1, \dots, d$ (in parallel)}
      \STATE betting fraction $\beta_t^{(j)} \leftarrow -\Sigma^{(j)} / t$
      \STATE $\tilde{x}_{t,j} \leftarrow \max\big(\beta_t^{(j)} \cdot W_{t-1}^{(j)},\, 0\big)$ \quad (\emph{$0$ encodes $\emptyset$})
    \ENDFOR
    \STATE $\bx_t \leftarrow \Pi_\sK(\bxtilde_t)$ \quad (\emph{unprojected:} $\bx_t \leftarrow \bxtilde_t$);\quad observe $S_t$;\quad $\bg_t \in \partial \sL_t(\bx_t)$ via~\eqref{eq:pinball-subgradient} \label{ln:cw-kt-proj}
    \FOR{$j = 1, \dots, d$ (in parallel)}
      \STATE $W_t^{(j)} \leftarrow W_{t-1}^{(j)} \cdot \big(1 - \beta_t^{(j)} \, g_{t,j}\big)$;\quad $\Sigma^{(j)} \leftarrow \Sigma^{(j)} + g_{t,j}$
    \ENDFOR
  \ENDFOR
\end{algorithmic}
\end{algorithm}

\subsubsection{\MultiQT}
\label{app:multiqt-rules}

\paragraph{Algorithm.} \MultiQT~\citep{ding2025calibrated} is the lazy-projection-wrapped instance of online gradient descent: in our notation it coincides with Algorithm~\ref{alg:pnorm-omd} restricted to $p = 2$ (so $q = 2$), composed with the isotonic projection. Unlike \PNormOMD with constant step size, \MultiQT uses the data-driven adaptive step-size schedule from~\citet[Section~5]{ding2025calibrated}: at each round $t$,
\[
  \eta_t \;=\; \max\!\left(0.1 \cdot Q_{0.9}\!\left(\set{S_s : s \in [\max(1, t - 50),\, t-1]}\right),\; 0.1\right),
\]
i.e.\ a multiplicative factor of the empirical $90$th-percentile of the most recent $50$ scores, lower-bounded by $0.1$ to prevent collapse early in the stream. This schedule effectively scales the step with the magnitude of recent residuals, eliminating sensitivity to the score scale at the cost of introducing a sliding window and a multiplicative constant; both are inherited verbatim from~\citet{ding2025calibrated}. Because \MultiQT is by design a projected algorithm, we report only its projected form and use it as the prior baseline against which we compare every other (base learner, projection) combination.

\paragraph{Hyperparameters and tuning.} The window length ($50$), tail quantile ($0.9$), multiplicative factor ($0.1$), and lower bound ($0.1$) are all taken from the public implementation of~\citet{ding2025calibrated} without modification. The dual iterate is initialized at $\btheta_1 = \bzero$, matching \PNormOMD. As with \PNormOMD, this adaptive schedule is best understood as a tuned baseline: its constants reflect choices made by~\citet{ding2025calibrated} on their own benchmarks rather than on ours.

\subsection{Width distribution analysis}
\label{app:width-distributions}

\begin{figure}[H]
\centering
\includegraphics[width=\textwidth]{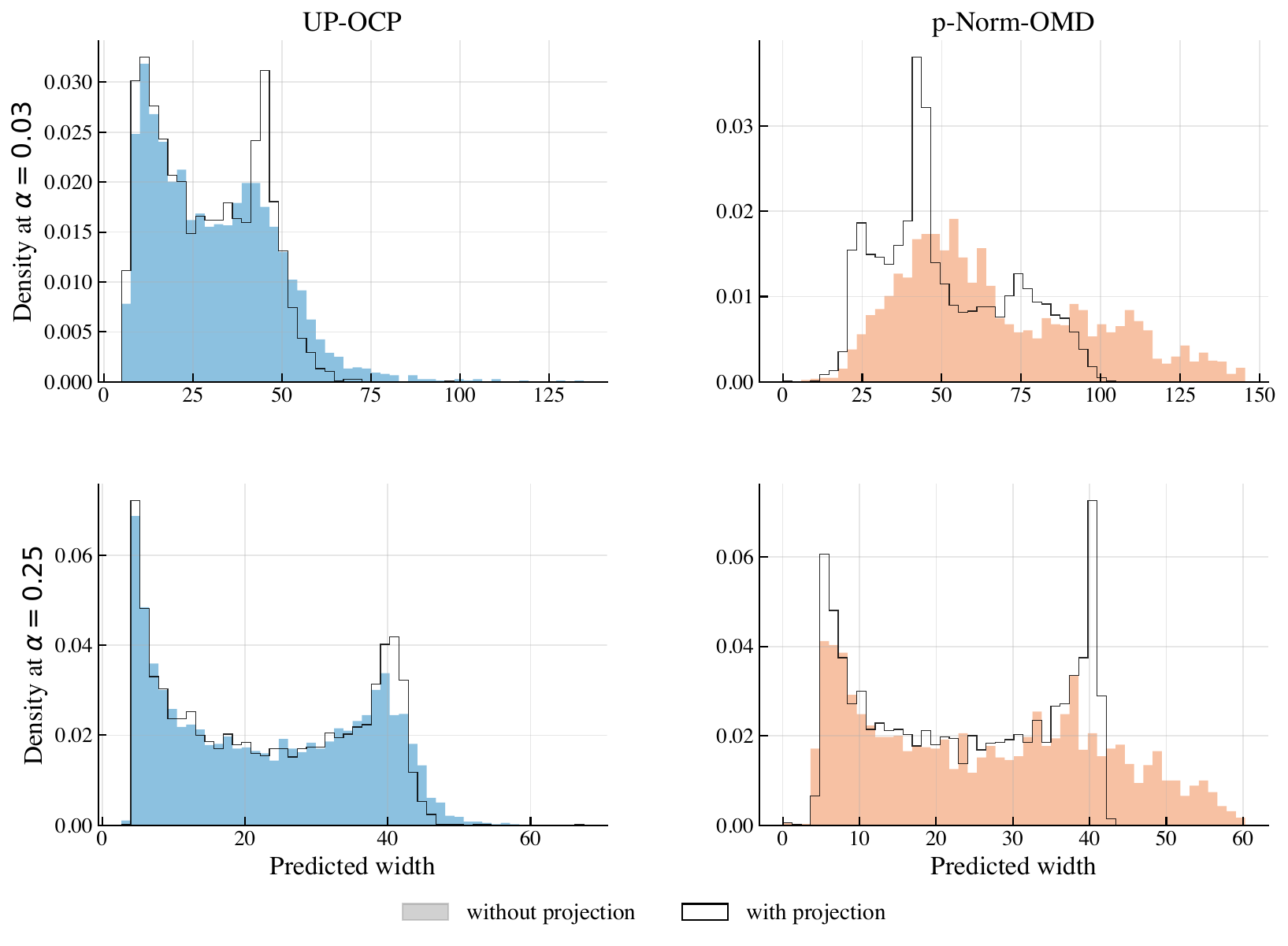}
\caption{Distribution of per-round prediction-set widths on the sinusoidal stream, at $\alpha = 0.03$ (top row) and $\alpha = 0.25$ (bottom row). Each panel overlays the histogram of widths for the unprojected base learner (filled, solid color) against the projected variant (outlined, black border), at the same $\alpha$ and base learner.}
\label{fig:paper-distributions}
\end{figure}

The mean and tail-quantile statistics in Section~\ref{sec:experiments} already give a rough picture of the projection's effect on sharpness, but they do not capture how it reshapes the underlying \emph{distribution} of widths. Figure~\ref{fig:paper-distributions} overlays the histograms of per-round predicted widths on the sinusoidal dataset at two representative levels, $\alpha = 0.03$ and $\alpha = 0.25$, for the two base learners \UPOCP\ and \PNormOMD that have a meaningful projected/unprojected pair.

The qualitative pattern is consistent across the two base learners. The projection consistently concentrates the distribution onto a tighter range: the heaviest right-tail mass is removed, and the overall distribution conforms closer to two narrower modes (the lower and higher modes correspond to the extrema pattern of the sinusoidal wave). From a downstream-decision perspective, this reshaping is more useful than a small mean improvement, because the cost of an unusually wide set is non-linear for most users (e.g., capacity reservation, contract pricing).

\subsection{Weighted Interval Score}
\label{app:wis-results}

Table~\ref{tab:wis_summary} extends Figure~\ref{fig:paper-wis} to every dataset. The projection lowers the WIS of \UPOCP and \PNormOMD on every dataset; for \KT, whose isotonicity violations are small (Table~\ref{tab:wis_within_run}), the two variants are within $0.3\%$ of each other. These comparisons are across two different runs, whose trajectories differ, so they are not covered by Theorem~\ref{thm:per-round-reduction}. Table~\ref{tab:wis_within_run} checks the theorem in its exact, within-run form: the hidden iterate violates isotonicity on $127{,}206$ of the $127{,}680$ post-burn-in rounds ($99.6\%$) across both datasets and the four projected methods, and on every one of them the projection strictly reduced both the WIS and its sharpness component without increasing the coverage penalty.

\begin{table}[H]
\centering
\caption{Average per-round WIS over all $d = 100$ levels (post burn-in; synthetic streams: mean over $10$ seeds), and its sharpness and coverage-penalty components. \textbf{Bold} marks the smallest WIS on each dataset.}
\label{tab:wis_summary}
\setlength{\tabcolsep}{4pt}
\renewcommand{\arraystretch}{1.05}
\resizebox{\textwidth}{!}{%
\begin{tabular}{@{}llrrrrrrr@{}}
\toprule
Dataset & & \textsc{UP-OCP} & \textsc{UP-OCP}-\textsc{Iso} & $p$-Norm-OMD & $p$-Norm-OMD-\textsc{Iso} & \textsc{KT} & \textsc{KT}-\textsc{Iso} & \textsc{MultiQT} \\
\midrule
AAPL & WIS & $456.4$ & $\mathbf{453.6}$ & $523.6$ & $466.5$ & $528.9$ & $528.3$ & $478.3$ \\
 & Sharpness & $394.5$ & $392.4$ & $451.7$ & $411.0$ & $394.1$ & $393.4$ & $405.0$ \\
 & Penalty & $61.8$ & $61.2$ & $71.9$ & $55.5$ & $134.8$ & $134.9$ & $73.3$ \\
\midrule
AMZN & WIS & $1746.6$ & $\mathbf{1734.3}$ & $2026.0$ & $1798.3$ & $2011.7$ & $2010.5$ & $1859.3$ \\
 & Sharpness & $1417.4$ & $1412.4$ & $1689.6$ & $1520.6$ & $1483.0$ & $1482.0$ & $1496.5$ \\
 & Penalty & $329.2$ & $321.9$ & $336.4$ & $277.7$ & $528.7$ & $528.5$ & $362.8$ \\
\midrule
GOOGL & WIS & $1777.3$ & $1771.6$ & $1944.6$ & $\mathbf{1756.4}$ & $2063.4$ & $2064.9$ & $1915.3$ \\
 & Sharpness & $1393.7$ & $1387.4$ & $1596.8$ & $1473.6$ & $1450.8$ & $1451.5$ & $1463.3$ \\
 & Penalty & $383.6$ & $384.2$ & $347.8$ & $282.9$ & $612.7$ & $613.4$ & $451.9$ \\
\midrule
\texttt{elec2} & WIS & $2.04$ & $2.03$ & $2.12$ & $2.06$ & $1.98$ & $\mathbf{1.98}$ & $2.38$ \\
 & Sharpness & $1.06$ & $1.05$ & $1.30$ & $1.25$ & $1.01$ & $1.00$ & $1.63$ \\
 & Penalty & $0.97$ & $0.97$ & $0.82$ & $0.81$ & $0.97$ & $0.98$ & $0.75$ \\
\midrule
Sinusoidal & WIS & $609.2$ & $\mathbf{602.1}$ & $709.9$ & $621.4$ & $643.2$ & $641.6$ & $606.6$ \\
 & Sharpness & $569.0$ & $564.1$ & $635.2$ & $577.4$ & $576.7$ & $575.7$ & $583.3$ \\
 & Penalty & $40.2$ & $38.0$ & $74.7$ & $44.0$ & $66.6$ & $65.9$ & $23.3$ \\
\midrule
Stationary & WIS & $11863.2$ & $\mathbf{11753.0}$ & $12551.5$ & $11810.6$ & $14208.3$ & $14202.9$ & $13238.4$ \\
 & Sharpness & $8770.9$ & $8671.8$ & $9867.2$ & $9267.3$ & $9128.4$ & $9114.4$ & $9669.0$ \\
 & Penalty & $3092.3$ & $3081.2$ & $2684.3$ & $2543.3$ & $5079.9$ & $5088.4$ & $3569.3$ \\
\midrule
Mixed & WIS & $8271.5$ & $\mathbf{8232.0}$ & $9378.0$ & $8484.8$ & $10008.3$ & $10010.9$ & $9212.5$ \\
 & Sharpness & $6006.8$ & $5961.4$ & $7271.2$ & $6490.6$ & $6150.6$ & $6149.1$ & $6637.6$ \\
 & Penalty & $2264.7$ & $2270.6$ & $2106.8$ & $1994.1$ & $3857.7$ & $3861.8$ & $2574.8$ \\
\bottomrule
\end{tabular}}
\end{table}

\begin{table}[H]
\centering
\caption{Within-run check of Theorem~\ref{thm:per-round-reduction}. For each projected method, the hidden iterate $\bxtilde_t$ and the played iterate $\bx_t = \Pi_\sK(\bxtilde_t)$ are logged at every post-burn-in round (sinusoidal: all $10$ seeds). Columns: share of rounds with $\bxtilde_t \notin \sK$; share of those rounds on which the WIS and its sharpness component are strictly reduced, and the coverage penalty is not increased (up to $10^{-12}$ relative rounding); median and maximum per-round relative WIS reduction $1 - \WIS(\bx_t)/\WIS(\bxtilde_t)$; average WIS of the hidden and of the played iterates.}
\label{tab:wis_within_run}
\setlength{\tabcolsep}{4pt}
\renewcommand{\arraystretch}{1.05}
\resizebox{\textwidth}{!}{%
\begin{tabular}{@{}llrrrrrr@{}}
\toprule
Dataset & Method & $\bxtilde_t \notin \sK$ & WIS, sharpness $\downarrow$, penalty $\not\uparrow$ & median red. & max red. & avg.\ WIS hidden & avg.\ WIS played \\
\midrule
AAPL & \textsc{UP-OCP}-\textsc{Iso} & $99.97\%$ & $100\%$ & $0.93\%$ & $39.9\%$ & $467.0$ & $453.6$ \\
 & $p$-Norm-OMD-\textsc{Iso} & $99.01\%$ & $100\%$ & $22.43\%$ & $80.1\%$ & $584.9$ & $466.5$ \\
 & \textsc{KT}-\textsc{Iso} & $99.93\%$ & $100\%$ & $0.03\%$ & $24.4\%$ & $529.1$ & $528.3$ \\
 & \textsc{MultiQT} & $99.86\%$ & $100\%$ & $0.28\%$ & $48.0\%$ & $482.2$ & $478.3$ \\
\midrule
Sinusoidal & \textsc{UP-OCP}-\textsc{Iso} & $99.97\%$ & $100\%$ & $4.51\%$ & $50.6\%$ & $644.9$ & $602.1$ \\
 & $p$-Norm-OMD-\textsc{Iso} & $99.10\%$ & $100\%$ & $23.45\%$ & $64.6\%$ & $828.4$ & $621.4$ \\
 & \textsc{KT}-\textsc{Iso} & $99.76\%$ & $100\%$ & $0.17\%$ & $27.1\%$ & $652.3$ & $641.6$ \\
 & \textsc{MultiQT} & $99.66\%$ & $100\%$ & $1.74\%$ & $31.6\%$ & $619.7$ & $606.6$ \\
\bottomrule
\end{tabular}}
\end{table}

\subsection{Step-size sensitivity and adaptive step sizes for $p$-norm OMD}
\label{app:eta-sensitivity}

Corollary~\ref{cor:p-norm} favors $\eta \to \infty$, whereas $F_T(\bu)$ diverges as $\eta \to \infty$ for every $\bu$. The two measure different things: small miscoverage alone is a weak target, attained even by data-free predictors~\citep[App.~H.5]{liu2026online}, while the regret bound controls efficiency, since it bounds half the WIS (Section~\ref{sec:projection-benefits}) of the played sets against every comparator. Our step-size sweep confirms the trade-off(Appendices~\ref{app:p-norm} and~\ref{app:eta-sensitivity}).

\paragraph{Sensitivity.} Table~\ref{tab:eta_sweep} reports, for projected $p$-norm OMD with $p = 2 \log d / (2 \log d - 1)$, the coverage error (mean over all $d = 100$ levels of $\abs{\Cov_{T,i} - (1 - \alpha_i)}$, post burn-in) and the average per-round WIS over the whole step-size grid. The coverage error decreases monotonically in $\eta$, as Corollary~\ref{cor:p-norm} suggests, while the WIS is minimized at a finite step size ($\eta = 64$ on AAPL and $\eta = 32$ on the sinusoidal stream) and roughly doubles between $\eta = 64$ and $\eta = 1024$. Every $\eta \in [16, 128]$ stays within about $10\%$ of the best WIS on both datasets, and the same $\eta = 64$ is within about $2.2\%$ of the best WIS on both streams, which are of very different nature (stock residuals and a synthetic wave).

\begin{table}[H]
\centering
\caption{Projected $p$-norm OMD: coverage error (\%, mean over all levels) / average per-round WIS, for each constant step size $\eta$ of the grid and for the plug-in step size~\eqref{eq:plug-in-eta}.}
\label{tab:eta_sweep}
\setlength{\tabcolsep}{4pt}
\renewcommand{\arraystretch}{1.05}
\resizebox{\textwidth}{!}{%
\begin{tabular}{@{}lrrrrrrrrrrrrrr@{}}
\toprule
$\eta$ & $0.25$ & $0.5$ & $1$ & $2$ & $4$ & $8$ & $16$ & $32$ & $64$ & $128$ & $256$ & $512$ & $1024$ & plug-in \\
\midrule
AAPL & $36.86$ & $28.16$ & $21.01$ & $14.88$ & $9.52$ & $5.46$ & $2.80$ & $1.38$ & $0.64$ & $0.31$ & $0.14$ & $0.06$ & $0.05$ & $3.31$ \\
 & $1075.8$ & $969.8$ & $853.5$ & $736.0$ & $639.1$ & $559.7$ & $500.0$ & $468.3$ & $\mathbf{466.5}$ & $500.0$ & $568.8$ & $706.6$ & $934.9$ & $482.9$ \\
\midrule
Sinusoidal & $48.12$ & $34.29$ & $20.27$ & $10.49$ & $5.40$ & $2.64$ & $1.26$ & $0.61$ & $0.30$ & $0.15$ & $0.08$ & $0.04$ & $0.02$ & $0.29$ \\
 & $1467.1$ & $1234.4$ & $1060.6$ & $940.9$ & $813.9$ & $688.3$ & $625.5$ & $\mathbf{608.0}$ & $621.4$ & $669.9$ & $767.6$ & $930.9$ & $1231.4$ & $629.3$ \\
\bottomrule
\end{tabular}}
\end{table}

\paragraph{A plug-in step size.} The balanced step size of Section~\ref{sub:pnorm}, $\eta^\star = 2 (p - 1) B_T / (d^{2/q} T)$, depends on the unknown average budget $B_T / T$. Replacing it by its running estimate gives the step size used for the update after round $t \geq 2$ (in our experiments, updates start after the warm-up),
\begin{equation}
\label{eq:plug-in-eta}
  \eta_t = \frac{2 (p - 1)}{d^{2/q}} \cdot \frac{1}{t - 1} \sum_{\tau = 1}^{t - 1} \sL_\tau(\bzero),
  \qquad \sL_\tau(\bzero) = S_\tau \sum_{i = 1}^d (1 - \alpha_i),
\end{equation}
which uses only revealed scores. Its WIS is about $3.5\%$ above the grid optimum on both datasets (Table~\ref{tab:eta_sweep}); its coverage error on the sinusoidal stream ($0.29\%$) matches that of the grid-selected $\eta = 64$ ($0.30\%$), while on AAPL it shows a mild under-coverage of $3.3\%$, comparable to the parameter-free methods. This schedule is only a heuristic: $(\eta_t)$ need not be monotone, so the standard analysis of OMD with non-increasing step sizes does not apply, and we claim no bound for it.

\paragraph{A doubling-trick variant with guarantees.} Run Algorithm~\ref{alg:pnorm-omd} in epochs $k = 0, 1, \dots$, restarting the base learner at the beginning of each epoch (the projection layer is stateless), with the constant step size $\eta_k = 2 (p - 1) b_k / d^{2/q}$ and $b_k = 2^k b_0$, where $b_0 = s_0 \sum_{i = 1}^d (1 - \alpha_i)$ for some $s_0 > 0$. Epoch $k$ ends after the first round at which the average of $\sL_\tau(\bzero)$ over the rounds of the epoch exceeds $2 b_k$. On the first $n_k - 1$ of its $n_k$ rounds, the epoch's budget is at most $2 b_k (n_k - 1)$, so Theorem~\ref{thm:level-set} applied to them, as in Corollary~\ref{cor:p-norm} with $q = 2 \log d$, gives $\norm{\bG^{(k)}}_q \leq \sqrt{3 e (2 \log d - 1) n_k} + \sqrt{e}$, where the last term accounts for the last round. Summing over the $K$ epochs and applying Cauchy--Schwarz,
\[
  \max_{i \in [d]} \MisCov_{T, i}
  \leq \sqrt{\frac{3 e (2 \log d - 1) K}{T}} + \frac{\sqrt{e}\, K}{T}.
\]
Since $\sL_\tau(\bzero) = S_\tau \sum_i (1 - \alpha_i) \leq D T^\gamma \sum_i (1 - \alpha_i)$ under Assumption~\ref{ass:poly-growth}, no epoch ends once $2 b_k \geq D T^\gamma \sum_i (1 - \alpha_i)$, hence $K \leq 1 + \log_2(D T^\gamma / s_0) = \sO(\gamma \log T + \log(D / s_0))$, independently of $d$~\citep[cf.][Sec.~2.3]{cesa2006prediction}. For bounded scores ($\gamma = 0$), this is the balanced rate up to a factor $\sqrt{3K/2}$ that depends only on $D / s_0$; under polynomial growth it pays a $\sqrt{\log T}$ factor, multiplicatively, whereas Theorem~\ref{thm:cw-upocp-per-level} pays it additively. In both cases the step size stays bounded, so the regret bound does not diverge.

\subsection{Varying the number of levels}
\label{app:d-ablation}

Table~\ref{tab:d_ablation} reports the realized worst-level miscoverage error $\max_i \MisCov_{T,i}$ on the sinusoidal stream ($T = 3{,}000$, levels uniformly spaced in $[0.001, 0.5]$, $10$ seeds; unlike Section~\ref{sec:experiments}, these runs use no warm-up period, so the $d = 100$ column differs slightly from the main experiments) as the number of levels grows from $d = 5$ to $d = 1{,}000$, for projected $p$-norm OMD with the balanced step size of Section~\ref{sub:pnorm}, projected coordinate-wise \UPOCP, and \MultiQT. The errors of \UPOCP-\textsc{Iso} and \MultiQT are flat in $d$, and that of $p$-norm OMD grows slowly while staying an order of magnitude below that of \MultiQT. The $\sqrt{d}$ and $\sqrt{\log d}$ rates are worst-case guarantees, and on this non-adversarial stream the ablation exhibits neither scaling; we do not claim that it does. What it shows is that the methods designed through the tighter analysis are more accurate at every $d$: $p$-norm OMD by an order of magnitude over \MultiQT, and \UPOCP by about $40\%$, with no degradation over more than two orders of magnitude in $d$. Small coverage error is only half of the picture, since it can be bought with width: the mean widths of \UPOCP-\textsc{Iso} and \MultiQT are within $1\%$ of each other for $d \geq 100$, while $p$-norm OMD with the balanced step size pays $23$--$36\%$ more mean width than \MultiQT for its smaller error.

\begin{table}[H]
\centering
\caption{Worst-level miscoverage error $\max_i \MisCov_{T,i}$ on the sinusoidal stream as a function of the number of levels $d$ (mean over $10$ seeds; standard errors are at most $1.5 \times 10^{-4}$).}
\label{tab:d_ablation}
\setlength{\tabcolsep}{5pt}
\renewcommand{\arraystretch}{1.05}
\begin{tabular}{@{}lrrrrrrrr@{}}
\toprule
$d$ & $5$ & $10$ & $20$ & $50$ & $100$ & $200$ & $500$ & $1000$ \\
\midrule
$p$-Norm-OMD-\textsc{Iso} & $0.0010$ & $0.0017$ & $0.0020$ & $0.0030$ & $0.0034$ & $0.0040$ & $0.0047$ & $0.0053$ \\
\textsc{UP-OCP}-\textsc{Iso} & $0.0377$ & $0.0379$ & $0.0379$ & $0.0380$ & $0.0380$ & $0.0380$ & $0.0378$ & $0.0381$ \\
\MultiQT & $0.0635$ & $0.0639$ & $0.0640$ & $0.0643$ & $0.0644$ & $0.0645$ & $0.0646$ & $0.0648$ \\
\bottomrule
\end{tabular}
\end{table}

\subsection{Additional results}
\label{app:experiments-results}

This subsection collects, for every dataset, the full set of artifacts referenced from Section~\ref{sec:experiments}: the calibration curve and three sharpness--coverage Pareto frontiers (using mean, median, and Q$_{75}$ widths) on a single 2$\times$2 grid, and a coverage--width summary table at three target levels $1 - \alpha \in \{0.95, 0.9, 0.8\}$ including \KT and its projected variant. The conclusions of Section~\ref{sec:experiments} carry over qualitatively to every dataset; we record only the deviations from the AAPL baseline.

For each tuned baseline (\PNormOMD with $p$ from Corollary~\ref{cor:p-norm}), the constant step size $\eta$ is selected by per-dataset grid search over $\eta \in \{2^{-2}, 2^{-1}, 1, 2, 4, 8, 16, 32, 64, 128, 256, 512, 1024\}$, exactly as in Section~\ref{sec:experiments}. The per-dataset selected $\eta$ is reported in each subsubsection.

\newpage

\subsubsection{AAPL (full results)}
\label{app:results-aapl}

\begin{figure}[H]
\centering
\includegraphics[width=\textwidth]{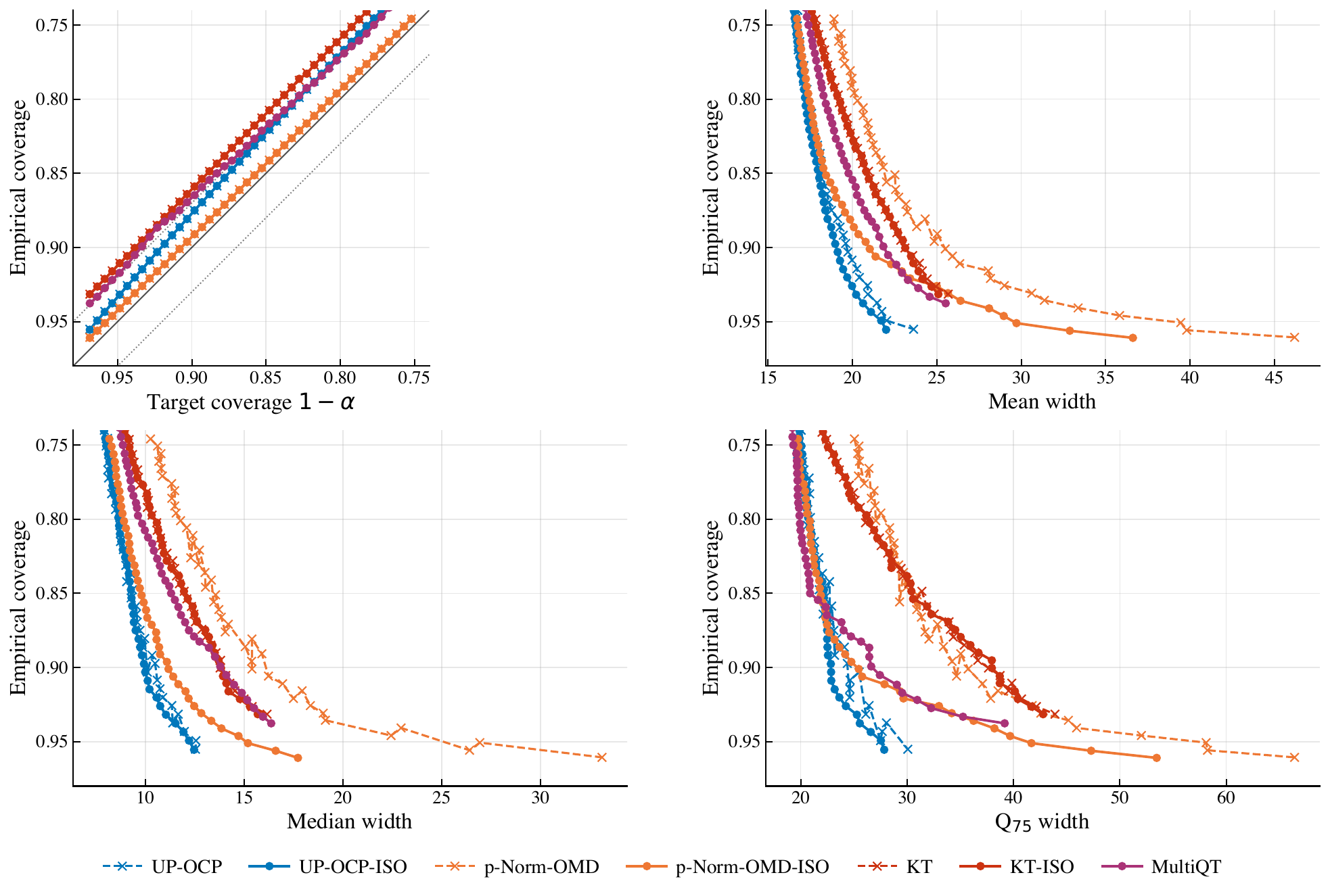}
\caption{AAPL: calibration (top-left) and three width Pareto frontiers (mean, median, Q$_{75}$) versus empirical coverage. \KT and \KT-\textsc{Iso} are now included, exposing their wider mean and median widths but competitive Q$_{75}$ tail.}
\label{fig:appendix-aapl-grid}
\end{figure}

This is the same dataset and forecasting pipeline as in Section~\ref{sec:experiments}; we extend Figure~\ref{fig:paper-main} with \KT and \KT-\textsc{Iso} (omitted there for scale) and add median and Q$_{75}$ width Pareto frontiers (Figure~\ref{fig:appendix-aapl-grid}). Tuned $\eta = 64$ for \PNormOMD with $p$ optimal. Table~\ref{tab:AAPL_summary} reports coverage and widths at three target levels.

\begin{table}[t]
\centering
\caption{AAPL: empirical coverage and prediction-set widths at three representative miscoverage levels. \textbf{Bold} marks the best value in each column (closest empirical coverage to the target; smallest width).}
\label{tab:AAPL_summary}
\setlength{\tabcolsep}{4pt}
\renewcommand{\arraystretch}{1.05}
\resizebox{\textwidth}{!}{%
\begin{tabular}{l|cccc|cccc|cccc}
\toprule
Method & \multicolumn{4}{c}{$1 - \alpha = 0.95$} & \multicolumn{4}{c}{$1 - \alpha = 0.9$} & \multicolumn{4}{c}{$1 - \alpha = 0.8$} \\
\cmidrule(lr){2-5}\cmidrule(lr){6-9}\cmidrule(lr){10-13}
 & Cov. & Mean width & Median width & Q$_{95}$ width & Cov. & Mean width & Median width & Q$_{95}$ width & Cov. & Mean width & Median width & Q$_{95}$ width \\
\midrule
\textsc{UP-OCP} & $0.93$ & $20.92$ & $11.67$ & $73.35$ & $0.88$ & $18.75$ & $9.73$ & $69.41$ & $0.77$ & $16.98$ & $\mathbf{8.14}$ & $66.16$ \\
\textsc{UP-OCP}-\textsc{Iso} & $0.93$ & $\mathbf{20.22}$ & $\mathbf{11.05}$ & $\mathbf{70.53}$ & $0.88$ & $\mathbf{18.39}$ & $\mathbf{9.51}$ & $\mathbf{67.17}$ & $0.77$ & $\mathbf{16.85}$ & $8.26$ & $65.07$ \\
\midrule
$p$-Norm-OMD & $0.94$ & $33.37$ & $22.95$ & $99.63$ & $0.89$ & $25.02$ & $15.90$ & $81.06$ & $0.80$ & $20.11$ & $11.56$ & $69.90$ \\
$p$-Norm-OMD-\textsc{Iso} & $\mathbf{0.94}$ & $28.10$ & $13.85$ & $101.58$ & $\mathbf{0.89}$ & $20.35$ & $10.75$ & $71.39$ & $\mathbf{0.80}$ & $17.38$ & $8.85$ & $\mathbf{64.99}$ \\
\midrule
\textsc{KT} & $0.91$ & $23.97$ & $14.28$ & $75.22$ & $0.86$ & $21.33$ & $12.34$ & $70.26$ & $0.76$ & $18.20$ & $9.44$ & $66.24$ \\
\textsc{KT}-\textsc{Iso} & $0.91$ & $23.60$ & $14.10$ & $72.86$ & $0.86$ & $21.24$ & $12.40$ & $70.02$ & $0.76$ & $18.11$ & $9.36$ & $65.63$ \\
\midrule
\textsc{MultiQT} & $0.92$ & $22.94$ & $14.88$ & $71.40$ & $0.86$ & $20.26$ & $11.81$ & $67.97$ & $0.77$ & $17.85$ & $9.26$ & $65.64$ \\
\bottomrule
\end{tabular}}
\end{table}

\newpage
\subsubsection{AMZN}
\label{app:results-amzn}

Same Prophet pipeline as AAPL (\citet{taylor2018forecasting}, log-prices, daily refit). Tuned $\eta = 256$ for \PNormOMD with $p$ optimal.

\begin{figure}[H]
\centering
\includegraphics[width=\textwidth]{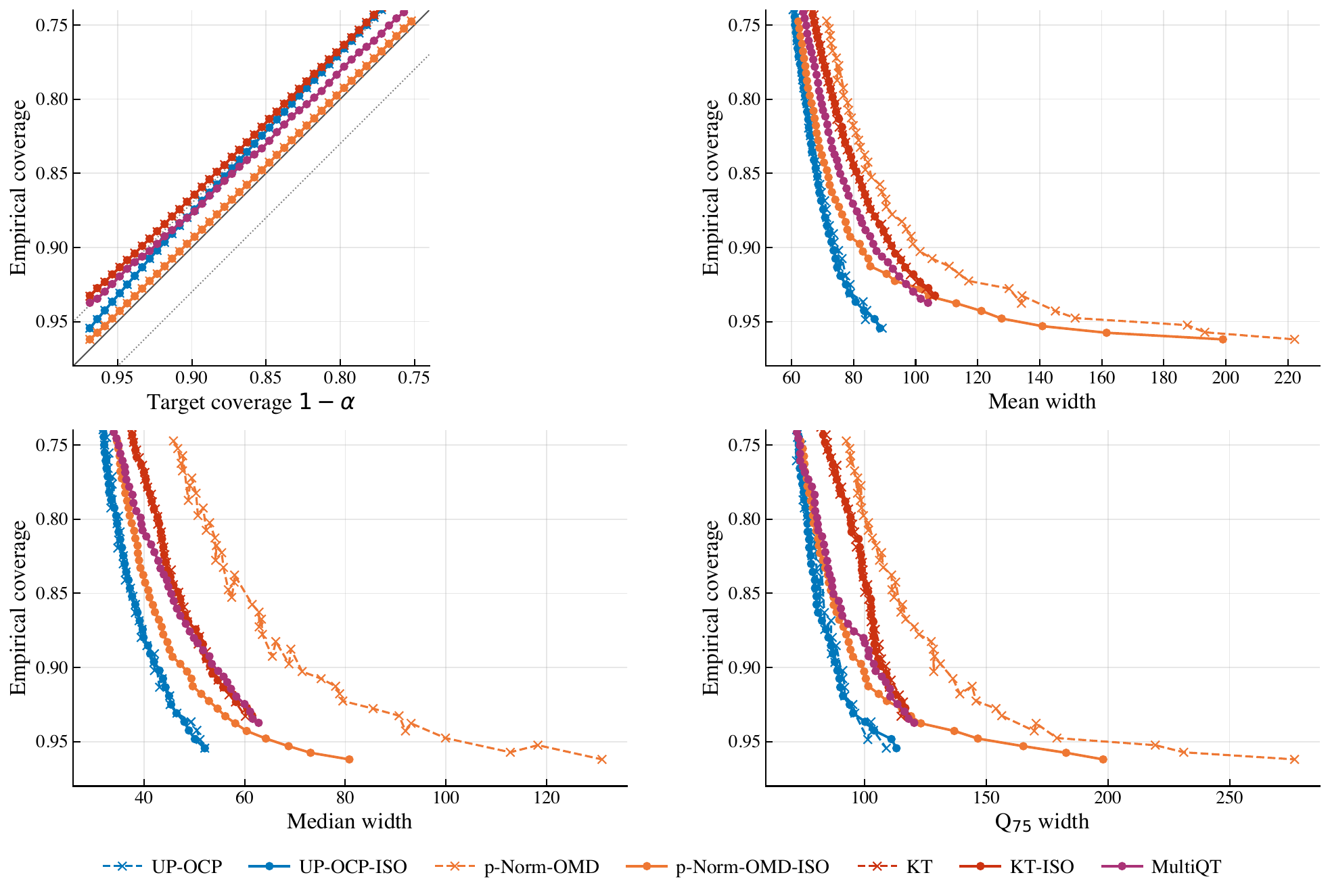}
\caption{AMZN: calibration and three width Pareto frontiers. Behavior is qualitatively identical to AAPL.}
\label{fig:appendix-amzn-grid}
\end{figure}

\begin{table}[t]
\centering
\caption{AMZN: empirical coverage and prediction-set widths at three representative miscoverage levels. \textbf{Bold} marks the best value in each column (closest empirical coverage to the target; smallest width).}
\label{tab:AMZN_summary}
\setlength{\tabcolsep}{4pt}
\renewcommand{\arraystretch}{1.05}
\resizebox{\textwidth}{!}{%
\begin{tabular}{l|cccc|cccc|cccc}
\toprule
Method & \multicolumn{4}{c}{$1 - \alpha = 0.95$} & \multicolumn{4}{c}{$1 - \alpha = 0.9$} & \multicolumn{4}{c}{$1 - \alpha = 0.8$} \\
\cmidrule(lr){2-5}\cmidrule(lr){6-9}\cmidrule(lr){10-13}
 & Cov. & Mean width & Median width & Q$_{95}$ width & Cov. & Mean width & Median width & Q$_{95}$ width & Cov. & Mean width & Median width & Q$_{95}$ width \\
\midrule
\textsc{UP-OCP} & $0.93$ & $79.01$ & $46.54$ & $285.27$ & $0.87$ & $71.19$ & $\mathbf{39.58}$ & $267.82$ & $0.77$ & $62.54$ & $33.70$ & $\mathbf{240.78}$ \\
\textsc{UP-OCP}-\textsc{Iso} & $0.93$ & $\mathbf{78.57}$ & $\mathbf{46.45}$ & $\mathbf{277.33}$ & $0.87$ & $\mathbf{70.21}$ & $39.73$ & $\mathbf{261.89}$ & $0.77$ & $\mathbf{62.25}$ & $\mathbf{32.77}$ & $241.45$ \\
\midrule
$p$-Norm-OMD & $\mathbf{0.94}$ & $145.04$ & $91.97$ & $514.42$ & $0.89$ & $98.53$ & $65.50$ & $324.27$ & $\mathbf{0.80}$ & $76.74$ & $50.70$ & $266.98$ \\
$p$-Norm-OMD-\textsc{Iso} & $0.94$ & $121.20$ & $60.39$ & $475.34$ & $\mathbf{0.89}$ & $78.89$ & $45.65$ & $287.50$ & $0.80$ & $65.85$ & $37.14$ & $253.93$ \\
\midrule
\textsc{KT} & $0.91$ & $97.00$ & $55.96$ & $328.28$ & $0.86$ & $83.57$ & $48.39$ & $287.34$ & $0.77$ & $69.39$ & $40.16$ & $248.53$ \\
\textsc{KT}-\textsc{Iso} & $0.91$ & $96.60$ & $55.74$ & $325.48$ & $0.86$ & $83.66$ & $48.10$ & $288.93$ & $0.77$ & $69.58$ & $39.97$ & $246.44$ \\
\midrule
\textsc{MultiQT} & $0.92$ & $94.68$ & $58.32$ & $326.20$ & $0.88$ & $81.05$ & $49.08$ & $298.66$ & $0.78$ & $68.05$ & $37.70$ & $263.80$ \\
\bottomrule
\end{tabular}}
\end{table}

\newpage
\subsubsection{GOOGL}
\label{app:results-googl}

Same Prophet pipeline as AAPL. Tuned $\eta = 256$ for \PNormOMD with $p$ optimal. The qualitative pattern matches AAPL/AMZN; the projection improves both tail and mean widths of \UPOCP and \PNormOMD without affecting calibration, while \KT-\textsc{Iso} is marginally wider than \KT in several cells.

\begin{figure}[H]
\centering
\includegraphics[width=\textwidth]{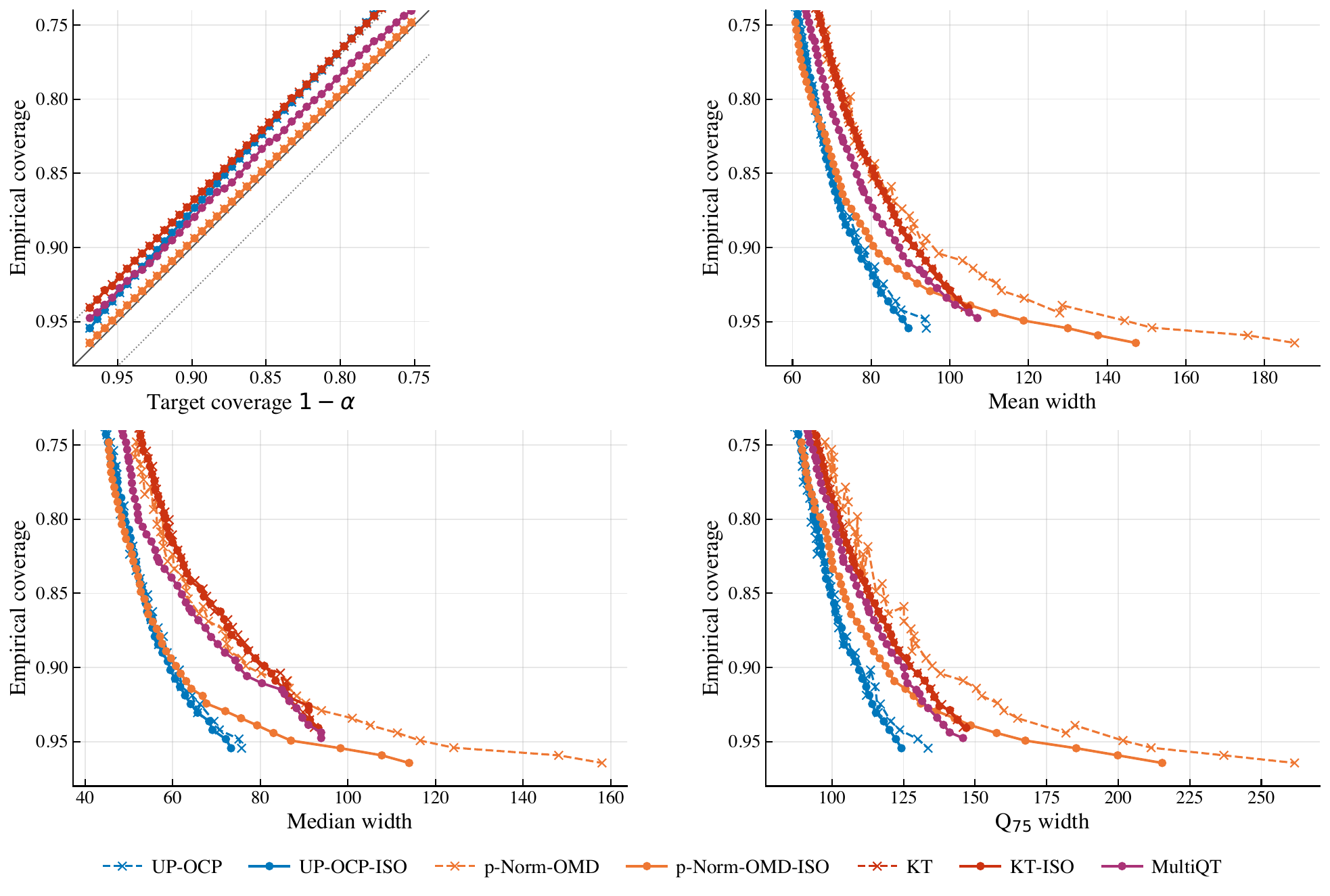}
\caption{GOOGL: calibration and three width Pareto frontiers.}
\label{fig:appendix-googl-grid}
\end{figure}

\begin{table}[t]
\centering
\caption{GOOGL: empirical coverage and prediction-set widths at three representative miscoverage levels. \textbf{Bold} marks the best value in each column (closest empirical coverage to the target; smallest width).}
\label{tab:GOOGL_summary}
\setlength{\tabcolsep}{4pt}
\renewcommand{\arraystretch}{1.05}
\resizebox{\textwidth}{!}{%
\begin{tabular}{l|cccc|cccc|cccc}
\toprule
Method & \multicolumn{4}{c}{$1 - \alpha = 0.95$} & \multicolumn{4}{c}{$1 - \alpha = 0.9$} & \multicolumn{4}{c}{$1 - \alpha = 0.8$} \\
\cmidrule(lr){2-5}\cmidrule(lr){6-9}\cmidrule(lr){10-13}
 & Cov. & Mean width & Median width & Q$_{95}$ width & Cov. & Mean width & Median width & Q$_{95}$ width & Cov. & Mean width & Median width & Q$_{95}$ width \\
\midrule
\textsc{UP-OCP} & $0.93$ & $82.74$ & $\mathbf{65.69}$ & $191.80$ & $0.87$ & $72.47$ & $56.68$ & $176.64$ & $0.77$ & $63.29$ & $47.28$ & $160.16$ \\
\textsc{UP-OCP}-\textsc{Iso} & $0.93$ & $\mathbf{82.49}$ & $65.69$ & $\mathbf{187.07}$ & $0.87$ & $\mathbf{72.17}$ & $\mathbf{55.36}$ & $173.32$ & $0.77$ & $\mathbf{63.24}$ & $\mathbf{47.12}$ & $159.41$ \\
\midrule
$p$-Norm-OMD & $\mathbf{0.94}$ & $127.93$ & $111.31$ & $286.65$ & $\mathbf{0.89}$ & $93.94$ & $75.78$ & $228.67$ & $0.80$ & $74.74$ & $57.80$ & $191.89$ \\
$p$-Norm-OMD-\textsc{Iso} & $0.94$ & $111.35$ & $83.02$ & $279.91$ & $0.89$ & $79.47$ & $59.60$ & $195.80$ & $\mathbf{0.80}$ & $64.69$ & $48.29$ & $163.53$ \\
\midrule
\textsc{KT} & $0.92$ & $96.45$ & $86.90$ & $187.69$ & $0.87$ & $84.03$ & $72.58$ & $173.19$ & $0.77$ & $69.33$ & $55.41$ & $153.74$ \\
\textsc{KT}-\textsc{Iso} & $0.92$ & $97.17$ & $86.45$ & $192.08$ & $0.87$ & $84.06$ & $71.74$ & $173.09$ & $0.77$ & $69.43$ & $55.29$ & $154.01$ \\
\midrule
\textsc{MultiQT} & $0.93$ & $96.73$ & $88.28$ & $189.76$ & $0.88$ & $81.45$ & $68.76$ & $\mathbf{172.88}$ & $0.79$ & $67.69$ & $51.35$ & $\mathbf{153.22}$ \\
\bottomrule
\end{tabular}}
\end{table}

\newpage
\subsubsection{Electricity demand (\texttt{elec2})}
\label{app:results-elec2}

Half-hourly New South Wales electricity demand from~\citep{harries1999splice}, with one-step-ahead forecasts produced by an autoregressive model (lag 3, refit at every step). Scores are absolute residuals on the demand variable, normalized to lie roughly in $[0, 1]$ --- hence the smaller numerical scale relative to the stock datasets. Tuned $\eta = 0.25$ for \PNormOMD with $p$ optimal.

\begin{figure}[H]
\centering
\includegraphics[width=\textwidth]{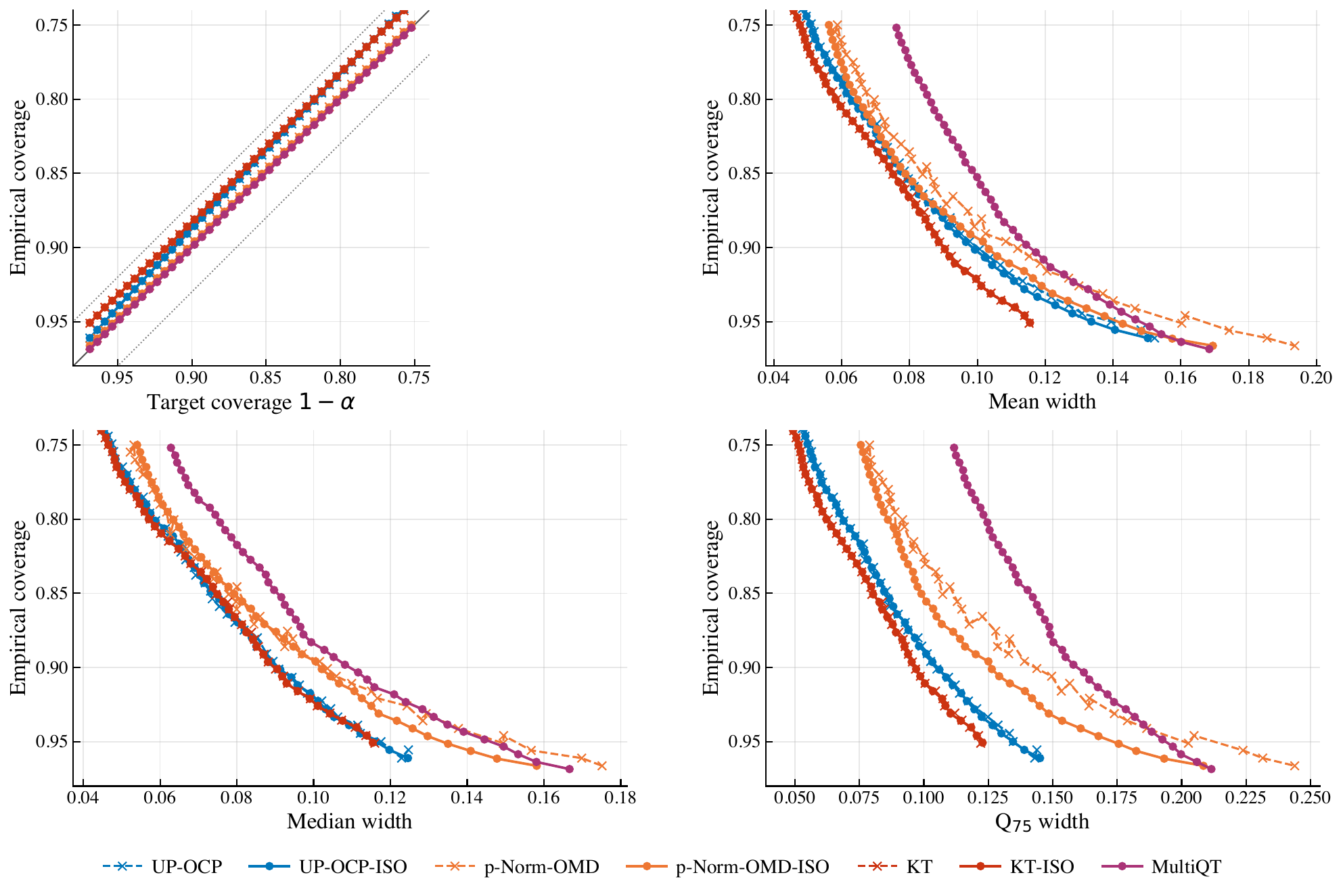}
\caption{\texttt{elec2}: calibration and three width Pareto frontiers.}
\label{fig:appendix-elec2-grid}
\end{figure}

\begin{table}[t]
\centering
\caption{elec2: empirical coverage and prediction-set widths at three representative miscoverage levels. \textbf{Bold} marks the best value in each column (closest empirical coverage to the target; smallest width).}
\label{tab:elec2_summary}
\setlength{\tabcolsep}{4pt}
\renewcommand{\arraystretch}{1.05}
\resizebox{\textwidth}{!}{%
\begin{tabular}{l|cccc|cccc|cccc}
\toprule
Method & \multicolumn{4}{c}{$1 - \alpha = 0.95$} & \multicolumn{4}{c}{$1 - \alpha = 0.9$} & \multicolumn{4}{c}{$1 - \alpha = 0.8$} \\
\cmidrule(lr){2-5}\cmidrule(lr){6-9}\cmidrule(lr){10-13}
 & Cov. & Mean width & Median width & Q$_{95}$ width & Cov. & Mean width & Median width & Q$_{95}$ width & Cov. & Mean width & Median width & Q$_{95}$ width \\
\midrule
\textsc{UP-OCP} & $0.94$ & $0.13$ & $0.11$ & $0.16$ & $0.89$ & $0.09$ & $0.09$ & $0.12$ & $0.79$ & $0.06$ & $0.06$ & $0.08$ \\
\textsc{UP-OCP}-\textsc{Iso} & $0.94$ & $0.12$ & $0.11$ & $0.15$ & $0.89$ & $0.09$ & $0.09$ & $0.12$ & $0.79$ & $0.06$ & $0.05$ & $0.08$ \\
\midrule
$p$-Norm-OMD & $0.95$ & $0.16$ & $0.15$ & $0.29$ & $0.90$ & $0.11$ & $0.10$ & $0.20$ & $0.80$ & $0.07$ & $0.06$ & $0.14$ \\
$p$-Norm-OMD-\textsc{Iso} & $0.95$ & $0.14$ & $0.13$ & $0.23$ & $0.90$ & $0.10$ & $0.10$ & $0.16$ & $0.80$ & $0.06$ & $0.06$ & $0.11$ \\
\midrule
\textsc{KT} & $0.93$ & $0.10$ & $\mathbf{0.10}$ & $0.12$ & $0.88$ & $0.08$ & $0.08$ & $0.10$ & $0.78$ & $0.05$ & $0.05$ & $0.07$ \\
\textsc{KT}-\textsc{Iso} & $0.93$ & $\mathbf{0.10}$ & $0.10$ & $\mathbf{0.12}$ & $0.88$ & $\mathbf{0.08}$ & $\mathbf{0.08}$ & $\mathbf{0.10}$ & $0.78$ & $\mathbf{0.05}$ & $\mathbf{0.05}$ & $\mathbf{0.07}$ \\
\midrule
\textsc{MultiQT} & $\mathbf{0.95}$ & $0.15$ & $0.14$ & $0.25$ & $\mathbf{0.90}$ & $0.11$ & $0.11$ & $0.23$ & $\mathbf{0.80}$ & $0.09$ & $0.08$ & $0.20$ \\
\bottomrule
\end{tabular}}
\end{table}

On this stream the relative ordering shifts slightly: \KT-\textsc{Iso} attains the smallest mean, median, and Q$_{95}$ widths in every column at $1 - \alpha \in \{0.9, 0.8\}$, while \UPOCP-\textsc{Iso} is within about $2\%$ in median width but about $8\%$ wider in mean and up to $22\%$ wider in Q$_{95}$ width. The reason is the small score range and the absence of fat tails on this stream: the variance budget $\alpha_j(1 - \alpha_j)$ that \UPOCP exploits is dominated here by the constant per-step quantization, which is precisely the regime where the simpler \KT update rule is competitive with the universal-portfolio update.

\newpage
\subsubsection{Sinusoidal (synthetic)}
\label{app:results-sinusoidal}

\begin{figure}[H]
\centering
\includegraphics[width=\textwidth]{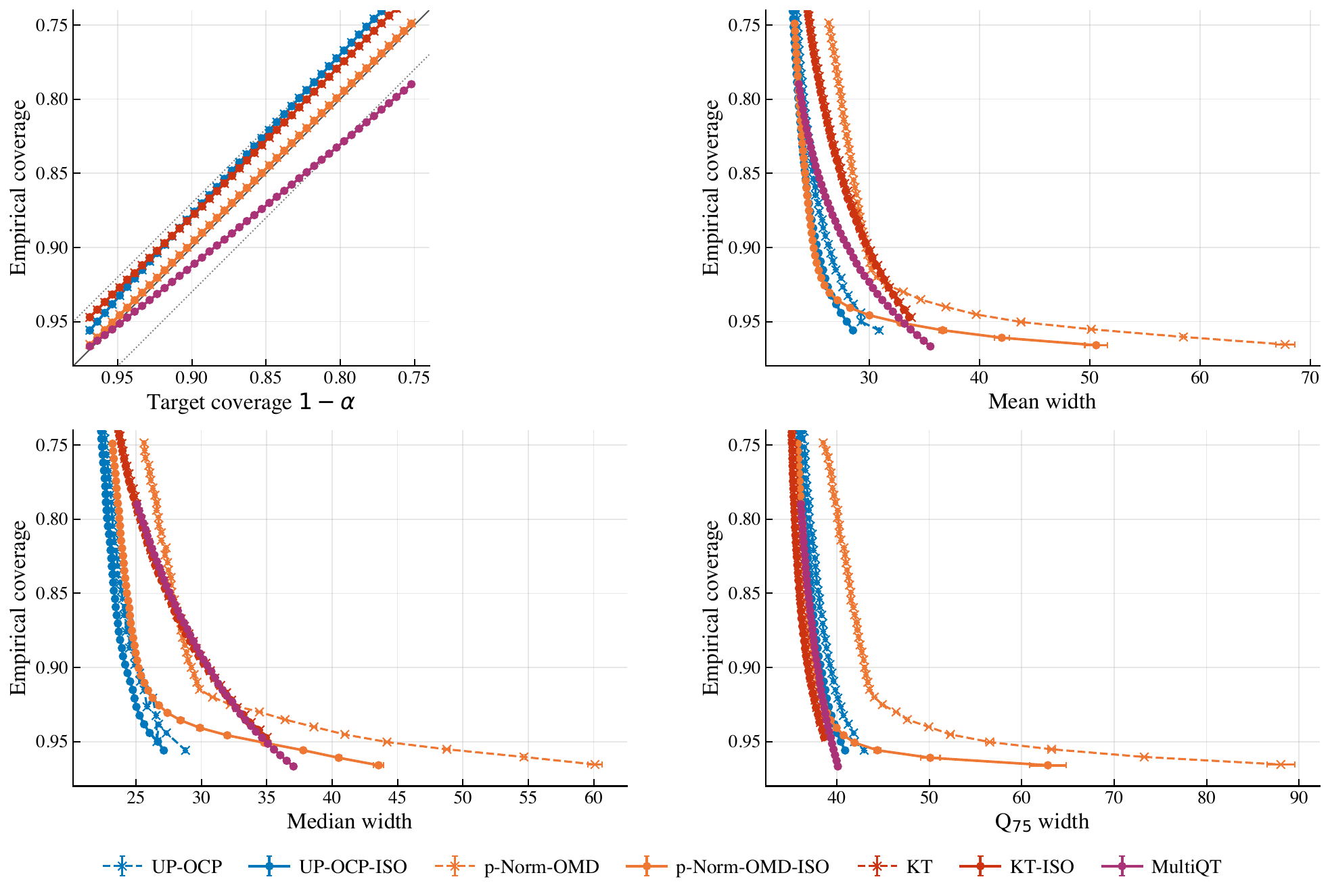}
\caption{Sinusoidal: calibration and three width Pareto frontiers. \UPOCP-\textsc{Iso} dominates in mean and median width across the entire $\alpha$ range, while \KT and \MultiQT lead on Q$_{75}$, mirroring the bounded twin-extrema structure of the wave.}
\label{fig:appendix-sinusoidal-grid}
\end{figure}

We generate the nonconformity scores $S_t$ as a sinusoidal wave distorted by Gaussian noise. With $T = 3{,}000$, $\alpha_{\min} = 2$, $\alpha_{\max} = 20$, period $P = 200$, noise scale $\sigma = 0.3$, and $\epsilon_t \mathop{\sim}\limits^{\text{i.i.d.}} \sN(0, \sigma^2)$, define the centre and amplitude $c \coloneqq  (\alpha_{\min} + \alpha_{\max})/2$ and $a \coloneqq  (\alpha_{\max} - \alpha_{\min})/2$ and set
\[
  S_t \;=\; \max\!\big(0,\; c + a \sin(2\pi t/P) + \epsilon_t\big),
  \qquad t = 1, \dots, T.
\]
Concretely $c = 11$, $a = 9$. Reported quantities are means over $10$ independent seeds (base seed $42$, seeds $42, 43, \dots, 51$) with standard errors. Tuned $\eta = 64$ for \PNormOMD with $p$ optimal. This is the same stream as in Section~\ref{sec:experiments}; we extend Figure~\ref{fig:paper-main} with \KT and \KT-\textsc{Iso} and the median and Q$_{75}$ frontiers (Figure~\ref{fig:appendix-sinusoidal-grid}), and Table~\ref{tab:sinusoidal_summary} reports the per-method coverage and width summary at three target levels.

\begin{table}[t]
\centering
\caption{sinusoidal: empirical coverage and prediction-set widths at three representative miscoverage levels. \textbf{Bold} marks the best value in each column (closest empirical coverage to the target; smallest width).}
\label{tab:sinusoidal_summary}
\setlength{\tabcolsep}{4pt}
\renewcommand{\arraystretch}{1.05}
\resizebox{\textwidth}{!}{%
\begin{tabular}{l|cccc|cccc|cccc}
\toprule
Method & \multicolumn{4}{c}{$1 - \alpha = 0.95$} & \multicolumn{4}{c}{$1 - \alpha = 0.9$} & \multicolumn{4}{c}{$1 - \alpha = 0.8$} \\
\cmidrule(lr){2-5}\cmidrule(lr){6-9}\cmidrule(lr){10-13}
 & Cov. & Mean width & Median width & Q$_{95}$ width & Cov. & Mean width & Median width & Q$_{95}$ width & Cov. & Mean width & Median width & Q$_{95}$ width \\
\midrule
\textsc{UP-OCP} & $0.93_{\scriptscriptstyle \pm .00}$ & $28.10_{\scriptscriptstyle \pm .02}$ & $26.54_{\scriptscriptstyle \pm .01}$ & $54.42_{\scriptscriptstyle \pm .03}$ & $0.88_{\scriptscriptstyle \pm .00}$ & $25.47_{\scriptscriptstyle \pm .02}$ & $24.24_{\scriptscriptstyle \pm .03}$ & $49.18_{\scriptscriptstyle \pm .06}$ & $0.77_{\scriptscriptstyle \pm .00}$ & $23.72_{\scriptscriptstyle \pm .01}$ & $22.91_{\scriptscriptstyle \pm .03}$ & $45.26_{\scriptscriptstyle \pm .04}$ \\
\textsc{UP-OCP}-\textsc{Iso} & $0.93_{\scriptscriptstyle \pm .00}$ & $\mathbf{26.58_{\scriptscriptstyle \pm .05}}$ & $\mathbf{25.28_{\scriptscriptstyle \pm .09}}$ & $50.12_{\scriptscriptstyle \pm .19}$ & $0.88_{\scriptscriptstyle \pm .00}$ & $\mathbf{24.66_{\scriptscriptstyle \pm .02}}$ & $\mathbf{23.69_{\scriptscriptstyle \pm .05}}$ & $46.42_{\scriptscriptstyle \pm .08}$ & $0.77_{\scriptscriptstyle \pm .00}$ & $\mathbf{23.32_{\scriptscriptstyle \pm .01}}$ & $\mathbf{22.61_{\scriptscriptstyle \pm .03}}$ & $43.84_{\scriptscriptstyle \pm .07}$ \\
\midrule
$p$-Norm-OMD & $0.95_{\scriptscriptstyle \pm .00}$ & $39.72_{\scriptscriptstyle \pm .20}$ & $40.97_{\scriptscriptstyle \pm .19}$ & $67.35_{\scriptscriptstyle \pm .35}$ & $0.89_{\scriptscriptstyle \pm .00}$ & $29.59_{\scriptscriptstyle \pm .02}$ & $28.98_{\scriptscriptstyle \pm .05}$ & $58.67_{\scriptscriptstyle \pm .08}$ & $0.80_{\scriptscriptstyle \pm .00}$ & $27.48_{\scriptscriptstyle \pm .02}$ & $26.72_{\scriptscriptstyle \pm .03}$ & $55.22_{\scriptscriptstyle \pm .05}$ \\
$p$-Norm-OMD-\textsc{Iso} & $0.95_{\scriptscriptstyle \pm .00}$ & $30.04_{\scriptscriptstyle \pm .14}$ & $31.99_{\scriptscriptstyle \pm .23}$ & $49.55_{\scriptscriptstyle \pm .43}$ & $\mathbf{0.90_{\scriptscriptstyle \pm .00}}$ & $24.89_{\scriptscriptstyle \pm .03}$ & $25.09_{\scriptscriptstyle \pm .04}$ & $42.73_{\scriptscriptstyle \pm .08}$ & $\mathbf{0.80_{\scriptscriptstyle \pm .00}}$ & $23.67_{\scriptscriptstyle \pm .02}$ & $23.73_{\scriptscriptstyle \pm .03}$ & $41.69_{\scriptscriptstyle \pm .05}$ \\
\midrule
\textsc{KT} & $0.93_{\scriptscriptstyle \pm .00}$ & $31.86_{\scriptscriptstyle \pm .04}$ & $32.75_{\scriptscriptstyle \pm .04}$ & $41.38_{\scriptscriptstyle \pm .04}$ & $0.88_{\scriptscriptstyle \pm .00}$ & $28.82_{\scriptscriptstyle \pm .02}$ & $29.05_{\scriptscriptstyle \pm .03}$ & $41.09_{\scriptscriptstyle \pm .02}$ & $0.78_{\scriptscriptstyle \pm .00}$ & $25.35_{\scriptscriptstyle \pm .01}$ & $24.80_{\scriptscriptstyle \pm .03}$ & $40.72_{\scriptscriptstyle \pm .03}$ \\
\textsc{KT}-\textsc{Iso} & $0.93_{\scriptscriptstyle \pm .00}$ & $31.78_{\scriptscriptstyle \pm .05}$ & $32.68_{\scriptscriptstyle \pm .05}$ & $\mathbf{41.24_{\scriptscriptstyle \pm .03}}$ & $0.88_{\scriptscriptstyle \pm .00}$ & $28.58_{\scriptscriptstyle \pm .02}$ & $28.81_{\scriptscriptstyle \pm .03}$ & $40.94_{\scriptscriptstyle \pm .02}$ & $0.78_{\scriptscriptstyle \pm .00}$ & $25.21_{\scriptscriptstyle \pm .02}$ & $24.62_{\scriptscriptstyle \pm .03}$ & $40.60_{\scriptscriptstyle \pm .02}$ \\
\midrule
\textsc{MultiQT} & $\mathbf{0.95_{\scriptscriptstyle \pm .00}}$ & $33.21_{\scriptscriptstyle \pm .04}$ & $35.09_{\scriptscriptstyle \pm .05}$ & $41.34_{\scriptscriptstyle \pm .05}$ & $0.91_{\scriptscriptstyle \pm .00}$ & $28.87_{\scriptscriptstyle \pm .04}$ & $31.29_{\scriptscriptstyle \pm .05}$ & $\mathbf{40.89_{\scriptscriptstyle \pm .04}}$ & $0.83_{\scriptscriptstyle \pm .00}$ & $24.73_{\scriptscriptstyle \pm .02}$ & $26.79_{\scriptscriptstyle \pm .04}$ & $\mathbf{40.52_{\scriptscriptstyle \pm .03}}$ \\
\bottomrule
\end{tabular}}
\end{table}

\newpage
\subsubsection{Stationary trend with random waves (synthetic)}
\label{app:results-stationary}

Nonconformity scores follow a constant baseline distorted by sparse, heavy-tailed multiplicative spikes and smoothed by a centred rolling maximum. With baseline $C = 10$, spike probability $p = 0.1$, exponential scale $\sigma = 10$, rolling window $W = 25$, and $T = 3{,}000$, we draw
\[
  B_t \sim \text{Bernoulli}(p),
  \qquad
  E_t \sim \text{Exponential}(1/\sigma),
  \qquad
  \tilde S_t \;=\; C\,\big(1 + B_t E_t\big),
\]
and obtain $S_t$ by a centered rolling-maximum smoother of half-width $\lfloor W/2 \rfloor$:
\[
  S_t \;=\; \max_{\tau \in [t - \lfloor W/2 \rfloor,\; t + \lfloor W/2 \rfloor]} \tilde S_\tau.
\]
This stresses heavy tails without an underlying trend. We report means over $10$ seeds (base seed $42$). Tuned $\eta = 1024$ for \PNormOMD with $p$ optimal.

\begin{figure}[H]
\centering
\includegraphics[width=\textwidth]{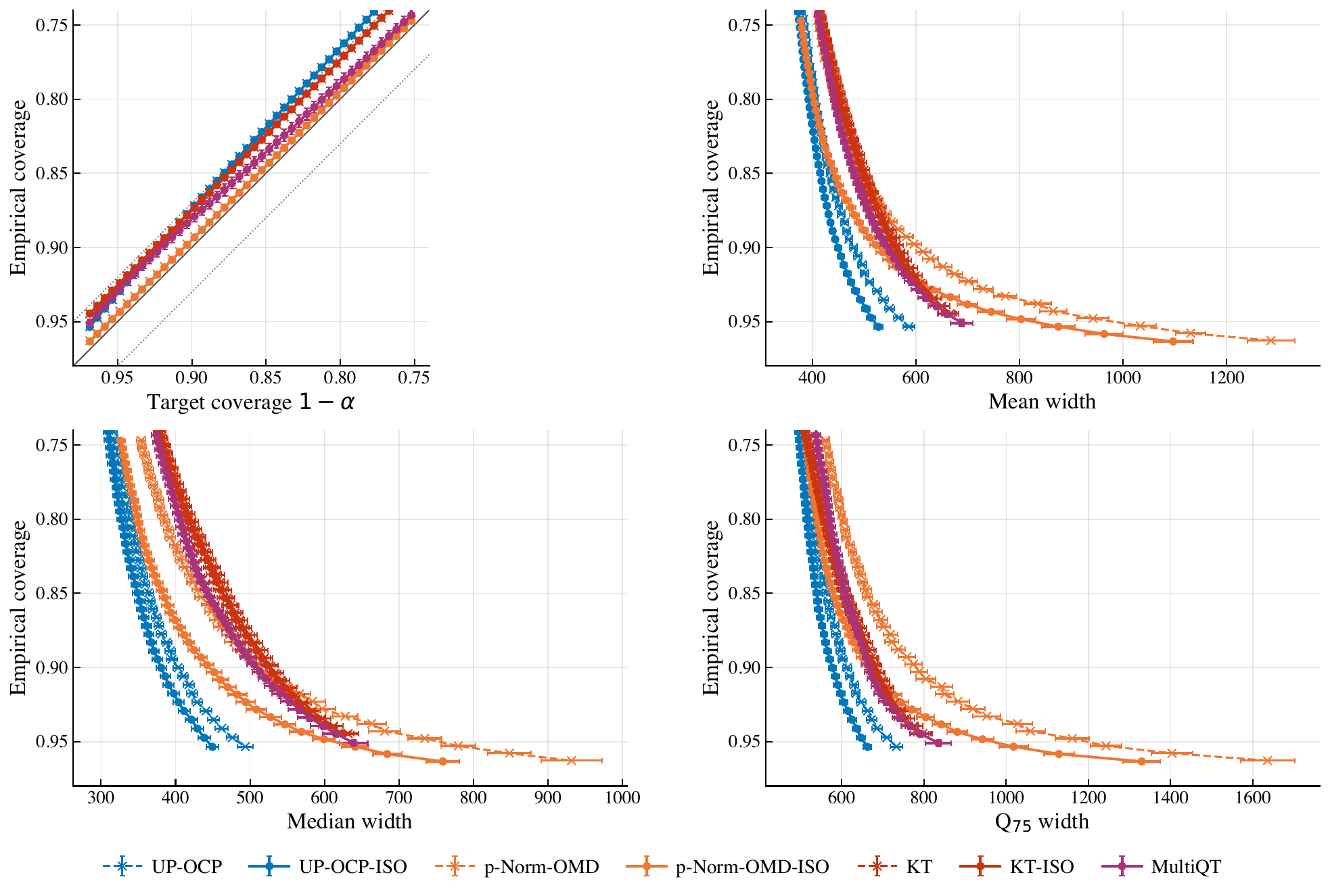}
\caption{Stationary trend with random waves: calibration and three width Pareto frontiers.}
\label{fig:appendix-stationary-grid}
\end{figure}

\begin{table}[t]
\centering
\caption{stationary: empirical coverage and prediction-set widths at three representative miscoverage levels. \textbf{Bold} marks the best value in each column (closest empirical coverage to the target; smallest width).}
\label{tab:stationary_summary}
\setlength{\tabcolsep}{4pt}
\renewcommand{\arraystretch}{1.05}
\resizebox{\textwidth}{!}{%
\begin{tabular}{l|cccc|cccc|cccc}
\toprule
Method & \multicolumn{4}{c}{$1 - \alpha = 0.95$} & \multicolumn{4}{c}{$1 - \alpha = 0.9$} & \multicolumn{4}{c}{$1 - \alpha = 0.8$} \\
\cmidrule(lr){2-5}\cmidrule(lr){6-9}\cmidrule(lr){10-13}
 & Cov. & Mean width & Median width & Q$_{95}$ width & Cov. & Mean width & Median width & Q$_{95}$ width & Cov. & Mean width & Median width & Q$_{95}$ width \\
\midrule
\textsc{UP-OCP} & $0.93_{\scriptscriptstyle \pm .00}$ & $525.10_{\scriptscriptstyle \pm 7.99}$ & $440.92_{\scriptscriptstyle \pm 7.39}$ & $1172_{\scriptscriptstyle \pm 29.01}$ & $0.87_{\scriptscriptstyle \pm .00}$ & $451.12_{\scriptscriptstyle \pm 6.20}$ & $377.72_{\scriptscriptstyle \pm 6.14}$ & $1026_{\scriptscriptstyle \pm 27.17}$ & $0.77_{\scriptscriptstyle \pm .00}$ & $391.88_{\scriptscriptstyle \pm 6.07}$ & $325.61_{\scriptscriptstyle \pm 6.29}$ & $912.26_{\scriptscriptstyle \pm 23.90}$ \\
\textsc{UP-OCP}-\textsc{Iso} & $0.93_{\scriptscriptstyle \pm .00}$ & $\mathbf{483.63_{\scriptscriptstyle \pm 7.43}}$ & $\mathbf{411.50_{\scriptscriptstyle \pm 7.77}}$ & $\mathbf{1059_{\scriptscriptstyle \pm 26.10}}$ & $0.87_{\scriptscriptstyle \pm .00}$ & $\mathbf{427.91_{\scriptscriptstyle \pm 6.05}}$ & $\mathbf{360.99_{\scriptscriptstyle \pm 6.36}}$ & $967.28_{\scriptscriptstyle \pm 24.87}$ & $0.77_{\scriptscriptstyle \pm .00}$ & $\mathbf{381.22_{\scriptscriptstyle \pm 5.34}}$ & $\mathbf{317.14_{\scriptscriptstyle \pm 5.90}}$ & $885.35_{\scriptscriptstyle \pm 22.26}$ \\
\midrule
$p$-Norm-OMD & $0.94_{\scriptscriptstyle \pm .00}$ & $865.25_{\scriptscriptstyle \pm 25.90}$ & $680.72_{\scriptscriptstyle \pm 21.55}$ & $2301_{\scriptscriptstyle \pm 102.11}$ & $0.89_{\scriptscriptstyle \pm .00}$ & $581.46_{\scriptscriptstyle \pm 12.87}$ & $497.91_{\scriptscriptstyle \pm 11.83}$ & $1347_{\scriptscriptstyle \pm 50.22}$ & $0.80_{\scriptscriptstyle \pm .00}$ & $441.37_{\scriptscriptstyle \pm 5.32}$ & $382.66_{\scriptscriptstyle \pm 6.51}$ & $1005_{\scriptscriptstyle \pm 20.14}$ \\
$p$-Norm-OMD-\textsc{Iso} & $\mathbf{0.94_{\scriptscriptstyle \pm .00}}$ & $745.34_{\scriptscriptstyle \pm 25.86}$ & $569.02_{\scriptscriptstyle \pm 15.81}$ & $2088_{\scriptscriptstyle \pm 130.20}$ & $\mathbf{0.89_{\scriptscriptstyle \pm .00}}$ & $510.48_{\scriptscriptstyle \pm 10.38}$ & $434.35_{\scriptscriptstyle \pm 8.26}$ & $1191_{\scriptscriptstyle \pm 48.89}$ & $\mathbf{0.80_{\scriptscriptstyle \pm .00}}$ & $401.20_{\scriptscriptstyle \pm 5.85}$ & $348.20_{\scriptscriptstyle \pm 5.97}$ & $915.13_{\scriptscriptstyle \pm 25.25}$ \\
\midrule
\textsc{KT} & $0.92_{\scriptscriptstyle \pm .00}$ & $608.70_{\scriptscriptstyle \pm 12.49}$ & $571.92_{\scriptscriptstyle \pm 10.53}$ & $1079_{\scriptscriptstyle \pm 41.79}$ & $0.87_{\scriptscriptstyle \pm .00}$ & $527.99_{\scriptscriptstyle \pm 12.18}$ & $491.64_{\scriptscriptstyle \pm 10.79}$ & $964.65_{\scriptscriptstyle \pm 37.28}$ & $0.78_{\scriptscriptstyle \pm .00}$ & $439.18_{\scriptscriptstyle \pm 8.64}$ & $402.11_{\scriptscriptstyle \pm 8.33}$ & $825.50_{\scriptscriptstyle \pm 23.61}$ \\
\textsc{KT}-\textsc{Iso} & $0.92_{\scriptscriptstyle \pm .00}$ & $601.46_{\scriptscriptstyle \pm 11.81}$ & $565.92_{\scriptscriptstyle \pm 10.37}$ & $1067_{\scriptscriptstyle \pm 38.87}$ & $0.87_{\scriptscriptstyle \pm .00}$ & $523.56_{\scriptscriptstyle \pm 11.66}$ & $488.15_{\scriptscriptstyle \pm 10.63}$ & $\mathbf{952.76_{\scriptscriptstyle \pm 34.55}}$ & $0.78_{\scriptscriptstyle \pm .00}$ & $435.85_{\scriptscriptstyle \pm 8.20}$ & $398.80_{\scriptscriptstyle \pm 8.15}$ & $\mathbf{817.11_{\scriptscriptstyle \pm 21.49}}$ \\
\midrule
\textsc{MultiQT} & $0.93_{\scriptscriptstyle \pm .00}$ & $605.17_{\scriptscriptstyle \pm 13.54}$ & $566.08_{\scriptscriptstyle \pm 14.59}$ & $1102_{\scriptscriptstyle \pm 46.29}$ & $0.88_{\scriptscriptstyle \pm .00}$ & $512.82_{\scriptscriptstyle \pm 9.02}$ & $476.98_{\scriptscriptstyle \pm 9.11}$ & $995.40_{\scriptscriptstyle \pm 33.23}$ & $0.79_{\scriptscriptstyle \pm .00}$ & $436.87_{\scriptscriptstyle \pm 6.50}$ & $400.76_{\scriptscriptstyle \pm 7.38}$ & $902.39_{\scriptscriptstyle \pm 23.39}$ \\
\bottomrule
\end{tabular}}
\end{table}

The width scale on this stream is large ($\sim$$10^2$), reflecting the rolling-maximum smoothing of exponential spikes. The projection still strictly improves the unprojected base learner on every cell of the table, and the parameter-free \UPOCP-\textsc{Iso} achieves the smallest mean and median widths at every target (\KT-\textsc{Iso} has the smallest Q$_{95}$ width at $1 - \alpha \in \{0.9, 0.8\}$).

\newpage
\subsubsection{Quadratic drift with random waves (\texttt{mixed}, synthetic)}
\label{app:results-mixed}

Same multiplicative-spike-plus-rolling-max structure as the stationary stream, now overlaid on a quadratic drift. With $T = 3{,}000$, baseline trajectory $T_t = (20/T^2)\,t^2$ on $t \in [0, T-1]$, spike probability $p = 0.1$, exponential scale $\sigma = 10$, and rolling window $W = 25$, we draw
\[
  B_t \sim \text{Bernoulli}(p),
  \qquad
  E_t \sim \text{Exponential}(1/\sigma),
  \qquad
  \tilde S_t \;=\; T_t\,\big(1 + B_t E_t\big),
\]
and again take $S_t = \max_{\tau \in [t - \lfloor W/2 \rfloor, t + \lfloor W/2 \rfloor]} \tilde S_\tau$. The deterministic trend $T_t$ has quadratic shape in $t$ and ranges over $[0, 20]$ across the horizon. We report means over $10$ seeds (base seed $42$). Tuned $\eta = 1024$ for \PNormOMD with $p$ optimal.

\begin{figure}[H]
\centering
\includegraphics[width=\textwidth]{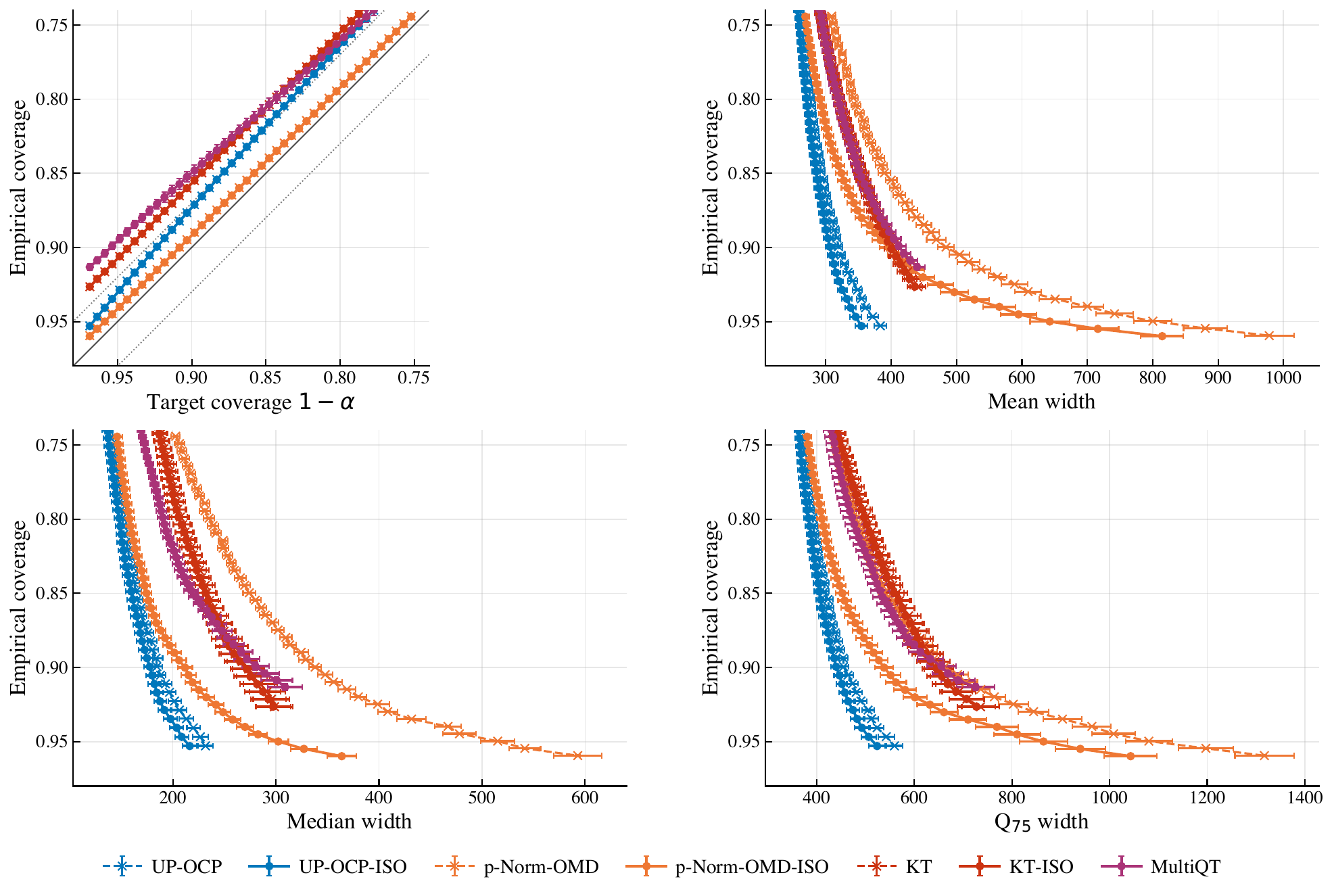}
\caption{Quadratic drift with random waves: calibration and three width Pareto frontiers.}
\label{fig:appendix-mixed-grid}
\end{figure}

\begin{table}[t]
\centering
\caption{mixed: empirical coverage and prediction-set widths at three representative miscoverage levels. \textbf{Bold} marks the best value in each column (closest empirical coverage to the target; smallest width).}
\label{tab:mixed_summary}
\setlength{\tabcolsep}{4pt}
\renewcommand{\arraystretch}{1.05}
\resizebox{\textwidth}{!}{%
\begin{tabular}{l|cccc|cccc|cccc}
\toprule
Method & \multicolumn{4}{c}{$1 - \alpha = 0.95$} & \multicolumn{4}{c}{$1 - \alpha = 0.9$} & \multicolumn{4}{c}{$1 - \alpha = 0.8$} \\
\cmidrule(lr){2-5}\cmidrule(lr){6-9}\cmidrule(lr){10-13}
 & Cov. & Mean width & Median width & Q$_{95}$ width & Cov. & Mean width & Median width & Q$_{95}$ width & Cov. & Mean width & Median width & Q$_{95}$ width \\
\midrule
\textsc{UP-OCP} & $0.93_{\scriptscriptstyle \pm .00}$ & $348.44_{\scriptscriptstyle \pm 7.14}$ & $206.36_{\scriptscriptstyle \pm 7.19}$ & $1158_{\scriptscriptstyle \pm 40.08}$ & $0.87_{\scriptscriptstyle \pm .00}$ & $303.10_{\scriptscriptstyle \pm 5.77}$ & $172.50_{\scriptscriptstyle \pm 5.22}$ & $1037_{\scriptscriptstyle \pm 34.24}$ & $0.77_{\scriptscriptstyle \pm .00}$ & $267.23_{\scriptscriptstyle \pm 5.44}$ & $143.56_{\scriptscriptstyle \pm 4.55}$ & $943.16_{\scriptscriptstyle \pm 35.66}$ \\
\textsc{UP-OCP}-\textsc{Iso} & $0.93_{\scriptscriptstyle \pm .00}$ & $\mathbf{325.17_{\scriptscriptstyle \pm 7.00}}$ & $\mathbf{191.73_{\scriptscriptstyle \pm 6.11}}$ & $\mathbf{1079_{\scriptscriptstyle \pm 37.51}}$ & $0.87_{\scriptscriptstyle \pm .00}$ & $\mathbf{293.20_{\scriptscriptstyle \pm 6.40}}$ & $\mathbf{166.90_{\scriptscriptstyle \pm 5.51}}$ & $\mathbf{1002_{\scriptscriptstyle \pm 36.87}}$ & $0.77_{\scriptscriptstyle \pm .00}$ & $\mathbf{262.85_{\scriptscriptstyle \pm 5.55}}$ & $\mathbf{141.34_{\scriptscriptstyle \pm 4.66}}$ & $930.76_{\scriptscriptstyle \pm 35.18}$ \\
\midrule
$p$-Norm-OMD & $0.94_{\scriptscriptstyle \pm .00}$ & $700.07_{\scriptscriptstyle \pm 24.17}$ & $466.97_{\scriptscriptstyle \pm 12.68}$ & $2249_{\scriptscriptstyle \pm 119.77}$ & $0.89_{\scriptscriptstyle \pm .00}$ & $460.82_{\scriptscriptstyle \pm 12.52}$ & $324.56_{\scriptscriptstyle \pm 6.93}$ & $1443_{\scriptscriptstyle \pm 65.93}$ & $0.79_{\scriptscriptstyle \pm .00}$ & $338.58_{\scriptscriptstyle \pm 6.98}$ & $231.17_{\scriptscriptstyle \pm 5.27}$ & $1070_{\scriptscriptstyle \pm 33.82}$ \\
$p$-Norm-OMD-\textsc{Iso} & $\mathbf{0.94_{\scriptscriptstyle \pm .00}}$ & $565.52_{\scriptscriptstyle \pm 24.49}$ & $270.02_{\scriptscriptstyle \pm 8.23}$ & $2122_{\scriptscriptstyle \pm 129.98}$ & $\mathbf{0.89_{\scriptscriptstyle \pm .00}}$ & $375.04_{\scriptscriptstyle \pm 12.77}$ & $201.36_{\scriptscriptstyle \pm 6.52}$ & $1294_{\scriptscriptstyle \pm 64.43}$ & $\mathbf{0.79_{\scriptscriptstyle \pm .00}}$ & $288.22_{\scriptscriptstyle \pm 6.59}$ & $156.79_{\scriptscriptstyle \pm 5.13}$ & $984.98_{\scriptscriptstyle \pm 35.75}$ \\
\midrule
\textsc{KT} & $0.91_{\scriptscriptstyle \pm .00}$ & $410.94_{\scriptscriptstyle \pm 11.68}$ & $277.76_{\scriptscriptstyle \pm 16.07}$ & $1283_{\scriptscriptstyle \pm 62.65}$ & $0.85_{\scriptscriptstyle \pm .00}$ & $358.83_{\scriptscriptstyle \pm 10.14}$ & $235.56_{\scriptscriptstyle \pm 10.34}$ & $1096_{\scriptscriptstyle \pm 43.06}$ & $0.76_{\scriptscriptstyle \pm .00}$ & $299.00_{\scriptscriptstyle \pm 7.45}$ & $192.79_{\scriptscriptstyle \pm 7.62}$ & $926.47_{\scriptscriptstyle \pm 33.12}$ \\
\textsc{KT}-\textsc{Iso} & $0.91_{\scriptscriptstyle \pm .00}$ & $407.09_{\scriptscriptstyle \pm 10.74}$ & $275.23_{\scriptscriptstyle \pm 16.03}$ & $1267_{\scriptscriptstyle \pm 53.10}$ & $0.85_{\scriptscriptstyle \pm .00}$ & $356.09_{\scriptscriptstyle \pm 9.65}$ & $234.22_{\scriptscriptstyle \pm 10.27}$ & $1086_{\scriptscriptstyle \pm 41.75}$ & $0.76_{\scriptscriptstyle \pm .00}$ & $297.40_{\scriptscriptstyle \pm 7.65}$ & $191.74_{\scriptscriptstyle \pm 7.54}$ & $\mathbf{921.67_{\scriptscriptstyle \pm 33.54}}$ \\
\midrule
\textsc{MultiQT} & $0.89_{\scriptscriptstyle \pm .00}$ & $404.10_{\scriptscriptstyle \pm 10.88}$ & $271.70_{\scriptscriptstyle \pm 12.32}$ & $1234_{\scriptscriptstyle \pm 43.23}$ & $0.85_{\scriptscriptstyle \pm .00}$ & $350.65_{\scriptscriptstyle \pm 9.81}$ & $217.81_{\scriptscriptstyle \pm 6.87}$ & $1111_{\scriptscriptstyle \pm 31.78}$ & $0.76_{\scriptscriptstyle \pm .00}$ & $300.23_{\scriptscriptstyle \pm 7.98}$ & $177.17_{\scriptscriptstyle \pm 3.82}$ & $996.18_{\scriptscriptstyle \pm 34.54}$ \\
\bottomrule
\end{tabular}}
\end{table}

\end{document}